\documentclass[12pt]{article} 
\pdfoutput=1
\usepackage{amsmath,amsthm,amssymb,bm,mathrsfs}
\usepackage{graphicx}
\usepackage[ruled,linesnumbered]{algorithm2e}
\usepackage{caption,enumerate,listings,setspace}
\usepackage[OT1]{fontenc}
\usepackage[colorlinks,linkcolor=red,anchorcolor=blue,citecolor=blue]{hyperref}
\usepackage[margin=1in]{geometry}
\usepackage[protrusion=false,expansion=true]{microtype}
\usepackage[title]{appendix}
\usepackage{bbm}
\usepackage{subcaption}
\usepackage{comment}
\theoremstyle{plain}
\usepackage{hyperref,bookmark}
\usepackage{multirow}
\usepackage{booktabs}       
\usepackage{enumitem}
\usepackage[dvipsnames]{xcolor}
\usepackage[authoryear]{natbib}

\usepackage{bibunits}

\newcommand{\argmin}{\mathop{\mathrm{argmin}}}

\newtheorem{lemma}{{\bf Lemma}}
\newtheorem{Example}{{\bf Example}}

\newtheorem{theorem}{{\bf Theorem}}

\newtheorem{assumption}{{\bf Assumption}}

\newtheorem{definition}{{\bf Definition}}

\usepackage{titlesec}
\titlespacing*{\section}
{-2pt}{-2pt}{-2pt}

\def\spacingset#1{\renewcommand{\baselinestretch}%
{#1}\small\normalsize} \spacingset{1}

\begin{document}

\title{\Large \bf 
Rate-Optimal Rank Aggregation with Private Pairwise Rankings
}
\author{Shirong Xu\thanks{Statistics and Data Science, University of California, Los Angeles. Email: shirongxu56@ucla.edu}
\and 
Will Wei Sun\thanks{Daniels School of Business, Purdue University. Email: sun244@purdue.edu}
\and
Guang Cheng\thanks{Statistics and Data Science, University of California, Los Angeles. Email: guangcheng@ucla.edu
}
}

\date{}
\maketitle

\begin{abstract}
In various real-world scenarios, such as recommender systems and political surveys, pairwise rankings are commonly collected and utilized for rank aggregation to derive an overall ranking of items. However, preference rankings can reveal individuals' personal preferences, highlighting the need to protect them from exposure in downstream analysis. In this paper, we address the challenge of preserving privacy while ensuring the utility of rank aggregation based on pairwise rankings generated from a general comparison model. A common privacy protection strategy in practice is the use of the randomized response mechanism to perturb raw pairwise rankings. However, a critical challenge arises because the privatized rankings no longer adhere to the original model, resulting in significant bias in downstream rank aggregation tasks. To address this, we propose an adaptive debiasing method for rankings from the randomized response mechanism, ensuring consistent estimation of true preferences and enhancing the utility of downstream rank aggregation. Theoretically, we provide insights into the relationship between overall privacy guarantees and estimation errors in private ranking data, and establish minimax rates for estimation errors. This enables the determination of optimal privacy guarantees that balance consistency in rank aggregation with privacy protection. We also investigate convergence rates of expected ranking errors for partial and full ranking recovery, quantifying how privacy protection affects the specification of top-$K$ item sets and complete rankings. Our findings are validated through extensive simulations and a real-world application.

\end{abstract}

\spacingset{1.7} 
\newpage

\section{Introduction}

Ranking data frequently arises in various scenarios, notably in recommender systems \citep{kalloori2018eliciting,oliveira2020rank}, political surveys \citep{ackerman2013elections, mccarthy2021ranked}, and search engines \citep{dwork2001rank}. Among ranking data, pairwise ranking stands out as a particular form that finds applications in collecting customer preferences and political investigations. Particularly, \citet{kalloori2018eliciting} developed a mobile app to collect users' pairwise comparisons in preference for ranking items, while \citet{ackerman2013elections} investigated the problem of collecting voters' pairwise comparisons to determine the committee members. The collected individual noisy rankings can be used for various rank aggregation analyses, such as estimating the true underlying ranking \citep{negahban2016rank}, specifying top-$K$ objects \citep{chen2015spectral, shah2018simple, AndersonZhang}, and inferring relative ranking of two objects \citep{liu2023lagrangian}.

In recent years, there has been a growing concern about the privacy of personal data \citep{dwork2006differential, bi2023distribution}, prompting numerous countries to respond by enacting specific regulations. One notable example is the General Data Protection Regulation (GDPR)\footnote{\url{https://gdpr-info.eu/}} in the European Union. The GDPR is designed to empower individuals with control over their personal data and mandates stringent data protection practices for businesses. Additionally, Canada has implemented the Personal Information Protection and Electronic Documents Act (PIPEDA)\footnote{\url{https://www.priv.gc.ca/en/privacy-topics/privacy-laws-in-canada/}} as a federal law governing the collection, use, and disclosure of personal data within the private sectors. These regulatory measures aim to address the evolving challenges associated with the protection of individuals' private information. 

As a distinctive form of data, ranking data also inherently carries sensitivity as it has the potential to reveal personal preferences \citep{ichihashi2020online} or political inclinations \citep{lee2015efficient}. In preference collection, users are asked to choose their preferred item in a pairwise comparison, which inevitably reveals their personal preferences for specific items. In recommender systems, this information can be exploited by malicious users for targeted advertising, allowing them to select advertisements that are most likely to result in conversions \citep{friedman2015privacy}. Protecting ranking data not only ensures regulatory compliance but also encourages data sharing without privacy concerns, minimizing potential misreporting due to privacy worries.

\begin{figure}[ht!]
\centering
\includegraphics[scale=0.32]{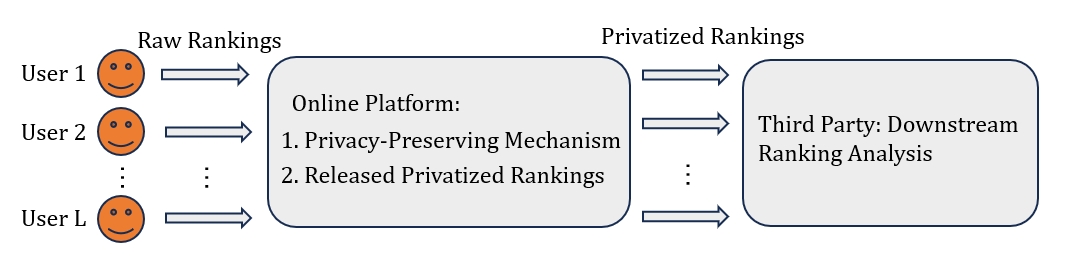}
\caption{Utility-Preserving Private Pairwise Ranking Mechanism.}
\label{Framework}
\end{figure}

Consider a scenario wherein a third party aims to collect pairwise rankings via a sequence of pairwise comparison questions through some online platforms such as survey websites \citep{carlson2017pairwise,tatli2024learning} or mobile apps \citep{kalloori2018eliciting}. The objective of the third party is to conduct the rank aggregation that integrates rankings from various users into a unified ranking that accurately reflects the collective preferences of the group. Hence, the objective of the online platform is to protect the privacy of the submitted pairwise rankings, while maintaining the utility of ranking data for the downstream rank aggregation task. In this context, a natural challenge for the online platform is understanding the fundamental limit in the privacy-utility tradeoff in rank aggregation and addressing the question of how strong privacy protection can be enforced on individuals' rankings given a specific number of respondents and items for comparison. See Figure \ref{Framework} for an illustration.

To model pairwise rankings, it is common to consider the class of linear stochastic transitivity (LST) models \citep{oliveira2018new}, in which $\bm{\theta}^\star = (\theta_1^\star, \ldots, \theta_m^\star)$ denotes the preference vector of a set of items $\Omega = \{1, \ldots, m\}$, and $\theta_i^\star > \theta_j^\star$ indicates that item $i$ is more preferred than item $j$ in population. Specifically, the LST models can be parameterized as:
\begin{align}
\label{General_Model_into}
\mathbb{P}\left(
\text{item } i \text{ is preferred over item } j
\right) = F(\theta_i^\star-\theta_j^\star),
\end{align}
where $F(\cdot)$ is the cumulative distribution function (CDF) of a zero-symmetric random variable. Typical examples include the Bradley-Terry-Luce (BTL) model \citep{bradley1952rank} and the Thurstone-Mosteller (TM) model \citep{thurstone1994law}, differing in the choice of $F(\cdot)$. 


The estimation of $\bm{\theta}^\star$ based on raw pairwise rankings has received increasing attentions \citep{negahban2016rank, chen2019spectral}. Nevertheless, protecting privacy of collected rankings in our framework inevitably distorts the raw distributions of rankings, rendering existing methods inapplicable. In practical scenarios, a widely employed method for protecting binary comparison data is the randomized response (RR) mechanism \citep{warner1965randomized}, wherein binary comparisons are flipped with a predetermined probability. 

Nonetheless, merely disclosing the private surrogate of ranking data using the classic RR mechanism is inadequate. This is because the privatized rankings no longer conform to the original comparison model outlined in (\ref{General_Model_into}), leading to notable biases in subsequent rank aggregation tasks. To address this challenge, in this paper we introduce an adaptive debiased RR mechanism aimed at achieving both differential privacy (DP; \citealt{dwork2006differential}) and enhancing utility in subsequent rank aggregation tasks. The conceptual framework developed in this paper is visually represented in Figure \ref{Framework}. Initially, individuals' rankings are collected through a ranking survey. These rankings then undergo permutation using a randomized mechanism and a subsequent debiasing-weighting step before being disclosed to a third party. The debiasing step is designed to align privatized rankings with the actual rankings in expectation. The weighting step aims to mitigate the impact of varying privacy guarantees. Following this, the resulting private surrogate of rankings are provided to the third party for preference ranking estimation via a regularized M-estimation procedure. Notably, our approach, which adapts to individuals' diverse privacy preferences rather than enforcing a uniform standard, proves valuable for managing the widely acknowledged variability in privacy preferences among individuals \citep{okazaki2009consumer, costante2013privacy}.

Our paper makes the following three methodological and theoretical contributions: 
\begin{itemize}

\item[1] Our first contribution is an adaptive debiased RR mechanism that reconciles the consistency of parameter estimation with utility enhancement in rank aggregation. This method comprises a debiasing step and an adaptive weighting step. The debiasing step aims to align the expected values of privatized rankings with those of the actual rankings, ensuring consistent estimation of $\bm{\theta}^\star$. The adaptive weighting step aims to enhance the robustness of the proposed estimator under varying user privacy preferences.
\item[2] Secondly, we study the asymptotic behavior of estimation errors measured via $\Vert \widehat{\bm{\theta}} - \bm{\theta}^\star \Vert_2$ and $\Vert \widehat{\bm{\theta}} - \bm{\theta}^\star \Vert_{\infty}$ to evaluate the utility of $\widehat{\bm{\theta}}$ based on private surrogate of pairwise rankings. The main technical challenge in this context lies in precisely assessing how diverse privacy guarantees affect $\Vert \widehat{\bm{\theta}} - \bm{\theta}^\star \Vert_2$ and $\Vert \widehat{\bm{\theta}} - \bm{\theta}^\star \Vert_{\infty}$ due to their varying impact on information loss in the raw rankings. Specifically, we show that $m^{-\frac{1}{2}}\Vert \widehat{\bm{\theta}} - \bm{\theta}^\star \Vert_2$ and $\Vert \widehat{\bm{\theta}} - \bm{\theta}^\star \Vert_{\infty}$ both converge to zero with probability at the order $O\left(\sqrt{\frac{\log(m)}{mLpB(\bm{\epsilon})}}\right)$, where $p$ denotes the observation probability of rankings, $m$ and $L$ represent the numbers of items and users, respectively. Here $B(\bm{\epsilon})=\frac{1}{L}\sum_{l=1}^L\left(\frac{e^{\epsilon_l}-1}{e^{\epsilon_l}+1}\right)^2$ represents the averaged privacy preferences among users. Moreover, we establish the minimax rate of the estimation errors based on the private surrogate of pairwise rankings, demonstrating that the proposed estimator is minimax-optimal, up to a logarithmic term. Our theoretical results shed light on the fundamental limit of the privacy-utility tradeoff.
\item[3] Another key issue in rank aggregation involves determining the minimum sample size necessary to accurately recover the true ranking of items. To address this issue, we establish the convergence rates of expected ranking errors in both top-$K$ recovery and full ranking recovery scenarios. Our theoretical findings unveil the sample complexity required to ensure consistent ranking recovery under varying privacy guarantees. 
\end{itemize}

\subsection{Related Work}

There are two closely related lines of work, including the application of differential privacy on ranking data and the privacy-utility tradeoff in statistical estimation problems. Next we provide an overview of relevant studies and discuss their distinctions from our own work.

\noindent
\textbf{Differential Privacy on Ranking Data.} \citet{shang2014application} proposed to utilize the Gaussian noise to contaminate the histogram of collected rankings for rank aggregation, while \cite{hay2017differentially,alabi2022private} employed the Laplace and Gaussian mechanisms to perturb pairwise comparisons. \citet{yan2020private} considered the rank aggregation based on rankings privatized via the randomized response mechanism or the Laplace mechanism. \citet{jeong2022ranking} studied the probability of recovering the individual ranking based on the privatized data and the level of privacy guarantee under different noise-additive mechanisms. \citet{cai2023score} considered rank aggregation under the BTL model by maximizing a perturbed likelihood function to protect individual comparison results. The existing literature primarily focuses on rank aggregation within a non-parametric framework, leaving the fundamental limit in the privacy-utility tradeoff of rank aggregation tasks under the general comparison model with privacy constraints unexplored. 

In our problem, while the classic RR mechanism has been shown to be more effective than noise-additive mechanisms, it inherently causes a shift in the underlying distribution of pairwise comparisons, resulting in significant biases in subsequent rank aggregation tasks. Moreover, the incorporation of user-level privacy in differentially private rank aggregation tasks has been neglected. To bridge these gaps, we approach the rank aggregation task through statistical estimation under the LST models, aiming to offer an optimal solution for online platforms to reconcile the utility of ranking data for third parties with the diverse privacy preferences of users. Additionally, we demonstrate that the proposed method is more efficient in both full ranking recovery and partial ranking recovery compared to the count method that uses comparison results from the RR mechanism through our simulation study and real-world application.


\noindent
\textbf{Fundamental Privacy-Utility Tradeoff.} Another relevant line of research delves into understanding the privacy-utility tradeoff in estimation based on privatized data. \citet{duchi2018minimax} derived minimax bounds for several canonical families of problems under privacy constraints, including mean estimation, median estimation, generalized linear models, and non-parametric density estimation. Their theoretical results shed light on the statistical cost of privacy and characterize the fundamental limit in the statistical utility that a privacy-preserving mechanism can achieve. \citet{cai2021cost} established minimax lower bounds for private mean estimation and linear regression problems. \citet{xu2023binary} considered the binary classification where responses are locally protected via the RR mechanism, quantifying the impact of privacy guarantees on the estimation of the optimal classifier. \citet{chhor2023robust} quantified the impact of differential privacy on the estimation of a discrete distribution. In contrast to previous studies, our focus is on the fundamental limit of statistical estimation within the rank aggregation problem, providing both an upper bound and a minimax lower bound to accurately assess the impact of privacy within the pairwise ranking data framework. Additionally, we consider the privatization of ranking data based on users' privacy preferences, aiming to quantify the overall impact of privacy guarantees on the rank aggregation task and provide insights into achievable privacy levels.

\subsection{Paper Organization}

After introducing necessary notations in Section \ref{Sec:Not}, we introduce the background information on ranking data and a general pairwise comparison model in Section \ref{Sec:Pre}. In Section \ref{Sec:RDP}, we explore the privacy protection aspects of the randomized response mechanism, highlighting the impracticality of the RR mechanism. We introduce an adaptive debiased RR mechanism to address this limitation. Section \ref{Sec:PE} introduces the estimation of true preference parameter under a general comparison model, demonstrating the statistical consistency of the resulting estimator in terms of parameter estimation, top-$K$ ranking recovery, and full ranking recovery. To support our findings, Section \ref{Rec:Sim} includes extensive simulations and a real application. Additional contents and all proofs are deferred to the supplementary file.

\subsection{Notation}
\label{Sec:Not}
For a positive integer $n$, denote $[n]=\{1, ... , n\}$ to be the $n$-set. Given two numbers $a$ and $b$, we use $a \vee b =\max\{a,b\}$ and $a \wedge b = \min\{a,b\}$. For a set $S$, we let $|S|$ denote its cardinality. For two positive sequences $\{f_n\}_{n=1}^{\infty}$ and $\{g_n\}_{n=1}^{\infty}$, we denote that $f_n = O(g_n)$ or $f_n \lesssim g_n$ if $\limsup_{n\rightarrow \infty} f_n/g_n <+\infty$. We let $f_n \asymp g_n$ if $f_n = O(g_n)$ and $g_n = O(f_n)$. A Bernoulli random variable $X$ with parameter $p$ is denoted as $X\sim \text{Bern}(p)$. For a vector $\bm{x}$, we let $\Vert \bm{x}\Vert_2$ denote its $l_2$-norm and $\Vert \bm{x}\Vert_{\infty}$ denote its $l_{\infty}$-norm. For a random variable $X_n$ and a sequence $\{a_n\}_{n=1}^{\infty}$, we denote that $X_n = o_p(a_n)$ is $X_n/a_n$ converges to zero in probability and $X_n = O_p(a_n)$ if $X_n/a_n$ is stochastically bounded.

\section{Preliminaries on Pairwise Ranking Model}
\label{Sec:Pre}

In the realm of ranking data, users engage in comparing a set of items based on users' relative preferences. In practical scenarios, ranking data typically manifests in two forms, contingent on the method of data collection. Primarily, rankings can be derived through pairwise comparisons between items \citep{kalloori2018eliciting}, or by assigning ordinal ranks to the items \citep{szorenyi2015online}. This paper predominantly delves into the context of pairwise rankings, a scenario frequently encountered in political investigations \citep{ackerman2013elections}, recommender systems \citep{kalloori2018eliciting}, and the collection of human feedback for large language models \citep{zhu2023principled}. 

Let $\Omega=\{1,\ldots,m\}$ denote a set of $m$ items. In ranking data, a true preference vector exists, denoted as $\bm{\theta}^\star=(\theta_1^\star,\ldots,\theta_m^\star)^T$. Here $\theta_i^\star>\theta_j^\star$ signifies that item $i$ is ranked higher (preferred) than item $j$ in terms of the ground truth. Pairwise rankings among $\Omega$ can be represented as a set of pairwise comparisons denoted by $\{\bm{y}^{(l)}\}_{l=1}^L$, where $\bm{y}^{(l)}$ represents the set of pairwise comparisons by $l$-th user and can be represented by $m(m-1)/2$ binary values as $\bm{y}^{(l)}= \left(y_{ij}^{(l)}\right)_{i<j} \in \{0,1\}^{m(m-1)/2}$, where $y_{ij}^{(l)}=1$ indicates that item $i$ is observed to be preferred over item $j$ by the $l$-th user. In this paper, we assume that $y_{ij}^{(l)}$'s are independent, as established in prior literature \citep{chen2019spectral, AndersonZhang, AndersonZhang2}. This assumption implies intransitive personal preferences, which is supported by substantial empirical evidence \citep{tversky1969intransitivity, klimenko2015intransitivity}, and also observed in our real application dataset. A more detailed discussion of this phenomenon can be found in the supplementary file.

In practice, the true parameter $\bm{\theta}^\star$ is not observable; instead, noisy pairwise rankings are collected from users. To model these pairwise rankings, a popular class of models is linear stochastic models, which assume that the probability of item $i$ being preferred over item $j$ is 
\begin{align}
\label{General_Model}
\mathbb{P}\left(
y_{ij}^{(l)}=1 
\right) = F(\theta_i^\star-\theta_j^\star),
\end{align}
where $F(\cdot)$ represents the cumulative distribution function (CDF) of a zero-symmetric random variable. In the literature, various types of LST models have been developed, differing in the choice of $F$. In what follows, we provide several examples as documented in previous studies.
\begin{itemize}[itemsep=-1pt]
    \item[1] \textit{Logistic Function}: $F(x)=(1+\exp(-x))^{-1}$ corresponds to the Bradley–Terry-Luce (BTL) model \citep{bradley1952rank,luce2012individual,chen2019spectral};
    \item[2] \textit{Normal CDF}: $F(x)=\int_{-\infty}^x (2\pi)^{-\frac{1}{2}}e^{-\frac{y^2}{2}} dy$ corresponds to the Thurstone-Mosteller (TM) model \citep{thurstone1994law,weng2011bayesian,chen2013pairwise};
    \item[3] \textit{Laplace CDF}: $F(x)=\int_{-\infty}^x \frac{1}{2}e^{-\frac{|y|}{2}}dy$ corresponds to the Dawkins' Threshold (DT) Model \citep{dawkins1969threshold,yellott1977relationship,oliveira2018new}.     
\end{itemize}

A notable characteristic of this class of parametric models outlined in (\ref{General_Model}) is its invariance to adding $\bm{\theta}^\star$ by a constant $c$. In other words, $\bm{\theta}^\star$ and $\bm{\theta}^\star+c \bm{1}_m$ are equivalent for any constant $c$. Therefore, to tackle the issue of identifiability of $\bm{\theta}^\star$, it is standard to impose an additional constraint on $\bm{\theta}^\star$ as $\sum_{i=1}^m \theta_i^\star=0$ \citep{chen2019spectral,liu2023lagrangian}. Furthermore, we suppose that  $\max_{i,j\in[m]} |\theta_i^\star-\theta_j^\star| <\kappa$, where $\kappa$ is the condition number. Compared to the DT model, the BTL and TM models are more popular, as evidenced by their applications in diverse fields such as sports tournaments \citep{karle2023dynamic, oliveira2018new} and social choice theory \citep{shah2018simple}.

\section{Differentially Private Pairwise Rankings}
\label{Sec:RDP}

In this section, we begin by examining the classic randomized response (RR) mechanism. We aim to explore why this conventional approach becomes impractical for our specific problem, demonstrating how the RR mechanism alters the original distribution of collected rankings, rendering inconsistent parameter estimation. To fix this issue, we introduce an adaptive debiased RR mechanism to reconcile privacy protection and utility in rank aggregation.

\subsection{Infeasibility of Classic RR Mechanism}\label{sec:RR}

To ensure the privacy of pairwise comparisons, a common method used in practice is the RR mechanism \citep{warner1965randomized}, which flips a binary value with a predetermined probability. In particular, we denote the randomized response (RR) mechanism with a flipping probability of $p$ as $\mathcal{A}_{p}$. For a pairwise comparison $y_{ij}^{(l)}$, a privacy-preserving ranking $\widetilde{y}_{ij}^{(l)}$ is generated as
\begin{align*}
\widetilde{y}_{ij}^{(l)} = 
	\mathcal{A}_{p}\left(y_{ij}^{(l)}\right) = 
	\begin{cases}
		y_{ij}^{(l)}, &\mbox{ with probability } 1-p, \\
		1-y_{ij}^{(l)},& \mbox{ with probability }  p,
	\end{cases}
\end{align*}
where $p$ is the probability of flipping the observed value. The RR mechanism outputs the true value with probability $1-p$ and the opposite value with probability $p<1/2$.

The privacy guarantee of the RR mechanism can be characterized under the local differential privacy (LDP; \citealt{wang2017locally,duchi2018minimax}). A key advantage of LDP is enabling randomized mechanisms to be implemented at the users' end, accommodating diverse privacy preferences. The formal definition of $\epsilon$-LDP is given as follows. 



\begin{definition}
($\epsilon$-local differential privacy) For any $\epsilon>0$, a randomized mechanism $\mathcal{A}$ is $\epsilon$-local differentially private if for any different input $y$ and $y'$, we have $\max_{\widetilde{y}}
\log\Big|
\frac{\mathbb{P}(\mathcal{A}(y)=\widetilde{y})}{\mathbb{P}(\mathcal{A}(y')=\widetilde{y})} 
\Big|
\leq \epsilon$, where the probability is taken with respect to the randomness of $\mathcal{A}$.
\end{definition}

Denote that $p_{\epsilon}=\frac{1}{e^{\epsilon}+1}$. The RR mechanism $\mathcal{A}_{p_{\epsilon}}$ ensures $\epsilon$-LDP for an observed pairwise comparison. As discussed in \citet{duchi2018minimax}, local DP can be conceptually understood in terms of the disclosure risk associated with the true values of individual data points. Suppose an observed pairwise comparison from user $l$ denoted as $\widetilde{y}_{ij}^{(l)}$, the inference attack is to distinguish two hypotheses regarding the actual value: $y_{ij}^{(l)}=1$ versus $y_{ij}^{(l)}=0$. The difficulty of achieving this distinction is controlled by the privacy parameter $\epsilon$. 

In this paper, we interpret privacy protection on pairwise comparison data as the probability of disclosing the actual comparison value. For example, if $y^{(l)}_{ij} = 1$, then the probabilities of observing $\widetilde{y}_{ij}^{(l)} = 1$ and $\widetilde{y}_{ij}^{(l)} = 0$ are $\frac{e^{\epsilon}}{e^{\epsilon} + 1}$ and $\frac{1}{e^{\epsilon} + 1}$, respectively. This implies that a third party can directly know user $l$ preferring item $i$ over item $j$ with probability $\frac{e^{\epsilon}}{e^{\epsilon} + 1}$. Protecting pairwise comparisons of an individual inherently leads to the protection of their overall preferences. For example, if pairwise comparisons are perturbed with a probability of 0.5 ($\epsilon=0$), the overall preferences of an individual become completely noisy. This connection is thoroughly examined in Section S.2 of the supplementary file.


As shown in Figure \ref{Framework}, as pairwise rankings undergo privatization, the privatized rankings are released to a third party for downstream rank aggregation analysis. Hence it is important to maintain the utility of the downstream analysis while protecting the raw ranking data. However, while it provides privacy protection, the classic RR mechanism alters the distribution of raw pairwise rankings. In Lemma \ref{Lemma:Consis}, we utilize the BTL model as an example to demonstrate that the private rankings released by the RR mechanism no longer follow the original model. Consequently, the true preference parameter vector $\bm{\theta}^\star$ may not be consistently estimated.

\begin{lemma}
\label{Lemma:Consis}
 If $Y_{ij}$ follows the BTL model with parameters $\theta_i^\star$ and $\theta_j^\star$. For any $\epsilon>0$, $\widetilde{Y}_{ij}$ output by the RR mechanism $\mathcal{A}_{p_{\epsilon}}$ follows the distribution: $\widetilde{Y}_{ij}=1$ with probability $\frac{1}{2}+
\frac{e^{\theta_i^\star} - e^{\theta_j^\star}}{e^{\theta_i^\star}+e^{\theta_j^\star}}(\frac{1}{2}-p_{\epsilon})$ and 0 otherwise. Here $\widetilde{Y}_{ij}$ no longer follows the BTL model. 

\end{lemma}

Lemma \ref{Lemma:Consis} demonstrates that $\widetilde{Y}_{ij}$ follows the Bernoulli distribution with a parameter of $\frac{1}{2}+ \frac{e^{\theta_i^\star} - e^{\theta_j^\star}}{e^{\theta_i^\star}+e^{\theta_j^\star}}(\frac{1}{2}-p_{\epsilon})$. As stronger privacy guarantee is imposed on $Y_{ij}$ (i.e., as $\epsilon\rightarrow 0$), the resulting pairwise comparison $\widetilde{Y}_{ij}$ becomes less informative. This is because the RR mechanism implicitly injects noise into pairwise comparisons, shrinking the probability of $\widetilde{Y}_{ij}=1$ towards $1/2$. Since $\widetilde{Y}_{ij}$ no longer follows BTL model, existing methods for estimating the true parameters based on privatized rankings in BTL the model may fail despite the BTL model being the correct model specification. Therefore, as an online platform, releasing $\widetilde{Y}_{ij}$ directly to the third party will lead to inaccurate estimation of the true preference vector $\bm{\theta}^\star$. 

To highlight this point, we provide an example using the BTL model in the following.

\begin{Example}
\label{Example_1}
Consider $\bm{\theta}^\star=(\theta_1^\star,\theta_2^\star,\theta_3^\star)$ satisfying $\theta_1^\star-\theta_2^\star=\theta_2^\star-\theta_3^\star=\log(2)$ and $p_{\epsilon}=1/3$. It can be verified that $\widetilde{Y}_{12}$ and $\widetilde{Y}_{23}$ follow the same Bernoulli distribution with parameter $5/9$, while $\widetilde{Y}_{13}$ follows the Bernoulli distribution with parameter $3/5$. If $\widetilde{Y}_{ij}$'s can be modeled under the BTL model, there should exists $\bm{\theta}'\in \mathbb{R}^3$ such that $A_1/(A_1+1)=5/9$, $A_2/(A_2+1)=3/5$, and $A_3/(A_3+1)=5/9$, where $A_1 = e^{\theta_1'-\theta_2'}$, $A_2 =e^{\theta_1'-\theta_3'}$, and $A_3 = e^{\theta_2'-\theta_3'}$
. However, there is no solution satisfying these equations.
\end{Example}

Example \ref{Example_1} demonstrates the impracticality of sharing privatized pairwise rankings with a third party, who may misinterpret them as being generated from the BTL model. This highlights that RR-based privatized rankings cannot be faithfully modeled by the BTL model, even though it is correct for raw rankings. Hence, releasing the privatized rankings from the classic RR mechanism will lead to significant errors in downstream rank aggregation analysis. Similar phenomena for other parametric models in our framework of (\ref{General_Model}) can be easily derived.

\subsection{Adaptive Debiased RR Mechanism}

To address the issues in the classic RR mechanism, we propose an adaptive debiased RR (ADRR) mechanism to protect pairwise rankings and enhance the utility in rank aggregation. 

In what follows, we utilize an example $\Omega = \{i,j\}$ to motivate the developed method. Specifically, the third party is interested in the preference comparison between items $i$ and $j$. We assume that $y_{ij}^{(l)}$ follows the general parametric pairwise ranking model as outlined in (\ref{General_Model}). After obtaining the privacy-preserving pairwise rankings $\{\widetilde{y}_{ij}^{(l)}\}_{l=1}^L$ output by the classic RR mechanism, given by $\widetilde{y}_{ij}^{(l)} = \mathcal{A}_{p_{\epsilon_l}}(y_{ij}^{(l)})$, where $\epsilon_l$ denotes the privacy preference of user $l$. As previously noted, $\widetilde{y}_{ij}^{(l)}$'s no longer follow the raw model. We consider a debiasing procedure: 
\begin{align}
\label{Eqn:Debias}
\mbox{Debiasing: }\widetilde{z}_{ij}^{(l)} =\frac{(e^{\epsilon_l}+1)\widetilde{y}_{ij}^{(l)}-1}{e^{\epsilon_l}-1}, l \in [L]
\end{align}
Here $\widetilde{z}_{ij}^{(l)}$ is affected by two sources of randomness: the RR mechanism and $Y_{ij}$. 
\begin{lemma}
\label{Lemma:Debias}
Given that $\mathbb{P}(y_{ij}^{(l)}=1)=F(\theta_i^\star-\theta_j^\star)$, the mean and variance of $\widetilde{z}_{ij}^{(l)}$ are given as
\begin{align*}
\mathbb{E}\left(\widetilde{z}_{ij}^{(l)}\right) = F(\theta_i^\star-\theta_j^\star)
\mbox{ and }
\mathrm{Var}\left(\widetilde{z}_{ij}^{(l)}\right) = 
 \frac{1}{4}
\left[
\left(
\frac{e^{\epsilon_l}+1}{e^{\epsilon_l}-1}\right)^2
- 
 \left(
2F(\theta_i^\star-\theta_j^\star)-1
\right)^2\right].
\end{align*}
\end{lemma}
As proved in Lemma \ref{Lemma:Debias}, the debiasing step allows us to derive an unbiased private surrogate ranking for $ y_{ij}^{(l)} $ using $ \widetilde{z}_{ij}^{(l)} $. Consequently, the average $ \frac{1}{L}\sum_{l=1}^L \widetilde{z}_{ij}^{(l)} $ serves as an unbiased and consistent estimator of $ F(\theta_i^\star - \theta_j^\star) $. Importantly, Lemma \ref{Lemma:Debias} also demonstrates that the variances of $ \widetilde{z}_{ij}^{(l)} $ vary significantly due to differing privacy preferences $ \epsilon_l $. When these differences are substantial, the robustness of $ \frac{1}{L}\sum_{l=1}^L \widetilde{z}_{ij}^{(l)} $ for estimating $ F(\theta_i^\star - \theta_j^\star) $ is significantly deteriorated. Specifically, 
as the privacy parameter $ \epsilon_l $ tends towards zero (indicating stronger privacy protection), the variance of $ \widetilde{z}_{ij}^{(l)} $ increases to infinity at the order $ O(\epsilon_l^{-2}) $. To mitigate the impact of $ \epsilon_l $ on the variance of $ \widetilde{z}_{ij}^{(l)} $, it is natural to consider the minimum-variance unbiased estimator \citep{lee2019u}, given as $ \sum_{l=1}^L w_l^\star \widetilde{z}_{ij}^{(l)} $, where $ w_l^\star = \left[\mathrm{Var}(\widetilde{z}_{ij}^{(l)})\right]^{-1} / \sum_{l=1}^L \left[\mathrm{Var}(\widetilde{z}_{ij}^{(l)})\right]^{-1}$. 

In practice, such optimal weights $ w_l^\star$'s are not available since $ F(\theta_i^\star - \theta_j^\star) $ is unknown. However, it is worthy mentioning that the first term of the variance $\left(
\frac{e^{\epsilon_l}+1}{e^{\epsilon_l}-1}\right)^2/4$ dominates the second term $ \left(2F(\theta_i^\star-\theta_j^\star)-1
\right)^2/4$, especially when the privacy parameter $\epsilon_l$ is small. 
For example, if $\epsilon_l=1$ and $F(\theta_i^\star-\theta_j^\star)=0.75$, $\left(\frac{e^{\epsilon_l}+1}{e^{\epsilon_l}-1}\right)^2/\left(
2F(\theta_i^\star-\theta_j^\star)-1
\right)^2 \approx 19$. Motivated from this fact, we can approximate the true variance via the first term $\left(
\frac{e^{\epsilon_l}+1}{e^{\epsilon_l}-1}\right)^2/4$. This leads to our proposed adaptive debiased RR mechanism:
\begin{align}
\label{Eqn:Weight}
\mbox{ Adaptively Debiased RR: }
z_{ij}^{(l)} = w_l \widetilde{z}_{ij}^{(l)} = \frac{\left(\frac{e^{\epsilon_l}-1}{e^{\epsilon_l}+1}\right)^2}{\sum_{l=1}^L \left(\frac{e^{\epsilon_l}-1}{e^{\epsilon_l}+1}\right)^2} \widetilde{z}_{ij}^{(l)}.
\end{align}

The intuition behind the weighting scheme is that users who prioritize stronger privacy protection will have their pairwise rankings perturbed with greater noise by the online platform. However, these noisier comparisons are less informative than those from users opting for weaker privacy protection. To address this, the weighting scheme assigns smaller weights to noisier comparisons during rank aggregation, ultimately improving parameter estimation. In the following sections, we demonstrate that the proposed ADRR mechanism achieves an optimal rate for subsequent rank aggregation and performs effectively across extensive numerical experiments.

\section{Differentially Private Rank Aggregation}
\label{Sec:PE}

In this section, we explore the rank aggregation based on privatized rankings under varying privacy guarantees, establishing theoretical results regarding the consistency regarding parameter estimation and ranking recovery. 

\begin{figure}[h]
    \centering
    \includegraphics[scale=0.25]{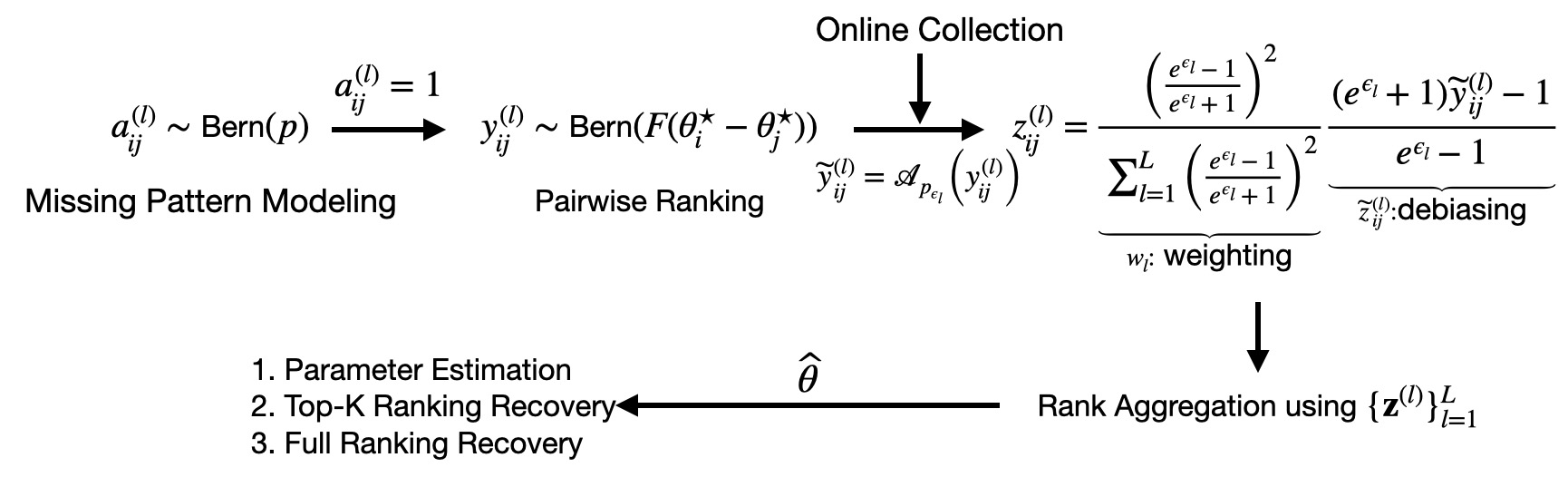}
    \caption{The process of data collection, privacy protection, and rank aggregation.}
    \label{fig:procedure}
\end{figure}

In practice, it is common that not all pairwise comparisons are observed. This is because, in online collection, it is impractical to obtain all pairwise comparisons between items. To address this, our framework allows for some pairwise comparisons to be missing. To characterize the missing pattern, we assume that the missing pattern is characterized by the Erd\H{o}s--R\'enyi random graph. We let $a_{ij}^{(l)} \in \{0, 1\}$ be a Bernoulli random variable with parameter $p$, indicating whether $y_{ij}^{(l)}$ is observed or not. Here $p$ represents the fraction of observed pairwise comparisons. It is typically required that $p \gtrsim \log m/m$ \citep{chen2019spectral}, implying that every item is compared and none remains isolated asymptotically. The general process for rank aggregation is illustrated in Figure \ref{fig:procedure}. In this process, we assume that the raw rankings $y_{ij}^{(l)}$'s follow the general parametric model outlined in (\ref{General_Model}). When the online platform collects pairwise comparisons from users, it permutes these comparisons according to each user’s selected privacy preferences. In other words, users choose their desired level of privacy protection for their responses.




In the subsequent sections, we develop theoretical results pertaining to the utility of rank aggregation tasks across three key aspects illustrated in Figure \ref{fig:procedure}: (1) parameter estimation, (2) top-$K$ ranking recovery, and (3) full ranking recovery. In Section \ref{SecSub:PE}, we develop a differentially private estimator based on collected private rankings and evaluate the overall influence of varying personalized privacy preferences on the convergence rates of the estimation errors, measured by $m^{-\frac{1}{2}}\Vert \widehat{\bm{\theta}}-\bm{\theta}^\star\Vert_2$ and $\Vert \widehat{\bm{\theta}}-\bm{\theta}^\star\Vert_{\infty}$. Furthermore, we establish the minimax rates for the estimation errors. We then analyze the asymptotic behavir of the expected top-$K$ ranking error and the full ranking error of the ranking derived by the proposed estimator in Sections \ref{SecSub:PRR} and \ref{SecSub:FRR}, respectively, where we also quantify the impact of privacy on the convergence rates of these metrics.

\subsection{Parameter Estimation}
\label{SecSub:PE}

We first present theoretical results related to the estimation of the true preference vector $\bm{\theta}^\star$ based on partially observed private pairwise rankings. In the literature on the pairwise comparison model, when there is no privacy guarantee $(\epsilon=\infty)$, the estimation of $\bm{\theta}^\star$ is commonly implemented through regularized maximum likelihood estimation (MLE) \citep{chen2019spectral}. In our framework, the released privatized surrogate rankings $\{\bm{z}^{(l)}\}_{l=1}^L$ are used as surrogates for $\{\bm{y}^{(l)}\}_{l=1}^L$, then we consider the optimization task 
\begin{align}
\label{Equ:OPT1}
\mathcal{L}_{\lambda}(\bm{\theta})=&
-\sum_{l=1}^L 
\sum_{i<j}
a_{ij}^{(l)}
\left\{
z_{ij}^{(l)} \log
F(\theta_i-\theta_j)
+
(w_l-z_{ij}^{(l)})
\log
\left(
1-F(\theta_i-\theta_j)\right)\right\}+\lambda \Vert \bm{\theta}\Vert_2^2 \notag \\
=&\sum_{l=1}^L  \sum_{i<j} \mathcal{L}_{ij}(\bm\theta,\bm{z}^{(l)})
+\lambda \Vert \bm{\theta}\Vert_2^2 \triangleq \mathcal{L}_0(\bm{\theta})+\lambda \Vert \bm{\theta}\Vert_2^2,
\end{align}
where $\lambda>0$ is a regularization parameter. As previously mentioned, the true preference parameter $\bm{\theta}^\star$ is non-unique due to its optimality being invariant to a constant shift. The regularization term ensures the uniqueness of the solution \citep{chen2019spectral,liu2023lagrangian}.

\begin{lemma}
\label{Lemma:Restricted}
For any $\lambda>0$, $\widehat{\bm{\theta}} = \arg\min_{\bm{\theta}} \mathcal{L}_{\lambda}(\bm{\theta})$ satisfies that $\bm{1}_m^T \widehat{\bm{\theta}} = 0$.
\end{lemma}


Lemma \ref{Lemma:Restricted} demonstrates that the regularization term ensures alignment with the true preference parameter within the same constraint $\bm{1}_m^T\widehat{\bm{\theta}} = 0$. In the following, we investigate a range of theoretical properties of \(\widehat{\bm{\theta}}\) to understand the tradeoff between utility and privacy guarantee. The utility of private rankings is quantified by the asymptotic behavior of \(\widehat{\bm{\theta}}\) to \(\bm{\theta}^\star\). Specifically, we primarily examine the following two quantities: \(\Vert \widehat{\bm{\theta}} - \bm{\theta}^\star \Vert_2\) and \(\Vert \widehat{\bm{\theta}} - \bm{\theta}^\star \Vert_{\infty}\). In Section \ref{Sec_sub:BTL}, we begin by establishing the theoretical results under the BTL model, which are later generalized for a general model \(F(\cdot)\) under mild assumptions in Section \ref{Sec_sub:general}.

\subsubsection{BTL model}
\label{Sec_sub:BTL}

In this section, we establish the consistency of parameter estimation under the BTL model as a special case. It is worth noting that under the BTL model, the optimization task in (\ref{Equ:OPT1}) is always strictly convex regardless of the value of $\bm{\epsilon}$, allowing for the existence of the optimal minimizer $\widehat{\bm{\theta}}$. Consequently, $\widehat{\bm{\theta}}$ can be simply obtained by applying gradient descent. In the following theorem, we present the efficiency of $\widehat{\bm{\theta}}$ under the BTL model within the framework in Figure \ref{fig:procedure}, quantifying the impact of privacy protection on the estimation errors.

\begin{theorem}
\label{Thm:Estimation}
Let $\widehat{\bm{\theta}}$ be the minimizer of (\ref{Equ:OPT1}) with $\lambda \lesssim \frac{\log(m)}{LB(\bm{\epsilon})}$ and suppose $\frac{\log(m)}{mLpB(\bm{\epsilon})}=o(1)$. Under the BTL model, we have the following results concerning the convergence of $\widehat{\bm{\theta}}$ to $\bm{\theta}^\star$:
 \begin{align}
 \label{Rate:BTL}
   \frac{1}{\sqrt{m}}\|\widehat{\bm\theta}-\bm\theta^\star\|_2  \lesssim  \sqrt{\frac{\log(m)}{mLp B(\bm{\epsilon})}} 
   \mbox{ and }
   \|\widehat{\bm\theta}-\bm\theta^\star\|_{\infty}\lesssim 
    \sqrt{\frac{\log(m)}{mLpB(\bm{\epsilon})}},
 \end{align}
 with probability at least $1-O(m^{-3})$, where $B(\bm{\epsilon})=\frac{1}{L}\sum_{l=1}^L \left(\frac{e^{\epsilon_l}-1}{e^{\epsilon_l}+1} \right)^2$.
\end{theorem}

Theorem \ref{Thm:Estimation} provides theoretical results regarding the asymptotic behavior of the estimation errors of $\widehat{\bm{\theta}}$ concerning its convergence toward $\bm{\theta}^\star$ under the BTL model. In this context, $mLp$ can be interpreted as the effective sample size for estimation. As the observation probability $p$ increases, a faster convergence rate for parameter estimation is achieved. It is evident that increasing the number of users ($L$) or items ($m$) results in a more accurate estimation of the true preference parameter vector $\bm{\theta}^\star$. Importantly, when privacy constraints are imposed on pairwise comparisons, the convergence rate of $\widehat{\bm{\theta}}$ is notably slowed by a multiplicative term of $1/\sqrt{B(\bm{\epsilon})}$, where $B(\bm{\epsilon})$ can be viewed as the averaged privacy preferences of the users.

A particular theoretical insight from Theorem \ref{Thm:Estimation} offers a quantification of how the average privacy protection of users impacts the convergence of $\widehat{\bm{\theta}}$. This result is significant in understanding the fundamental limit of privacy protection that can be achieved while maintaining consistency of $\widehat{\bm{\theta}}$ in estimating $\bm{\theta}^\star$. Additionally, this convergence rate in (\ref{Rate:BTL}) matches the minimax lower bounds for a general comparison model $F(\cdot)$ in Theorem \ref{Thm_1-Minimax}, demonstrating the optimality of the proposed estimator. In the absence of privacy constraints (as $\epsilon_l$ approaches infinity), our results reduce to the existing non-private one $O\left(\sqrt{\frac{1}{mLp}}\right)$, omitting the logarithmic term  \citep{chen2019spectral}.

\subsubsection{General Pairwise Comparison Model}
\label{Sec_sub:general}

In this section, we extend the results in Theorem \ref{Thm:Estimation} to encompass a broader range of $F(x)$. We let $f(x)$ be the derivative of $F(x)$ and define $g(x) = \frac{f(x)}{F(x)}$. Subsequently, we impose several mild assumptions on $F$.
\begin{assumption}[Log-concavity]
\label{Ass1}
 The cumulative distribution function $F(x)$ is log-concave.
\end{assumption}

\begin{assumption}[Boundedness]
\label{Ass2}
Assume that both $F(x)$ and $f(x)$ are bounded away from zero on any closed interval in $\mathbb{R}$.
\end{assumption}

Assumption \ref{Ass1} states that $F(x)$ is log-concave, and Assumption \ref{Ass2} assumes that both $F(x)$ and $f(x)$ are lower bounded by positive constants when $x$ is within a bounded range. These mild assumptions are satisfied in the BTL, TM, and DT models introduced in Section \ref{Sec:Pre}.

For a general $F(x)$ satisfying Assumptions \ref{Ass1} and \ref{Ass2}, the optimization problem in (\ref{Equ:OPT1}) may not be convex due to the inclusion of privacy protection. This is because $z_{ij}^{(l)}$ can take negative values, thereby potentially disrupting the convexity. However, we establish the convexity of its population-level counterpart in Lemma \ref{Lemma:Restricted}, implying the asymptotic convexity of $\mathcal{L}_{\lambda}(\bm \theta)$. To substantiate this claim, we present a tail probability bound for the smallest non-zero eigenvalue of $\nabla^2\mathcal{L}_{0}(\bm{\theta})$ in Lemma \ref{Lemma:Hessian}, demonstrating that the probability of $\mathcal{L}_{0}(\bm{\theta})$ being non-convex is exponentially small.

We define $\bm{\Theta}=\{\bm{\theta} \in \mathbb{R}^m: \bm{1}_m^T \bm{\theta}=0,\max_{i\neq j}|\theta_i-\theta_j| \leq \kappa\}$. Let $C_{g',L}$ and $C_{g',U}$ denote the minimum and maximum of $-g'(x)$ on $[-\kappa,\kappa]$, respectively. These constants $C_{g',L}$ and $C_{g',U}$ are positive due to the log-concavity of $F$, and they depend on the choice of $F(\cdot)$. 

\begin{lemma}
    \label{Lemma:Hessian}
    $\mathbb{E}\left(\mathcal{L}_{\lambda}(\bm{\theta})\right)$ is strictly convex with respect to $\bm{\theta}$. For any $\bm{\theta} \in \mathcal{C}(q)=\{\bm{\theta}: \Vert \bm{\theta} - \bm{\theta}^\star\Vert_{\infty} \leq q \}$, it holds true that 
    \begin{align}
    \label{Eqn:Conv}
\mathbb{P}
    \left(
\Lambda_{min,\perp}
\left(
\nabla^2\mathcal{L}_{0}(\bm{\theta})
\right)>\frac{C_{g',L}(q)mp}{4}
    \right)
    \geq 1-
    m \exp\left(-\frac{3C_{g',L}^2(q)mLpB(\bm{\epsilon})}{128  C_{g',U}^2(q) }\right),
    \end{align}
    where $\Lambda_{min,\perp}(\cdot)$ denotes the smallest non-zero eigenvalue.
\end{lemma}

\begin{figure}[ht!]
        \centering
        \begin{subfigure}[b]{0.45\textwidth}  
            \centering 
            \includegraphics[width=\textwidth]{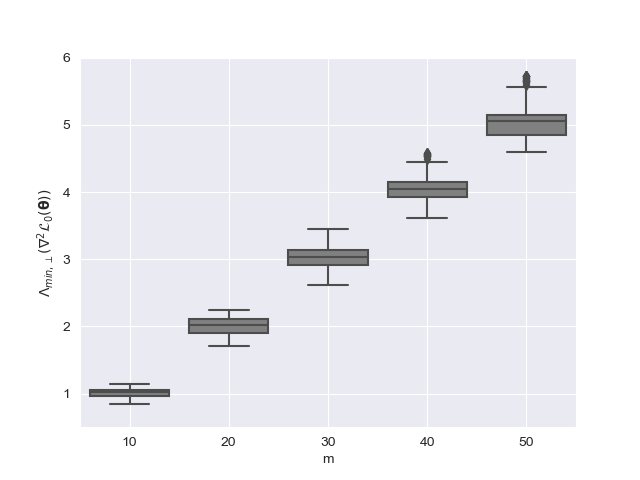}
        \end{subfigure}
                        \begin{subfigure}[b]{0.45\textwidth}  
            \centering 
            \includegraphics[width=\textwidth]{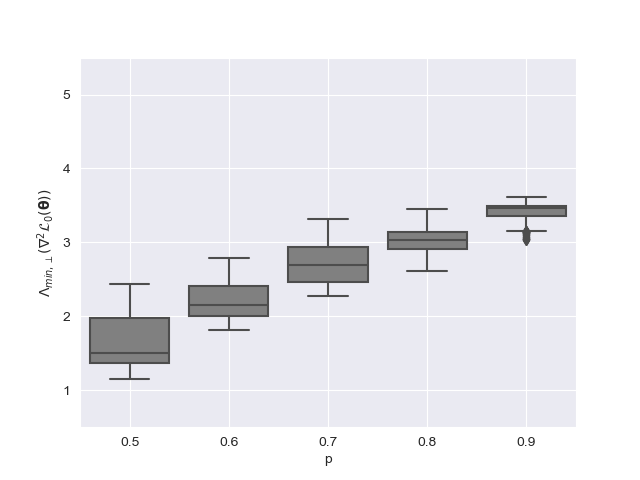}
        \end{subfigure}
        \caption{The boxplots of the smallest non-zero eigenvalues of cases $(L,p,\epsilon)=(30,0.8,2)$ (Left) and $(m,L,\epsilon)=(30,30,2)$ (Right) under the TM model with $\bm{\theta}$ being equally spaced in $[0, 1]$, showing that $\Lambda_{min,\perp}\left(\nabla^2\mathcal{L}_{0}(\bm\theta)\right)$ exhibits a linear relationship with $m$ and $p$. }
        \label{fig:min-eigenvalue}
    \end{figure}

Lemma \ref{Lemma:Hessian} demonstrates that the expectation of $\mathcal{L}_{\lambda}(\bm{\theta})$ is strongly convex for any $\lambda > 0$. Moreover, the smallest non-zero eigenvalue of $\nabla^2 \mathcal{L}_0(\bm{\theta})$ is lower bounded by $O(mp)$ with a probability approaching 1 as either $m$ or $L$ increases, see Figure \ref{fig:min-eigenvalue} for an illustration. A key insight from \eqref{Eqn:Conv} is that larger values of $m$, $L$, and $p$ suggest a higher probability of $\nabla^2\mathcal{L}_{0}(\bm{\theta})$ being convex. Specifically, convexity holds asymptotically when $mLpB(\bm{\epsilon}) \rightarrow \infty$.

\begin{theorem}
\label{Thm:Estimation_General}
Suppose $\frac{\log(m)}{mLpB(\bm{\epsilon})}=o(1)$. For a general $F(x)$ under Assumptions \ref{Ass1}-\ref{Ass2}, we have
 \begin{align*}
   \frac{1}{\sqrt{m}}\|\widehat{\bm\theta}-\bm\theta^\star\|_2  \lesssim   
\sqrt{ \frac{\log(m)}{mLpB(\bm{\epsilon})}}
   \mbox{ and }
   \|\widehat{\bm\theta}-\bm\theta^\star\|_{\infty}\lesssim 
    \sqrt{\frac{\log(m)}{mLpB(\bm{\epsilon})}},
 \end{align*}
 with probability at least $1-O(m^{-3})$, where constants depending on $F(\cdot)$ are omitted.
 
\end{theorem}

Theorem \ref{Thm:Estimation_General} extends the theoretical results of Theorem \ref{Thm:Estimation} regarding the asymptotic behavior of the estimation errors of $\widehat{\bm{\theta}}$ to a general $F(x)$. Clearly, the asymptotic behavior of the estimation errors in terms of $m$, $L$, and $\bm{\epsilon}$ remains invariant to $F(\cdot)$, up to some multiplicative constants. Additionally, the impact of privacy guarantee is also invariant to $F(\cdot)$. For any $F$, the convergence rate of $\widehat{\bm{\theta}}$ is always slowed by a multiplicative constant of $1/\sqrt{B(\bm{\epsilon})}$.


To gain deeper insights into the privacy-utility tradeoff, we establish the minimax rates of the estimation errors under a general $F(x)$ based on privatized rankings in Theorem \ref{Thm_1-Minimax}. 
\begin{theorem}
Let $\widetilde{\mathcal{D}}=\{\bm{z}^{(l)}\}_{l=1}^L$ denote a set of privatized pairwise rankings under a general $F(\cdot)$. For any estimator $\widehat{\bm\theta}$ based on $\widetilde{\mathcal{D}}$, we have
\label{Thm_1-Minimax}
\begin{align}
\inf_{\widehat{\bm{\theta}}}
    \sup_{\bm{\theta}^\star \in \bm{\Theta}}
\mathbb{E}
\left[
\Vert \widehat{\bm{\theta}} - \bm{\theta}^\star \Vert_{\infty}
\right] \geq 
\inf_{\widehat{\bm{\theta}}}
    \sup_{\bm{\theta}^\star \in \bm{\Theta}}
\mathbb{E}
\left[
\frac{1}{\sqrt{m}}
\Vert  \widehat{\bm{\theta}} - \bm{\theta}^\star  \Vert_2
\right] \gtrsim 
\sqrt{
\frac{1}{m L p B(\bm{\epsilon}) }},
\end{align}
where constants depending on $F(\cdot)$ are omitted.
\end{theorem}

Theorem \ref{Thm_1-Minimax} introduces the minimax rates of the estimation errors. The minimax rate aligns with the upper bound on the convergence rate of $\widehat{\bm{\theta}}$ up to a logarithmic term, showing that the estimator $\widehat{\bm{\theta}}$ derived from the proposed estimator is minimax optimal, achieving the best possible performance while allowing for a logarithmic term. This result showcases the efficiency and effectiveness of the proposed estimator $\widehat{\bm{\theta}}$ under privacy protection constraints. Additionally, a main characteristic of our minimax rate is considering varying privacy concerns among users, which remains largely unexplored in the current literature regarding estimation based on differentially private data.

\subsection{Top-$K$ Ranking Recovery}
\label{SecSub:PRR}

In addition to the parameter estimation, another challenge in the domain of ranking is obtaining the list of top-$K$ items \citep{chen2015spectral,suh2017adversarial}. For example, in recommender systems, only products placed in the first few positions are readily accessible to customer, and hence recommender systems usually care about whether they place the $K$ most preferred items for each customer. This endeavor inherently revolves around distinguishing between the $K$-th and $(K+1)$-th most preferred items. In this section, our focus is on establishing the convergence rate of the ranking error associated with $\widehat{\bm{\theta}}$ when selecting the top-$K$ items. In comparison to parameter estimation, achieving an accurate ranking order is a more manageable task, as obtaining the correct ordering of items does not necessarily imply consistency in parameter estimation.

Specifically, let $\theta_{(k)}^\star$ represent the preference parameter of the $k$-th most preferred item. The difficulty of this challenge inherently hinges on the degree of separation between $\theta_{(k)}^\star$ and $\theta_{(k+1)}^\star$. In other words, when $\theta_{(k)}^\star \gg \theta_{(k+1)}^\star$, the task of identifying the top-$K$ items becomes considerably easy. In the following Theorem, we shed light on the sample complexity required for recovering the exact top-$K$ identification.

Next, we aim to analyze the asymptotic behavior of the error about selecting the top-$K$ items. To this end, we define the function $\sigma(\bm{\theta}) = (\sigma(\theta_i))_{i \in [m]}$, where $\sigma(\theta_i)$ represents the rank of $\theta_i$ within the values of $\bm{\theta}$. To illustrate, if $\sigma(\theta_i) = k$, it signifies that item $i$ is the $k$-th highest value among $\theta_i$'s. Then we consider the normalized Hamming distance between $\widehat{\bm{\theta}}$ and $\bm{\theta}^\star$ \citep{shah2018simple,AndersonZhang}, which is defined as
\begin{align*}
H_K(\widehat{\bm{\theta}},\bm{\theta}^\star)=
\frac{1}{2K}
\left(
\sum_{i=1}^m
I\left(\sigma(\widehat{\theta}_i)>K, 
\sigma(\theta_i^\star) \leq K \right)
 + 
\sum_{i=1}^m
I\left(\sigma(\widehat{\theta}_i) \leq K,
\sigma(\theta_i^\star) > K
\right)
\right).
\end{align*}
Here $H_K(\widehat{\bm{\theta}},\bm{\theta}^\star)$ takes values in $\{0,\frac{1}{K},\frac{2}{K},\ldots,1\}$ and $H_K(\widehat{\bm{\theta}},\bm{\theta}^\star)=0$ only when $\{i\in [m]:\sigma(\widehat{\theta}_i) \leq K\} = \{i\in [m]:\sigma(\theta_i^\star) \leq K \}$. It is worth noting that achieving a top-$K$ ranking error of zero does not require the consistency of $\widehat{\bm{\theta}}$. It merely necessitates alignment between $\widehat{\bm{\theta}}$ and $\bm{\theta}^\star$ in distinguishing between the Top-$K$ items and the others. In the following lemma, we establish an upper bound for $H_K(\widehat{\bm{\theta}},\bm{\theta}^\star)$, building upon the probabilistic behavior of $\widehat{\bm{\theta}}$.

\begin{theorem}
\label{Thm: partial}
Under the assumptions of Theorem \ref{Thm:Estimation_General}, the top-$K$ ranking error of $\widehat{\bm{\theta}}$ satisfies
$
H_K(\widehat{\bm{\theta}},\bm{\theta}^\star)
= O_p\left(
\exp\left(
- C_0 mLpB(\bm{\epsilon})\Delta_K^2
\right)\right)$, for some positive constants $C_0$ depending on $F(x)$. 
\end{theorem}

Theorem \ref{Thm: partial} reveals that the expected ranking error exhibits exponential convergence under privacy constraints. Notably, an increase in the number of rankings or items augments the precision of identifying the top-$K$ item set for any fixed value of $\bm{\epsilon}$. Furthermore, Theorem \ref{Thm: partial} delivers a similar insight into the influence of privacy guarantees.

\subsection{Full Ranking Recovery}
\label{SecSub:FRR}

Acquiring a full ranking of items constitutes a fundamental challenge within the domains of ranking and preference modeling. Obtaining the full ranking of items based on user preferences is instrumental in enhancing the efficiency of a recommender system, enabling the system to place items according to the established preference order. Compared with the top-$K$ ranking recovery problem, the difficulty of obtaining the accurate full ranking depends on the differences between $\theta_{(i)}^\star$ and $\theta_{(i+1)}^\star$ for all $i \in [m]$. In this section, we delve into the theoretical aspects of the asymptotic behavior of full ranking error, with the aim of quantifying the impact of privacy guarantee on the error of recovering the full ranking. To this end, we consider the normalized versions of Kendall's tau distance and Spearman’s footrule \citep{diaconis1977spearman}, as defined by
\begin{align*}&
K(\widehat{\bm{\theta}},\bm{\theta}^\star)=
\frac{2}{m(m-1)}
\sum_{1 \leq i <j \leq m}
I\left(
\big(\sigma(\widehat{\theta}_i)-\sigma(\widehat{\theta}_j)\big)
\big(\sigma(\theta_i^\star)-\sigma(\theta_j^\star)\big)
<0
\right), \\
&S(\widehat{\bm{\theta}},\bm{\theta}^\star) = \frac{2}{m^2}
\sum_{i=1}^m |\sigma(\widehat{\theta}_i)-\sigma(\theta_i^\star)|,
\end{align*}
respectively. Here, $K(\widehat{\bm{\theta}},\bm{\theta}^\star)$ measures the percentage of pairs of items that are ranked inaccurately between the rankings induced from $\widehat{\bm{\theta}}$ and $\bm{\theta}^\star$ and $S(\widehat{\bm{\theta}},\bm{\theta}^\star)$ calculates the averaged absolute difference between two rankings. Additionally, using the Diaconis-Graham inequality \citep{diaconis1977spearman}, we can show that $S(\widehat{\bm{\theta}},\bm{\theta}^\star) \asymp K(\widehat{\bm{\theta}},\bm{\theta}^\star)$. This relationship shows that Kendall’s tau provides a tight bound for Spearman’s footrule, allowing us to derive the asymptotic behavior of $S(\widehat{\bm{\theta}},\bm{\theta}^\star)$ based on that of $K(\widehat{\bm{\theta}},\bm{\theta}^\star)$.

\begin{theorem}
Under the assumptions of Theorem \ref{Thm:Estimation_General}, the full ranking errors of $\widehat{\bm{\theta}}$ satisfies
\label{Thm:FRE}
\begin{align}
\label{Rate:FR}
S(\widehat{\bm{\theta}},\bm{\theta}^\star)
\leq  2
K(\widehat{\bm{\theta}},\bm{\theta}^\star)
= O_p \left(
\frac{1}{m-1}
\sum_{k=1}^{m-1}
\exp\left(
- C_0mLpB(\bm{\epsilon})\Delta_k^2
\right)
\right)
,
\end{align}
for some positive constants $C_0$ depending on $F(x)$, where $\Delta_k  = \theta_{(k)}^\star - \theta_{(k+1)}^\star$.
\end{theorem}

Theorem \ref{Thm:FRE} presents the convergence rate of the expected full ranking error, demonstrating that, similar to the top-$K$ ranking error, the full ranking error decreases exponentially as the number of users $(L)$ and items $(m)$ increases. A noteworthy distinction lies in the influence of all adjacent true parameters, specifically $\theta_{(k)}^\star-\theta_{(k+1)}^\star$ for $k \in [m-1]$, on the rate presented in (\ref{Rate:FR}). This intriguing phenomenon can be explained by the fact that achieving a precise ranking of all items is inherently equivalent to identifying all top-$K$ item sets.

\section{Numerical Experiments}
\label{Rec:Sim}

In this section, we undertake a series of numerical experiments using both simulated data and the Car preference dataset to substantiate our theoretical conclusions. Particularly, our experiments focus on demonstrating the efficacy of the proposed ADRR mechanism on two aspects: parameter estimation and ranking recovery. In the main paper, we focus on plot results. Detailed experimental results, including tables of average estimation errors, ranking errors, and their standard errors, are provided in Section S.3 of the Appendix.

\subsection{Simulation}

The simulated datasets are created via the following procedure. Firstly, we generate the true preference parameter vector $\bm{\theta}^\star = (\theta_1^\star, \ldots, \theta_m^\star)$ from a uniform distribution, specifically, $\theta_i^\star \sim \text{Uniform}(-1, 1)$. Secondly, we generate pairwise comparisons as outlined in the framework in Figure \ref{fig:procedure}. We then repeat this second step $L$ times to generate $L$ individual rankings and let $p$ represent the observation probability. For each experimental setting, the regularization parameter $\lambda$ is set as $\lambda \asymp (LB(\bm{\epsilon}))^{-1}$ in accordance with our theoretical framework. For illustration, we conduct all experiments using both the BTL model and the TM model.

\noindent
\textbf{Scenario I: Consistency of Parameter Estimation.} In the first scenario, our objective is to provide empirical validations of the convergence of $\widehat{\bm{\theta}}$ to $\bm{\theta}^\star$ when $m$ or $L$ diverges. To this end, we consider cases $(L,m)\in \{100,200,300,400\} \times \{10,20,30\}$ with $\epsilon_l$'s being randomly generated from $\text{Uniform}(1,5)$ and $p=0.5$.  

\begin{figure}[h!]
        \centering
        \begin{subfigure}[b]{0.45\textwidth}   
            \centering 
            \includegraphics[width=\textwidth]{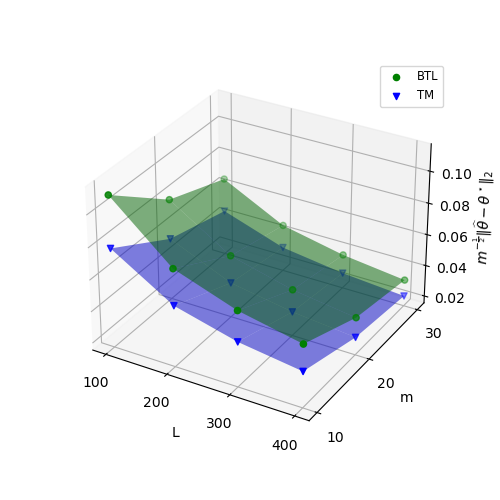}
        \end{subfigure}
                \begin{subfigure}[b]{0.45\textwidth}  
            \centering 
            \includegraphics[width=\textwidth]{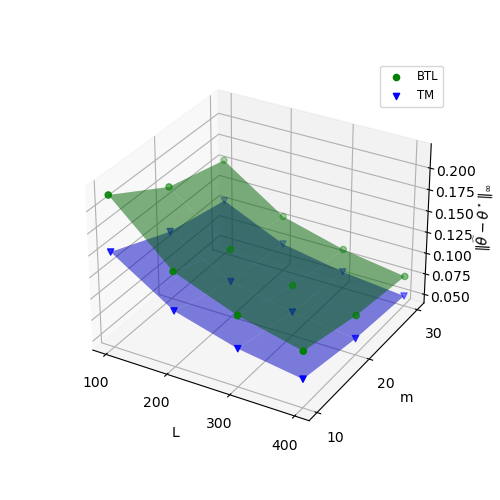}
        \end{subfigure}

        \caption{The averaged estimation errors $m^{-\frac{1}{2}} \Vert \widehat{\bm{\theta}} - \bm{\theta}^\star \Vert_2$ (Left) and $\Vert \widehat{\bm{\theta}} - \bm{\theta}^\star \Vert_\infty$ (Right) of all cases under both the BTL and TM models in Scenario I.}
        \label{fig:PG_S1}
    \end{figure}

In Figure \ref{fig:PG_S1}, the averaged estimation errors in 200 replications, measured as $m^{-\frac{1}{2}} \Vert \widehat{\bm{\theta}} - \bm{\theta}^\star \Vert_2$ and $\Vert \widehat{\bm{\theta}}-\bm{\theta}^\star \Vert_{\infty}$, are reported. As depicted in Figure \ref{fig:PG_S1}, the estimator $\widehat{\bm{\theta}}$ converges to $\bm{\theta}^\star$ as $L$ or $m$ increases. This observation highlights that the proposed estimator derived from the proposed weighted regularized MLE is a consistent estimator of $\bm{\theta}^\star$, which aligns with our theoretical findings outlined in Theorem \ref{Thm:Estimation_General}.

\begin{figure}[ht!]
        \centering
        \begin{subfigure}[b]{0.45\textwidth}  
            \centering 
            \includegraphics[width=\textwidth]{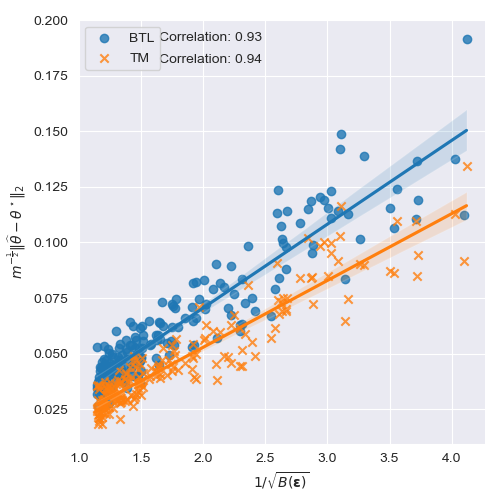}   
        \end{subfigure}
                \begin{subfigure}[b]{0.45\textwidth}  
            \centering 
            \includegraphics[width=\textwidth]{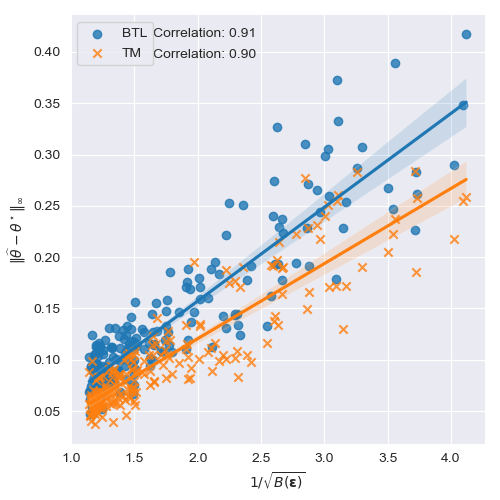}  
        \end{subfigure}
        \caption{The linear models between estimation errors and $\frac{1}{\sqrt{B(\bm{\epsilon})}}$ in Scenario II. Here, the dots represent the estimation errors of 200 replicates, and the shaded areas represent the corresponding 95\% prediction intervals.}
        \label{fig:S2}
    \end{figure}

\noindent
\textbf{Scenario II: Effect of Varying Privacy Guarantee.} In the second scenario, our objective is to empirically confirm the theoretical findings presented in Theorems \ref{Thm:Estimation_General} and \ref{Thm_1-Minimax}, which assert that the privacy guarantee influences estimation errors through a multiplicative constant of $\sqrt{1/B(\bm{\epsilon})}$. To substantiate this assertion, it is sufficient to examine whether the estimation errors $\Vert \widehat{\bm{\theta}} - \bm{\theta}^\star \Vert_\infty$ and $m^{-\frac{1}{2}}\Vert \widehat{\bm{\theta}} - \bm{\theta}^\star \Vert_2$ demonstrate a linear correlation with $\sqrt{1/B(\bm{\epsilon})}$. To this end, we set $(L,m,p)$ to $(200,20,1)$ and generate $\bm{\epsilon}=(\epsilon_l)_{l \in [m]}$ from a uniform distribution $\text{Uniform}(A, A+0.5)$, where $A$ is randomly drawn from the interval $[0.5, 2.5]$. Here the parameter $A$ governs the value of $\sqrt{1/B(\bm{\epsilon})}$. We conduct this experiment 200 times, fitting linear models between $\sqrt{1/B(\bm{\epsilon})}$ and estimation errors for both BTL and TM models. Additionally, we display the corresponding correlation coefficients in Figure \ref{fig:S2}. 

Clearly, both the BTL and TM models demonstrate a linear relationship between the privacy constant $\sqrt{1/B(\bm{\epsilon})}$ and the estimation errors $\Vert \widehat{\bm{\theta}} - \bm{\theta}^\star \Vert_\infty$ and $m^{-\frac{1}{2}}\Vert \widehat{\bm{\theta}} - \bm{\theta}^\star \Vert_2$. Particularly, the linear correlation coefficients are all greater than 0.9. This finding corroborates the theoretical results outlined in Theorems \ref{Thm:Estimation_General} and \ref{Thm_1-Minimax}.

\noindent
\textbf{Scenario III: Fundamental Privacy-Utility Tradeoff.} In this scenario, we aim to verify our theoretical results that there exists an dividing line for the order of privacy guarantee determining the convergence of $\widehat{\bm{\theta}}$ to $\bm{\theta}^\star$ when $\epsilon$ decreases with $m$ and $L$. To this end, we consider cases $(L,m)\in \{100,200,400,800\} \times \{10,20,30\}$ with identical privacy preferences among users under three data-dependent privacy schemes: (1) $\epsilon_1(m,L) \asymp \frac{\log^2(mL)}{\sqrt{mL}}$; (2) $\epsilon_2(m,L) \asymp \frac{1}{\sqrt{mL}}$; (3) $\epsilon_3(m,L) \asymp \frac{1}{\log^2(mL)\sqrt{mL}}$.

\begin{figure}[!ht]
        \centering
                \begin{subfigure}[b]{0.45\textwidth}  
            \centering 
            \includegraphics[width=\textwidth]{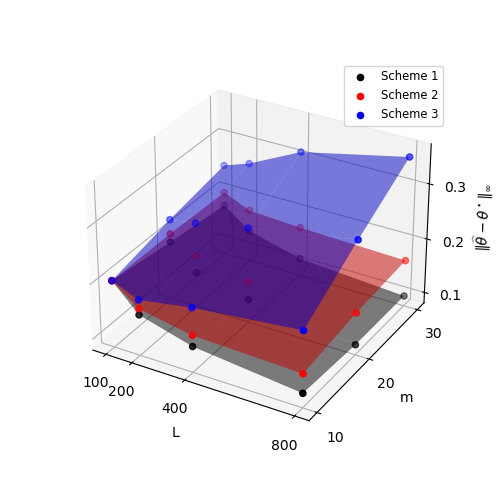}  
        \end{subfigure}
                        \begin{subfigure}[b]{0.45\textwidth}  
            \centering 
            \includegraphics[width=\textwidth]{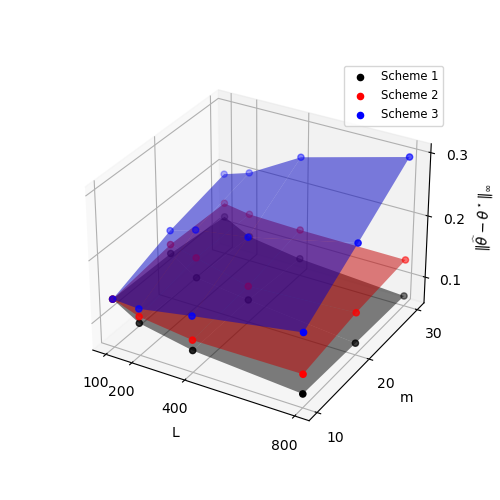}  
        \end{subfigure}
        \caption{The averaged estimation errors measured by $\Vert \widehat{\bm{\theta}} - \bm{\theta}^* \Vert_{\infty}$ in three privacy adaptive schemes in Scenario III under the BTL model (Left) and the TM model (Right).}
        \label{fig:Simu_S3}
    \end{figure}

The estimation error of $\widehat{\bm{\theta}}$ is expected to display distinct patterns as $L$ or $m$ increases under three different schemes. Notably, the convergence of $\widehat{\bm{\theta}}$ towards the true parameter vector $\bm{\theta}^\star$ is not assured under Schemes 2 and 3, which becomes evident when we substitute $\epsilon_{2}(m, L)$ and $\epsilon_{3}(m, L)$ into the minimax lower bound as established in Theorem \ref{Thm_1-Minimax}. Under Scheme 1, we can guarantee the convergence of $\widehat{\bm{\theta}}$, which is supported by the upper bound presented in Theorem \ref{Thm:Estimation_General}. This phenomenon arises from the delicate balance between the increase in information gain resulting from a higher number of rankings and the information loss due to enhanced privacy guarantee. Notably, the introduction of an additional logarithmic term exerts a substantial impact on the original curve pattern. More specifically, for Scheme 3, the estimation errors of $\widehat{\bm{\theta}}$ worsen considerably as $L$ or $m$ increases, while in the case of Scheme 1, there is a significant improvement.


\noindent
\textbf{Scenario IV: Larger Separation Facilitates Ranking Recovery.} In this scenario, our objective is to demonstrate that ranking recovery becomes more achievable as the parameter gap $\Delta_i=\theta_{(i)}^\star - \theta_{(i+1)}^\star$ increases when $(m,L)$ is fixed. We assume that the true preference parameters are evenly spaced, such that $\Delta_i = \Delta$ for $i \in [m-1]$. We examine cases where $\Delta \in \{0.05,0.1,0.15\}$, and consider the pairs $(L, m) \in \{100,200,300\} \times \{10,20,30\}$ with $\bm{\epsilon}$ being generated as described in Scenario I and $p = 0.5$. Results for the TM model are included in the Appendix.

\begin{figure}[h!]
        \centering
        \begin{subfigure}[b]{0.45\textwidth}  
            \centering 
            \includegraphics[width=\textwidth]{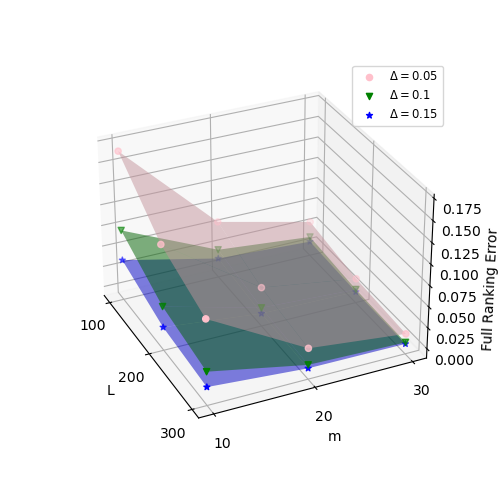}
 
        \end{subfigure}
        \begin{subfigure}[b]{0.45\textwidth}   
            \centering 
            \includegraphics[width=\textwidth]{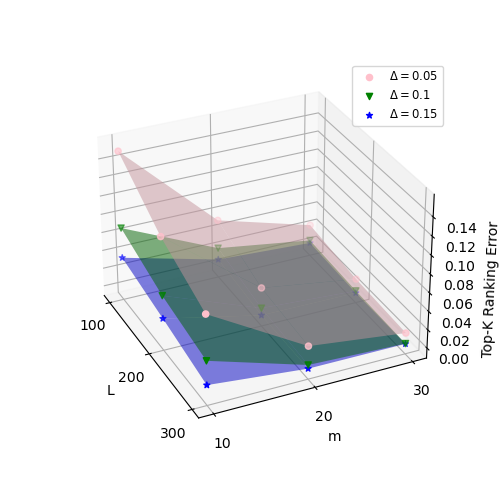}   
        \end{subfigure}

        \caption{The averaged full ranking error (Left) and top-$K$ ranking error (Right) of all cases in Scenario IV under the BTL model.}
        \label{fig:PG4}
    \end{figure}

In Figure \ref{fig:PG4}, we present the averaged top-$K$ ranking errors and full ranking errors across all cases under the BTL model. We set $K$ to be $m/2$ and replicate each case 200 times. As depicted in Figure \ref{fig:PG4}, we observe a clear trend of diminishing ranking errors as $\Delta$ increases. Comparing this trend to parameter estimation, it becomes evident that achieving a low ranking error is more attainable. Notably, when $(L, m, \Delta) = (300, 30, 0.15)$, the top-$K$ ranking error is zero. This result is consistent with our theories outlined in Theorems \ref{Thm: partial} and \ref{Thm:FRE} that ranking recovery is an easier task compared with parameter estimation.

\noindent
\textbf{Scenario V: Comparison to Competitive Methods.} We evaluate our proposed ADRR mechanism against three competitive methods for parameter estimation. Specifically, we compare it with the classic RR method discussed in Section \ref{sec:RR} and the Laplace mechanism. The Laplace mechanism injects Laplace noise into pairwise comparisons to achieve $\epsilon$-LDP, i.e., $\widetilde{y}_{ij}^{(l)}=y_{ij}^{(l)}+\xi_l$ with $\xi_l$ following the Laplace distribution with mean zero and scale $1/\epsilon_l$. Additionally, we include the objective perturbation method \citep{cai2023score}, denoted as \textit{NoisyObj}, for comparison. This method protects pairwise comparisons by perturbing the objective function with Gaussian noise and achieves $(\epsilon,\delta)$-differential privacy. There are two key differences between the objective perturbation method and the others. First, the objective perturbation method provides identical privacy protection for each comparison value. Second, in the context of central DP, the objective perturbation method benefits from injecting less noise into the comparison data, leading to better rank aggregation results. Therefore, we set $(\epsilon,\delta)=(\min \epsilon_l,10^{-6})$ to meet the privacy preferences of all users. 

 For ranking recovery, we compare our method with the nonparametric count method under privacy constraints \citep{yan2020private,alabi2022private}, where $\widetilde{S}_i = \sum_{j \in [m]\setminus \{i\}}
\sum_{l=1}^L a_{ij}^{(l)} \widetilde{y}_{ij}^{(l)}$ with $\widetilde{y}_{ij}^{(l)} = \mathcal{A}_{p_{\epsilon_l}}(y_{ij}^{(l)})$. Here a larger value of $\widetilde{S}_i$ indicates more wins of item $i$ in pairwise comparisons. The estimated full ranking of items is derived from the ordering of $\widetilde{\bm{S}}=(\widetilde{S}_1,\ldots,\widetilde{S}_m)$. In the experiment, we uniformly sample the numbers of users and items from $\{150, 151,\ldots, 400\}$ and $\{10, 11,\ldots, 30\}$, respectively, and users' privacy preferences $\epsilon_l$'s are generated from $\text{Uniform}(0.2, 2)$.

\begin{figure}[h!]
        \centering
        \begin{subfigure}[b]{0.49\textwidth}  
            \centering 
            \includegraphics[width=\textwidth]{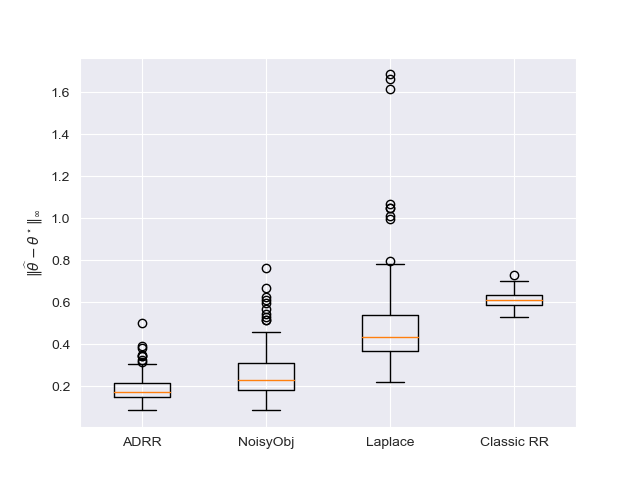}
        \end{subfigure}
        \begin{subfigure}[b]{0.49\textwidth}   
            \centering 
            \includegraphics[width=\textwidth]{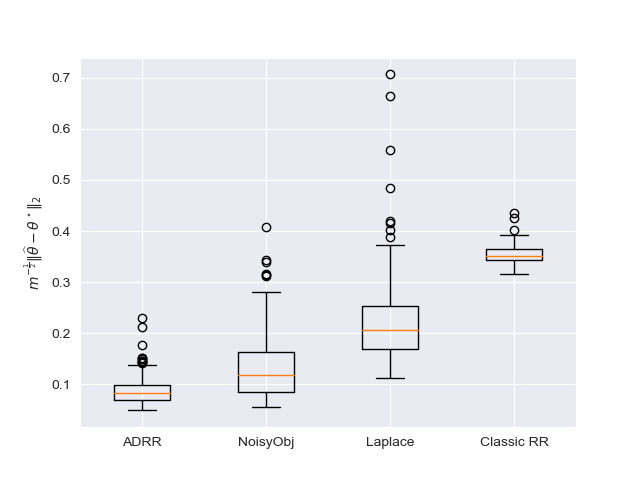}
        \end{subfigure}
        \caption{Comparison of the proposed ADRR method and others under the BTL model regarding $\Vert \widehat{\bm{\theta}} - \bm{\theta}^\star\Vert_{\infty}$ (Left) and $m^{-\frac{1}{2}}\Vert \widehat{\bm{\theta}} - \bm{\theta}^\star\Vert_{2}$ (right).}
        \label{fig:PG5}
    \end{figure}

\begin{figure}[h]
        \centering
                \begin{subfigure}[b]{0.49\textwidth}  
            \centering 
            \includegraphics[width=\textwidth]{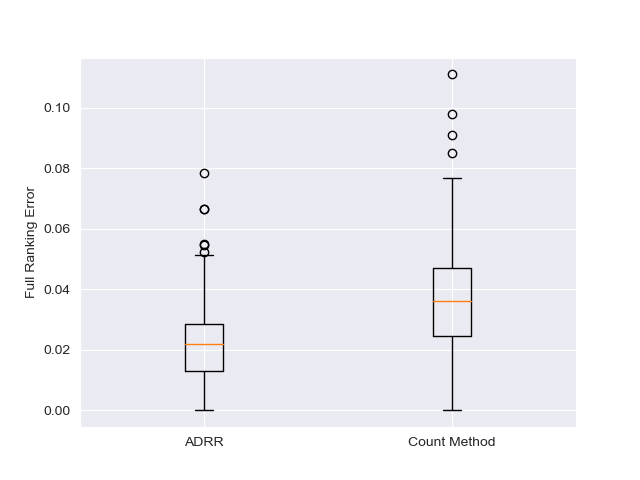} 
        \end{subfigure}
       \begin{subfigure}[b]{0.49\textwidth}   
            \centering 
            \includegraphics[width=\textwidth]{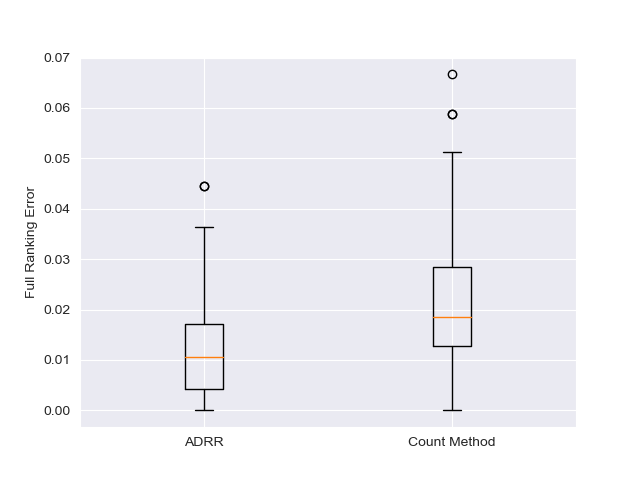}   
       \end{subfigure}
        \caption{The comparison of our method against the nonparametric count method in terms of the full ranking error under the BTL model (Left) and the TM (Right) model.}
        \label{fig:PG6}
        \end{figure}

Figure \ref{fig:PG5} demonstrates that using privatized rankings generated by the ADRR mechanism improves the performance of the rank aggregation task compared to both the Laplace mechanism and the classic RR mechanism, primarily due to the weighting scheme. Moreover, \textit{NoisyObj} surpasses both the Laplace mechanism and the classic RR mechanism in parameter estimation, highlighting the efficiency of central DP over local DP. However, a significant disadvantage of the perturbation method is its inability to accommodate varying privacy preferences of users. To meet all users' privacy preferences, setting $\epsilon$ to the minimal value of $(\epsilon_l)_{l=1}^L$ results in suboptimal performance compared to ADRR. Moreover, when implementing rank aggregation based on output rankings by the classic RR mechanism, achieving consistent estimation of $\bm{\theta}^\star$ becomes unattainable as anticipated. This stems from the fact that output rankings generated by the RR mechanism deviate from the underlying comparison model. This is particularly disastrous if the third party aims to study the popularity of items. Inconsistent parameter estimation leads to misunderstandings regarding the popularity of these items. Furthermore, the debiasing step rectifies this deviation, while the weighting step enhances the utility in ranking items. Additionally, we compare the full ranking recovery of our method with that of the non-parametric count method under the same privacy setting. As shown in Figure \ref{fig:PG6}, our proposed method outperforms the count method when pairwise comparisons are generated from both the BTL and TM models, indicating that the proposed method utilizes privatized rankings more efficiently.

\subsection{Real Application}

In this section, we conduct experiments using a real-world car preference dataset collected by \citet{abbasnejad2013learning} via Amazon Mechanical Turk, a commercial crowdsourcing platform. This dataset is publicly available at \url{https://users.cecs.anu.edu.au/~u4940058/CarPreferences.html}, consisting of genuine pairwise preferences from users. In this experiment, users were asked to choose their preferred car from pairs based on the cars' attributes. The dataset contains 45 pairwise preferences from 10 types of cars, assigned to 60 users, and collects their binary responses. In this dataset, 42 users exhibit intransitive preferences in their comparisons. In other words, 70\% of users fail to provide consistent rankings, which partially aligns with the independence assumption of pairwise comparisons. Furthermore, by estimating users' listwise rankings based on the number of items chosen in all comparisons, we found that intransitive preferences account for 6.98\% of all observed comparisons.

Initially, we employ both the BTL and TM models, as well as the nonparametric count method, to aggregate rankings in the dataset without implementing privacy protection measures. It is important to note that all three methods produce an identical ranking of cars, as illustrated in Table \ref{tab:car_1}. Therefore, we consider this ranking as the ground truth for assessing the ranking errors associated with various privacy-preserving methods.

\begin{table}[ht]
    \centering
    \caption{Comparison of 10 cars in the first experiment and the rankings by three methods. The 6-th type of car is the most preferred, while the 3rd type is the least preferred.}
\scriptsize
    \begin{tabular}{l|cccccccccc}
        \toprule
        Aspects & 1 & 2 &3&4&5& 6&7&8&9&10 \\
        \midrule
Body type &  SUV & Sedan&Sedan&Sedan&SUV&  SUV & Sedan& SUV & Sedan& SUV             \\
Transmission & Manu & Auto & Manu &Manu& Manu&   Auto & Auto & Auto & Auto & Auto    \\
Engine &  2.5L & 5.5L& 4.5& 6.2L& 3.5L& 3.5L& 3.5L &  2.5L& 3.5L& 4.5L          \\
Fuel &  Non-Hy &  Hybrid & Non-Hy &Non-Hy &Non-Hy &Hybrid &Hybrid &Hybrid &Non-Hy &Non-Hy         \\
  \midrule 
Ranking & 
8 &  3 & 10 &  9 & 7&  1&  5&  4&  6&  2
\\
        \bottomrule
    \end{tabular}
    \label{tab:car_1}
\end{table}

In this application, our focus is to compare the full ranking error of our private method with that of various competitive privacy-preserving methods. To this end, we generate privacy parameters $\epsilon_l$'s for 60 users from $\text{Uniform}(A,A+1)$, where $A$ follows $\text{Uniform}(0.2,2)$. Here, $A$ aims to imitate both low and high privacy scenarios. This configuration aims to mimic various levels of privacy protection, and this setup is repeated in 200 times. To evaluate the efficacy of our method, we compute the full ranking errors of all methods and conduct pairwise t-tests to compare the full ranking error of our method with those of other methods, as presented in Table \ref{tab:car_2}. The results indicate that our method significantly outperforms the other methods in full ranking recovery under both the BTL and TM models. This result demonstrates that the proposed method attains greater utility while maintaining the same level of privacy protection, illustrating a more delicate privacy-utility tradeoff in our approach.

    \begin{table}[ht]
    \centering
    \caption{The pairwise t-tests comparing our method to competing methods.}
\scriptsize
    \begin{tabular}{c|c|ccc}
        \toprule
    &    ADRR (BTL)   & Count Method                & Classic RR (BTL)            & Laplace (BTL)   \\
    \midrule
    Mean Full Ranking Error & 
    0.0757 (0.0030)            & 0.0859 (0.0037)         & 0.0841 (0.0036)        &  0.1198 (0.0046)  \\
     \midrule
    \multirow{2}{*}{Pairwise T-test} &
 t-statistic &    4.1219                 & 3.4277                & 9.4314   \\
& p-value & $5.5 \times 10^{-5}$ & $7.4\times 10^{-4}$ & $1.1 \times 10^{-17}$ \\
        \midrule
    &    ADRR (TM)   & Count Method                & Classic RR (TM)            & Laplace (TM)   \\
    \midrule
    Mean Full Ranking Error & 
     0.0767 (0.0031)            & 0.0859 (0.0037)        & 0.0844 (0.0036)       & 0.1193 (0.0046)   \\
     \midrule
    \multirow{2}{*}{Pairwise T-test} &
 t-statistic &   3.6713                 & 3.0994                & 9.1284     \\
& p-value &  $3.1 \times 10^{-4}$  & $2.2\times 10^{-3}$ & $7.9 \times 10^{-17}$
\\
        \bottomrule
    \end{tabular}
    \label{tab:car_2}
\end{table}

\section*{Supplementary Materials}

The supplementary file provides a detailed discussion of our method and explores several potential extensions. Specifically, it discusses the phenomenon of intransitive preferences in pairwise comparisons, which arises from the independence assumption across users. To address this issue, we propose an extension based on mixed-effects models, where user-specific effects are treated as random effects to relax the independence assumption of pairwise comparisons for a user, and we validate this approach through experimental analysis. We also present the model selection process for ranking models within the class of Linear Stochastic Transitivity (LST) models, introducing a likelihood-based method. The effectiveness of this approach is demonstrated through an experiment. Additionally, we offer practical guidance on analyzing the privacy budget based on our theoretical findings. Furthermore, we investigate the relationship between central DP and local DP in the context of ranking data. Finally, the supplementary file includes additional experimental results, along with detailed proofs of the theorems and lemmas presented in the main text.

\section*{Acknowledgments}
The authors sincerely thank the co-editor Professor Annie Qu, the guest editor, the associate editor, and four anonymous reviewers for their insightful comments and suggestions, which have significantly improved the quality of this article.

\section*{Conflict of Interest Statement}
The authors report there are no competing interests to declare.

\baselineskip=16pt
\renewcommand\refname{References}
\bibliographystyle{apalike}
\bibliography{Pairwise_Ranking_DP}

\begin{thebibliography}{}

\bibitem[Abbasnejad et~al., 2013]{abbasnejad2013learning}
Abbasnejad, E., Sanner, S., Bonilla, E.~V., and Poupart, P. (2013).
\newblock Learning community-based preferences via dirichlet process mixtures of gaussian processes.
\newblock In {\em Twenty-third International Joint Conference on Artificial Intelligence}.

\bibitem[Ackerman et~al., 2013]{ackerman2013elections}
Ackerman, M., Choi, S.-Y., Coughlin, P., Gottlieb, E., and Wood, J. (2013).
\newblock Elections with partially ordered preferences.
\newblock {\em Public Choice}, 157:145--168.

\bibitem[Alabi et~al., 2022]{alabi2022private}
Alabi, D., Ghazi, B., Kumar, R., and Manurangsi, P. (2022).
\newblock Private rank aggregation in central and local models.
\newblock In {\em Proceedings of the AAAI Conference on Artificial Intelligence}, volume~36, pages 5984--5991.

\bibitem[Bi and Shen, 2022]{bi2023distribution}
Bi, X. and Shen, X. (2022).
\newblock Distribution-invariant differential privacy.
\newblock {\em Journal of Econometrics}.

\bibitem[Bradley and Terry, 1952]{bradley1952rank}
Bradley, R.~A. and Terry, M.~E. (1952).
\newblock Rank analysis of incomplete block designs: I. the method of paired comparisons.
\newblock {\em Biometrika}, 39(3/4):324--345.

\bibitem[Cai et~al., 2021]{cai2021cost}
Cai, T.~T., Wang, Y., and Zhang, L. (2021).
\newblock The cost of privacy: Optimal rates of convergence for parameter estimation with differential privacy.
\newblock {\em The Annals of Statistics}, 49(5):2825--2850.

\bibitem[Cai et~al., 2023]{cai2023score}
Cai, T.~T., Wang, Y., and Zhang, L. (2023).
\newblock Score attack: A lower bound technique for optimal differentially private learning.
\newblock {\em arXiv preprint arXiv:2303.07152}.

\bibitem[Carlson and Montgomery, 2017]{carlson2017pairwise}
Carlson, D. and Montgomery, J.~M. (2017).
\newblock A pairwise comparison framework for fast, flexible, and reliable human coding of political texts.
\newblock {\em American Political Science Review}, 111(4):835--843.

\bibitem[Chen et~al., 2022a]{AndersonZhang2}
Chen, P., Gao, C., and Zhang, A.~Y. (2022a).
\newblock Optimal full ranking from pairwise comparisons.
\newblock {\em The Annals of Statistics}, 50(3):1775--1805.

\bibitem[Chen et~al., 2022b]{AndersonZhang}
Chen, P., Gao, C., and Zhang, A.~Y. (2022b).
\newblock Partial recovery for top-{$k$} ranking: optimality of {MLE}, and suboptimality of the spectral method.
\newblock {\em The Annals of Statistics}, 50(3):1618--1652.

\bibitem[Chen et~al., 2013]{chen2013pairwise}
Chen, X., Bennett, P.~N., Collins-Thompson, K., and Horvitz, E. (2013).
\newblock Pairwise ranking aggregation in a crowdsourced setting.
\newblock In {\em Proceedings of the sixth ACM International Conference on Web Search and Data Mining}, pages 193--202.

\bibitem[Chen et~al., 2019]{chen2019spectral}
Chen, Y., Fan, J., Ma, C., and Wang, K. (2019).
\newblock Spectral method and regularized mle are both optimal for top-k ranking.
\newblock {\em Annals of statistics}, 47(4):2204.

\bibitem[Chen and Suh, 2015]{chen2015spectral}
Chen, Y. and Suh, C. (2015).
\newblock Spectral mle: Top-k rank aggregation from pairwise comparisons.
\newblock In {\em International Conference on Machine Learning}, pages 371--380. PMLR.

\bibitem[Chhor and Sentenac, 2023]{chhor2023robust}
Chhor, J. and Sentenac, F. (2023).
\newblock Robust estimation of discrete distributions under local differential privacy.
\newblock In {\em International Conference on Algorithmic Learning Theory}, pages 411--446. PMLR.

\bibitem[Costante et~al., 2013]{costante2013privacy}
Costante, E., Paci, F., and Zannone, N. (2013).
\newblock Privacy-aware web service composition and ranking.
\newblock In {\em 2013 IEEE 20th International Conference on Web Services}, pages 131--138. IEEE.

\bibitem[Dawkins, 1969]{dawkins1969threshold}
Dawkins, R. (1969).
\newblock A threshold model of choice behaviour.
\newblock {\em Animal Behaviour}, 17:120--133.

\bibitem[De~la Cruz et~al., 2011]{AE8}
De~la Cruz, R., Marshall, G., and Quintana, F.~A. (2011).
\newblock Logistic regression when covariates are random effects from a non-linear mixed model.
\newblock {\em Biometrical journal}, 53(5):735--749.

\bibitem[Diaconis and Graham, 1977]{diaconis1977spearman}
Diaconis, P. and Graham, R.~L. (1977).
\newblock Spearman's footrule as a measure of disarray.
\newblock {\em Journal of the Royal Statistical Society Series B: Statistical Methodology}, 39(2):262--268.

\bibitem[Duchi et~al., 2018]{duchi2018minimax}
Duchi, J.~C., Jordan, M.~I., and Wainwright, M.~J. (2018).
\newblock Minimax optimal procedures for locally private estimation.
\newblock {\em Journal of the American Statistical Association}, 113(521):182--201.

\bibitem[Dwork, 2006]{dwork2006differential}
Dwork, C. (2006).
\newblock Differential privacy.
\newblock In {\em International Colloquium on Automata, Languages, and Programming}, pages 1--12. Springer.

\bibitem[Dwork et~al., 2001]{dwork2001rank}
Dwork, C., Kumar, R., Naor, M., and Sivakumar, D. (2001).
\newblock Rank aggregation methods for the web.
\newblock In {\em Proceedings of the 10th International Conference on World Wide Web}, pages 613--622.

\bibitem[Franklin, 2012]{franklin2012matrix}
Franklin, J.~N. (2012).
\newblock {\em Matrix theory}.
\newblock Courier Corporation.

\bibitem[Friedman et~al., 2015]{friedman2015privacy}
Friedman, A., Knijnenburg, B.~P., Vanhecke, K., Martens, L., and Berkovsky, S. (2015).
\newblock Privacy aspects of recommender systems.
\newblock {\em Recommender systems handbook}, pages 649--688.

\bibitem[Hay et~al., 2017]{hay2017differentially}
Hay, M., Elagina, L., and Miklau, G. (2017).
\newblock Differentially private rank aggregation.
\newblock In {\em Proceedings of the 2017 SIAM International Conference on Data Mining}, pages 669--677. SIAM.

\bibitem[Horn and Johnson, 2012]{horn2012matrix}
Horn, R.~A. and Johnson, C.~R. (2012).
\newblock {\em Matrix analysis}.
\newblock Cambridge university press.

\bibitem[Ichihashi, 2020]{ichihashi2020online}
Ichihashi, S. (2020).
\newblock Online privacy and information disclosure by consumers.
\newblock {\em American Economic Review}, 110(2):569--595.

\bibitem[Jeong et~al., 2022]{jeong2022ranking}
Jeong, M., Dytso, A., and Cardone, M. (2022).
\newblock Ranking recovery under privacy considerations.
\newblock {\em Transactions on Machine Learning Research}.

\bibitem[Kalloori et~al., 2018]{kalloori2018eliciting}
Kalloori, S., Ricci, F., and Gennari, R. (2018).
\newblock Eliciting pairwise preferences in recommender systems.
\newblock In {\em Proceedings of the 12th ACM Conference on Recommender Systems}, pages 329--337.

\bibitem[Karl{\'e} and Tyagi, 2023]{karle2023dynamic}
Karl{\'e}, E. and Tyagi, H. (2023).
\newblock Dynamic ranking with the btl model: a nearest neighbor based rank centrality method.
\newblock {\em Journal of Machine Learning Research}, 24(269):1--57.

\bibitem[Klein and Rio, 2005]{klein2005concentration}
Klein, T. and Rio, E. (2005).
\newblock Concentration around the mean for maxima of empirical processes.
\newblock {\em The Annals of Probability}, 33(3):1060--1077.

\bibitem[Klimenko, 2015]{klimenko2015intransitivity}
Klimenko, A.~Y. (2015).
\newblock Intransitivity in theory and in the real world.
\newblock {\em Entropy}, 17(6):4364--4412.

\bibitem[Koltchinskii, 2011]{koltchinskii2011oracle}
Koltchinskii, V. (2011).
\newblock {\em Oracle inequalities in empirical risk minimization and sparse recovery problems: {\'E}cole D’{\'E}t{\'e} de Probabilit{\'e}s de Saint-Flour XXXVIII-2008}, volume 2033.
\newblock Springer Science \& Business Media.

\bibitem[Lee, 2019]{lee2019u}
Lee, A.~J. (2019).
\newblock {\em U-statistics: Theory and Practice}.
\newblock Routledge.

\bibitem[Lee, 2015]{lee2015efficient}
Lee, D.~T. (2015).
\newblock Efficient, private, and eps-strategyproof elicitation of tournament voting rules.
\newblock In {\em Twenty-Fourth International Joint Conference on Artificial Intelligence}.

\bibitem[Liu et~al., 2023]{liu2023lagrangian}
Liu, Y., Fang, E.~X., and Lu, J. (2023).
\newblock Lagrangian inference for ranking problems.
\newblock {\em Operations Research}, 71(1):202--223.

\bibitem[Luce, 2012]{luce2012individual}
Luce, R.~D. (2012).
\newblock {\em Individual choice behavior: A theoretical analysis}.
\newblock Courier Corporation.

\bibitem[McCarthy and Santucci, 2021]{mccarthy2021ranked}
McCarthy, D. and Santucci, J. (2021).
\newblock Ranked choice voting as a generational issue in modern american politics.
\newblock {\em Politics \& Policy}, 49(1):33--60.

\bibitem[Negahban et~al., 2016]{negahban2016rank}
Negahban, S., Oh, S., and Shah, D. (2016).
\newblock Rank centrality: Ranking from pairwise com-parisons.
\newblock {\em Operations Research}, 65(1):266--287.

\bibitem[Okazaki et~al., 2009]{okazaki2009consumer}
Okazaki, S., Li, H., and Hirose, M. (2009).
\newblock Consumer privacy concerns and preference for degree of regulatory control.
\newblock {\em Journal of Advertising}, 38(4):63--77.

\bibitem[Oliveira et~al., 2018]{oliveira2018new}
Oliveira, I.~F., Ailon, N., and Davidov, O. (2018).
\newblock A new and flexible approach to the analysis of paired comparison data.
\newblock {\em Journal of Machine Learning Research}, 19(60):1--29.

\bibitem[Oliveira et~al., 2020]{oliveira2020rank}
Oliveira, S.~E., Diniz, V., Lacerda, A., Merschmanm, L., and Pappa, G.~L. (2020).
\newblock Is rank aggregation effective in recommender systems? an experimental analysis.
\newblock {\em ACM Transactions on Intelligent Systems and Technology (TIST)}, 11(2):1--26.

\bibitem[Pinheiro and Bates, 1995]{AE7}
Pinheiro, J.~C. and Bates, D.~M. (1995).
\newblock Approximations to the log-likelihood function in the nonlinear mixed-effects model.
\newblock {\em Journal of Computational and Graphical Statistics}, 4(1):12--35.

\bibitem[Shah and Wainwright, 2018]{shah2018simple}
Shah, N.~B. and Wainwright, M.~J. (2018).
\newblock Simple, robust and optimal ranking from pairwise comparisons.
\newblock {\em Journal of Machine Learning Research}, 18(199):1--38.

\bibitem[Shang et~al., 2014]{shang2014application}
Shang, S., Wang, T., Cuff, P., and Kulkarni, S. (2014).
\newblock The application of differential privacy for rank aggregation: Privacy and accuracy.
\newblock In {\em 17th International Conference on Information Fusion (FUSION)}, pages 1--7. IEEE.

\bibitem[Suh et~al., 2017]{suh2017adversarial}
Suh, C., Tan, V.~Y., and Zhao, R. (2017).
\newblock Adversarial top-$ k $ ranking.
\newblock {\em IEEE Transactions on Information Theory}, 63(4):2201--2225.

\bibitem[Sz{\"o}r{\'e}nyi et~al., 2015]{szorenyi2015online}
Sz{\"o}r{\'e}nyi, B., Busa-Fekete, R., Paul, A., and H{\"u}llermeier, E. (2015).
\newblock Online rank elicitation for plackett-luce: A dueling bandits approach.
\newblock {\em Advances in Neural Information Processing Systems}, 28.

\bibitem[Tatli et~al., 2024]{tatli2024learning}
Tatli, G., Chen, Y., and Vinayak, R.~K. (2024).
\newblock Learning populations of preferences via pairwise comparison queries.
\newblock In {\em International Conference on Artificial Intelligence and Statistics}, pages 1720--1728. PMLR.

\bibitem[Thurstone, 1994]{thurstone1994law}
Thurstone, L.~L. (1994).
\newblock A law of comparative judgment.
\newblock {\em Psychological Review}, 101(2):266.

\bibitem[Tropp, 2012]{tropp2012user}
Tropp, J.~A. (2012).
\newblock User-friendly tail bounds for sums of random matrices.
\newblock {\em Foundations of Computational Mathematics}, 12:389--434.

\bibitem[Tuerlinckx et~al., 2006]{AE6}
Tuerlinckx, F., Rijmen, F., Verbeke, G., and De~Boeck, P. (2006).
\newblock Statistical inference in generalized linear mixed models: A review.
\newblock {\em British Journal of Mathematical and Statistical Psychology}, 59(2):225--255.

\bibitem[Tversky, 1969]{tversky1969intransitivity}
Tversky, A. (1969).
\newblock Intransitivity of preferences.
\newblock {\em Psychological review}, 76(1):31.

\bibitem[Van~Erven and Harremos, 2014]{van2014renyi}
Van~Erven, T. and Harremos, P. (2014).
\newblock R{\'e}nyi divergence and kullback-leibler divergence.
\newblock {\em IEEE Transactions on Information Theory}, 60(7):3797--3820.

\bibitem[Wang et~al., 2017]{wang2017locally}
Wang, T., Blocki, J., Li, N., and Jha, S. (2017).
\newblock Locally differentially private protocols for frequency estimation.
\newblock In {\em 26th USENIX Security Symposium (USENIX Security 17)}, pages 729--745.

\bibitem[Warner, 1965]{warner1965randomized}
Warner, S.~L. (1965).
\newblock Randomized response: A survey technique for eliminating evasive answer bias.
\newblock {\em Journal of the American Statistical Association}, 60(309):63--69.

\bibitem[Weng and Lin, 2011]{weng2011bayesian}
Weng, R.~C. and Lin, C.-J. (2011).
\newblock A bayesian approximation method for online ranking.
\newblock {\em Journal of Machine Learning Research}, 12(1).

\bibitem[Xu et~al., 2023]{xu2023binary}
Xu, S., Wang, C., Sun, W.~W., and Cheng, G. (2023).
\newblock Binary classification under local label differential privacy using randomized response mechanisms.
\newblock {\em Transactions on Machine Learning Research}.

\bibitem[Yan et~al., 2020]{yan2020private}
Yan, Z., Li, G., and Liu, J. (2020).
\newblock Private rank aggregation under local differential privacy.
\newblock {\em International Journal of Intelligent Systems}, 35(10):1492--1519.

\bibitem[Yellott~Jr, 1977]{yellott1977relationship}
Yellott~Jr, J.~I. (1977).
\newblock The relationship between luce's choice axiom, thurstone's theory of comparative judgment, and the double exponential distribution.
\newblock {\em Journal of Mathematical Psychology}, 15(2):109--144.

\bibitem[Zhu et~al., 2023]{zhu2023principled}
Zhu, B., Jiao, J., and Jordan, M.~I. (2023).
\newblock Principled reinforcement learning with human feedback from pairwise or $ k $-wise comparisons.
\newblock {\em arXiv preprint arXiv:2301.11270}.

\end{thebibliography}

\appendix
\newpage
\baselineskip=24pt
\setcounter{page}{1}
\setcounter{equation}{0}
\setcounter{section}{0}
\renewcommand{\thesection}{S.\arabic{section}}
\renewcommand{\thelemma}{S\arabic{lemma}}
\renewcommand{\thetheorem}{S\arabic{theorem}}
\renewcommand{\theequation}{S\arabic{equation}}
\begin{center}
{\Large\bf Supplementary Materials} \\
\medskip
{\Large\bf ``Rate-Optimal Rank Aggregation with Private Pairwise Rankings "}  \\
\bigskip
\vspace{0.2in} 
\end{center}
\bigskip

In this supplementary file, we first present various extensions of the proposed rank aggregation framework in Section \ref{AR:Discussion}. Next, we establish the connection between central DP and local DP in pairwise rankings in Section \ref{Sec:Connection}. The former aims to protect an individual's overall preferences, while the latter safeguards each comparison answer. In Section \ref{AR:Experiments}, we present additional experimental results, including tables of all scenarios in the simulation and the privacy-utility tradeoff in our real application. In Section \ref{SuppSec:Thm}, we introduce the necessary notations and present the proofs of all theorems and lemmas.

\vspace{5mm}

\section{Summary and Extensions}
\label{AR:Discussion}

\vspace{5mm}
In this paper, we address the challenge of ranking aggregation with privacy protection for pairwise ranking data from individuals. Specifically, we focus on pairwise ranking data and employ the randomized response (RR) mechanism to ensure privacy in pairwise comparison values. However, two practical issues arise in this process. First, the RR mechanism inevitably distorts the raw distributions, leading to inconsistent parameter estimation. Second, varying privacy preferences under local DP undermine the robustness of existing estimators. To address these issues, we developed an adaptive debiased RR mechanism to perturb pairwise rankings in the context of rank aggregation. We demonstrate that, under the proposed mechanism, the resulting estimator achieves the minimax optimal rate. The efficacy of the proposed method is verified through extensive simulations and a real-world dataset.

In this section, we explore various extensions of the rank aggregation framework considered in this paper. In Section \ref{AR_sub_Indep}, we discuss the independence assumption of pairwise comparisons within the LST models. Section \ref{AR_sub_UserDep} delves into potential extensions of the LST models to accommodate user-dependent rankings via the mixed-effects model. Section \ref{AR_sub_MS} introduces a method for model selection within our rank aggregation framework. Finally, in Section \ref{AR_sub_Guidance}, we discuss the practical insights derived from our theoretical results.

\subsection{Independence Assumption of Pairwise Comparisons}
\label{AR_sub_Indep}
In the general LST models, a common assumption is the independence of pairwise comparisons. Under this assumption, it is possible to observe intransitive preferences in the collected data. This phenomenon is significant in practice because decision-makers often exhibit intransitive preferences due to cognitive biases and heuristics \citep{tversky1969intransitivity,klimenko2015intransitivity}, leading to inconsistent choices. For example, \citet{tversky1969intransitivity} presented each participant with 10 unordered pairs of gambles 20 times, finding that individual participants could behave intransitively in pairwise choices. This phenomenon is also observed in the car preference dataset used in our real application \citep{abbasnejad2013learning}. The authors included several control questions, randomly selected from the preferences and presented in reverse order to verify the consistency of the responses. They found that some users presented inconsistent results compared with their previous choices. Additionally, for each user's pairwise comparisons, we obtained their listwise rankings by counting the number of times items were chosen and calculated the proportion of intransitive preferences. Among all pairwise comparisons, roughly 7\% exhibited conflicting preferences with their previous choices. This is expected in reality, as people often make pairwise comparisons based on first impressions, with no guarantee that these comparisons accurately reflect their true preferences. Despite this, from a different perspective, the presence of intransitive preferences may indicate irrationality, suggesting significant noise in pairwise comparison data. 

A natural solution to address this issue is to collect listwise or partial listwise rankings rather than pairwise comparisons. However, in practice, users may lack the patience to provide complete listwise rankings when the item set is large. Additionally, privacy protection definitions at the individual level, along with the fundamental trade-off between privacy and utility, remain unclear for these two types of ranking data. For instance, some users may wish to protect only their top-$K$ preferences, while others may want to safeguard their top choice alone. Developing effective mechanisms that accommodate diverse privacy preferences is of practical importance. Moreover, as privacy-preserving mechanisms are developed, understanding the privacy-utility trade-off for listwise rankings remains a critical challenge. There is a pressing need for robust algorithms that can handle diverse types of ranking data while ensuring differential privacy. Such algorithms should be adaptable to various ranking formats and user preferences without compromising privacy. In summary, addressing these challenges involves defining individual-level privacy across different types of ranking data, developing robust and scalable algorithms, and conducting empirical evaluations to assess performance.

\subsection{Mixed-Effect Models: User-Dependent Pairwise Comparison}
\label{AR_sub_UserDep}

In this paper, we examine rank aggregation within the general linear stochastic transitivity (LST) models, which assume that users' pairwise rankings are independent and influenced solely by the preference parameters of the items. However, this assumption may be overly strong, as pairwise comparisons are highly correlated. To address the issue of independence in comparisons, we apply a mixed-effects model by incorporating random user effects $\{\gamma_l^\star\}_{l=1}^L$ into the general LST models. Specifically, we consider
\begin{align*}
  & \gamma_l^\star \sim \mathbb{P}_{\sigma} \quad \mbox{and} \quad y_{ij}^{(l)} \sim F(\gamma_l^\star + \theta_i^\star - \theta_j^\star),
\end{align*}
for $l \in [L]$ and $i < j$, where $\sigma$ represents the parameter of $\mathbb{P}_{\sigma}$. For example, $\mathbb{P}_{\sigma} = N(0,\sigma^2)$. Suppose $\gamma_l^\star \sim N(0,\sigma^2)$, then the objective function (without privacy protection and missing comparisons) becomes
\begin{align}
\label{RE1}
    L(\bm{\theta},\sigma)  \varpropto
    \prod_{l=1}^L \prod_{i<j}
    \int_{-\infty}^{+\infty}
   \left( F(\gamma_l+\theta_i-\theta_j)
    \right)^{y_{ij}^{(l)}}
       \left( 1-F(\gamma_l+\theta_i-\theta_j)
    \right)^{1-y_{ij}^{(l)}} e^{-\frac{\gamma_l^2}{2\sigma^2}}d \gamma_l.
\end{align}
This objective function is challenging to optimize due to the difficulty of solving the integral. This is a common problem encountered in nonlinear mixed-effects models \citep{AE6,AE7}. Solving this optimization task relies on some approximation techniques for $L(\bm{\theta}, \sigma)$.

\textbf{Estimation \& Experiments.} Following \citet{AE8}, we employ a two-step estimator for parameter estimation. The main idea is to treat random effects as fixed effects and iteratively optimize \(\bm{\gamma}\) and \(\bm{\theta}\). After this, we can estimate \(\sigma\) based on the resulting estimator of \(\bm{\gamma}^\star\). For example, in the BTL model, the optimization task for solving \(\bm{\theta}\) and \(\bm{\gamma}\) can be organized as
\begin{align*}
    \min_{\bm{\theta},\bm{\gamma}}
    \mathcal{L}(\bm{\theta},\bm{\gamma})=
    \sum_{i<j}\sum_{l=1}^L \left\{ -y_{ij}^{(l)}(\gamma_{l}+\theta_i-\theta_j)
    +\log(1+e^{\gamma_l+\theta_i-\theta_j})
    \right\}+\lambda \Vert \bm{\theta}\Vert_2^2.
\end{align*}
Here, the optimization task is jointly convex in $(\bm{\gamma}, \bm{\theta})$. Therefore, applying gradient descent leads to convergence to the global minimizer. Let $(\widehat{\bm{\theta}}, \widehat{\bm{\gamma}}) = \argmin_{\bm{\theta}, \bm{\gamma}} \mathcal{L}(\bm{\theta}, \bm{\gamma})$ be the resultant estimator. Once we obtain $\widehat{\bm{\gamma}}$, we estimate $\sigma$ as
\begin{align*}
    \widehat{\sigma} = \sqrt{\frac{1}{L-1}\sum_{l=1}^L (\widehat{\gamma}_l-\overline{\gamma})^2},
\end{align*}
where $\overline{\gamma} = L^{-1}\sum_{l=1}^L \widehat{\gamma}_l$.

To assess the effectiveness of this estimation procedure, we have added new experiments to analyze this estimator. Specifically, we generate $\gamma_l^\star$, for $l = 1, \ldots, L$, from a normal distribution $N(0,1)$ and consider $(L, m) \in \{100, 200, 300\} \times \{10, 20, 30\}$. For each experimental setting, we perform 200 replications and report the average estimation errors, given by $|\widehat{\sigma} - \sigma|$ with $\sigma = 1$ and $m^{-\frac{1}{2}} \Vert \widehat{\bm{\theta}} - \bm{\theta}^\star \Vert_2$, which are summarized in the following table.
\begin{table}[h]
\caption{The averaged estimation errors $L^{-\frac{1}{2}}\Vert \widehat{\bm \gamma}-\bm{\gamma}^\star\Vert_2$ and $|\widehat{\sigma}-\sigma|$ in 200 replicates under varying values of $(L,m)$.}
    \centering
    \label{tab:RadnomEffect}
    \begin{tabular}{c|c|c|c|c}
\toprule
                        &   $m$ & $L=100 $          & $L=200 $          & $L=300$           \\
                        \hline
$m^{-\frac{1}{2}}\Vert \widehat{\bm \theta}-\bm{\theta}^\star\Vert_2$ &  10 &  0.0815 (0.0224) & 0.0596 (0.0175) & 0.0476 (0.0145) \\
\hline
$m^{-\frac{1}{2}}\Vert \widehat{\bm \theta}-\bm{\theta}^\star\Vert_2$ &  20 & 0.0535 (0.0087) & 0.0384 (0.0071) & 0.0319 (0.0064) \\
\hline
$m^{-\frac{1}{2}}\Vert \widehat{\bm \theta}-\bm{\theta}^\star\Vert_2$ &  30 & 0.0434 (0.0065) & 0.0300 (0.0042) & 0.0252 (0.0032) \\
\hline
$|\widehat{\sigma}-\sigma|$ &  10 &  0.1974 (0.1527) & 0.1576 (0.0951) & 0.1511 (0.0726) \\
\hline
$|\widehat{\sigma}-\sigma|$ &  20 & 0.0658 (0.0549) & 0.0445 (0.0363) & 0.0391 (0.0306) \\
\hline
$|\widehat{\sigma}-\sigma|$ &  30 & 0.0589 (0.0430) & 0.0388 (0.0299) & 0.0319 (0.0229) \\
\bottomrule
\end{tabular}
\end{table}
Clearly, as either \(L\) or \(m\) increases, the estimation error of \(\widehat{\bm{\theta}}\) decreases. This improvement occurs because an increase in either \(L\) or \(m\) provides more information about \(\bm{\theta}^\star\). Similarly, the estimation error of \(\widehat{\sigma}\) also decreases with an increase in \(m\) or \(L\), but for two distinct reasons. First, as \(m\) increases, \(\gamma_l^\star\) can be estimated more accurately, which in turn leads to a more precise estimation of \(\sigma\). Second, as \(L\) increases, \(\widehat{\sigma}\) utilizes more \(\widehat{\gamma}_l\) values to estimate \(\sigma\), thereby improving the accuracy of the estimation.

\subsection{Model Selection of Ranking Models}
\label{AR_sub_MS}

In our framework, a key assumption is that the downstream task accurately specifies the model for pairwise comparisons. However, a major challenge is selecting the appropriate model for these comparisons. To address this, we propose a model selection approach based on likelihood. In the following sections, we outline the general procedure for model selection among the linear stochastic transitivity (LST) models, leveraging likelihood-based criteria to identify the correct model from the set of all LST models.

Define $\mathcal{L}_{\lambda}(\bm{\theta},F)$ as:
\begin{align*}
\mathcal{L}_{\lambda}(\bm{\theta},F)=&
-\sum_{l=1}^L 
\sum_{i<j}
a_{ij}^{(l)}
\left\{
z_{ij}^{(l)} \log
F(\theta_i-\theta_j)
+
(w_l-z_{ij}^{(l)})
\log
\left(
1-F(\theta_i-\theta_j)\right)\right\}+\lambda \Vert \bm{\theta}\Vert_2^2,
\end{align*}
where \( F \) represents any parametric model within the class of LST models, such as the BTL model \( F^{BTL}(x) = \frac{e^x}{1 + e^x} \) and the TM model \( F^{TM}(x) = \int_{-\infty}^{x} \frac{1}{\sqrt{2\pi}} e^{-x^2 / 2} \, dx \). For demonstration purposes, we use the BTL and TM models. We define
\begin{align*}
    \widehat{\bm{\theta}}^{BTL} = \argmin_{\bm{\theta}}
    \mathcal{L}_{\lambda}(\bm{\theta},F^{BTL}) \mbox{ and }
    \widehat{\bm{\theta}}^{TM} = \argmin_{\bm{\theta}}
    \mathcal{L}_{\lambda}(\bm{\theta},F^{TM}).
\end{align*}
Then we compare the likelihoods for model selection.
\begin{align*}
   \mathcal{L}_{0}(\widehat{\bm{\theta}}^{BTL},F^{BTL})<
   \mathcal{L}_{0}(\widehat{\bm{\theta}}^{TM},F^{TM})
   \Rightarrow
   F(\cdot) = F^{BTL}.
\end{align*}
The basic idea is to treat \( F(\cdot) \) as a parameter and choose the best parametric model.

To demonstrate the effectiveness of the proposed model selection method, we conduct experiments for the cases \( (L, m) \in \{100, 200, 400, 800\} \times \{10, 20, 30\} \). In each scenario, we randomly select \( F(\cdot) \) from the BTL and TM models, repeating each setup 200 times. The results are reported as the percentage of correct model specification using the proposed method.
\begin{table}[h]

\caption{ The percentage of correct model specification in 200 replicates under varying values of $(L,m)$.}
    \centering
    \begin{tabular}{c|cccc}
        \toprule
      $m$  & $L=100$ & $L=200$ & $L=400$ & $L=800$ \\
      \midrule
    10  & 0.630 & 0.680 & 0.750 & 0.795 \\
        20  & 0.695 & 0.795 & 0.870 & 0.895 \\
        30  & 0.795 & 0.845 & 0.930 & 0.940\\
\bottomrule
        \end{tabular}
\end{table}

From the table above, it can be observed that as either \( L \) or \( m \) increases, the probability of selecting the correct model specification also increases. Specifically, when \( (L, m) = (800, 30) \), the probability of correctly specifying the model reaches 94\%.

\subsection{Practical Guidance}
\label{AR_sub_Guidance}
In this section, we provide some practical guidance derived from our fundamental privacy-utility tradeoff in rank aggregation. Suppose a company (the third party) aims to study the popularity of its products and commissions an online platform to collect users' pairwise comparison answers for rank aggregation. To evaluate the utility of privatized rankings, the online platform can utilize the result from Theorem \ref{Thm:Estimation_General}.
\begin{align*}
    \Vert \widehat{\bm{\theta}} - \bm{\theta}^\star\Vert_{\infty}
    \asymp \sqrt{\frac{1}{mLpB(\bm{\epsilon})}}=
        \sqrt{\frac{1}{mp\sum_{l=1}^L \left(\frac{e^{\epsilon_l}-1}{e^{\epsilon_l}+1}\right)^2}},
\end{align*}
where the logarithmic term is omitted. Based on this result, when an online platform collects pairwise comparisons along with users' privacy preferences, it can use the term $\sum_{l=1}^L \left(\frac{e^{\epsilon_l} - 1}{e^{\epsilon_l} + 1}\right)^2$ to assess the quality of the collected data. In practice, the platform can establish a threshold $\alpha$ for this term, where a higher value of $\alpha$ corresponds to greater information about the underlying true ranking. The platform can then continue collecting data until the condition $\sum_{l=1}^L \left(\frac{e^{\epsilon_l} - 1}{e^{\epsilon_l} + 1}\right)^2 > \alpha$ is met, as this term naturally increases with $L$. A large value of $\sum_{l=1}^L \left(\frac{e^{\epsilon_l} - 1}{e^{\epsilon_l} + 1}\right)^2$ indicates that the privatized rankings contain sufficient information to ensure that the estimation error $\Vert \widehat{\bm{\theta}} - \bm{\theta}^\star \Vert_{\infty}$ remains small. Therefore, in practice, the online platform can collect privatized rankings until a specific criterion is met, ensuring that users' preferences are protected. Additionally, in our setting, the privacy protections for users are independent and based solely on their personal preferences. Therefore, no matter what privacy preferences other users choose, they will not affect the privacy protection of an individual user.

\vspace{5mm}

\section{The Relationship Between Central and Local DP in Ranking Data}
\label{Sec:Connection}

\vspace{5mm}
In this section, we study the application of $\epsilon$-DP on the ranking data and establish its connection to the local DP considered in this paper. Let $\mathcal{D} = \{\bm{y}^{(l)}\}_{l=1}^L$ be a set of rankings and $\mathcal{D}'$ be a neighboring set of rankings that differs from $\mathcal{D}$ in only one user's ranking, say $\bm{y}^{(l_0)\prime}$. The $\epsilon$-DP on the ranking data \citep{hay2017differentially,alabi2022private} is given as: a mechanism $\mathcal{M}$ is $\epsilon$-differentially private if 
\begin{align}
\label{Anq1}
\sup_{\mathcal{S}}
    \frac{\mathbb{P}(\mathcal{M}(\mathcal{D}) \in \mathcal{S})}{
    \mathbb{P}(\mathcal{M}(\mathcal{D}')\in \mathcal{S})
    } \leq e^{\epsilon},
\end{align}
where $\mathcal{S}$ is a measurable set of the output space. In rank aggregation tasks, $\mathcal{M}(\cdot)$ maps the set of rankings to either a full ranking or ranking scores of items. The $\epsilon$-DP for ranking requires that the output distribution remains robust to the change of any individual's overall ranking data. Therefore, the classic $\epsilon$-DP basically protects an individual's overall preference.

If we apply the RR mechanism $\mathcal{A}$ independently to each user's paiwise rankings, then (\ref{Anq1}) can be written as
\begin{align*}
    \frac{\mathbb{P}(\mathcal{M}(\mathcal{D}) \in \mathcal{S})}{
    \mathbb{P}(\mathcal{M}(\mathcal{D}')\in \mathcal{S})
    } =
    \frac{\prod_{l=1, l \neq l_0}^L \mathbb{P}(\mathcal{A}(\bm{y}^{(l)})) \mathbb{P}(\mathcal{A}(\bm{y}^{(l_0)}))}{\prod_{l=1,l \neq l_0}^L \mathbb{P}(\mathcal{A}(\bm{y}^{(l)}))\mathbb{P}(\mathcal{A}(\bm{y}^{(l_0)\prime}))}=
    \frac{\mathbb{P}(\mathcal{A}(\bm{y}^{(l_0)}))}{\mathbb{P}(\mathcal{A}(\bm{y}^{(l_0)\prime}))}.
\end{align*}
Given the independence $y_{ij}^{(l)}$'s and that the RR mechanism satisfies $\epsilon_l$-local DP for user $l$, we have
\begin{align*}
    \frac{\mathbb{P}(\mathcal{A}(\bm{y}^{(l_0)}))}{\mathbb{P}(\mathcal{A}(\bm{y}^{(l_0)\prime}))} = 
    \prod_{i<j,a_{ij}^{(l_0)}=1}\frac{\mathbb{P}(\mathcal{A}(y^{(l_0)}_{ij}))}{\mathbb{P}(\mathcal{A}(y^{(l_0)\prime}_{ij}))} \leq e^{S_{l_0} \epsilon_{l_0}},
\end{align*}
where $S_{l_0} = \sum_{i<j}a_{ij}^{(l_0)}$ denotes the total number of pairwise comparison collected from user $l_0$ and $a_{ij}^{(l_0)}$ indicates whether $y^{(l_0)}_{ij}$ is observed or not. Clearly, if we find an $\epsilon$ such that $S_l \epsilon_l \leq \epsilon$, then the RR mechanism satisfies the classic $\epsilon$-DP on the ranking data, thereby protecting an individual's overall preferences.

\vspace{5mm}

\section{Additional Experimental Results}
\label{AR:Experiments}

\begin{figure}[h]
    \centering
            \includegraphics[scale=0.475]{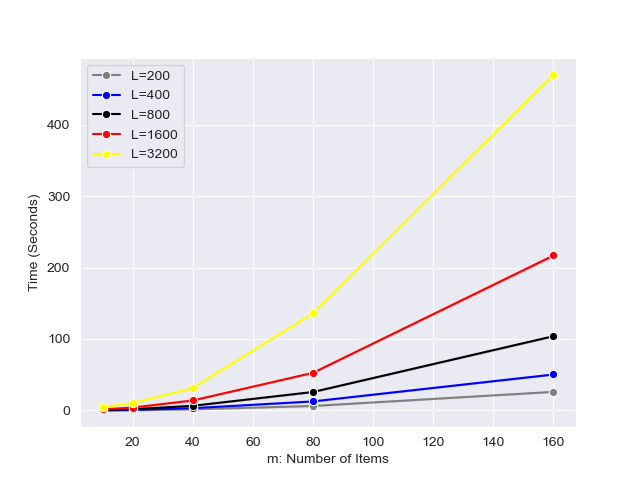}
            \includegraphics[scale=0.475]{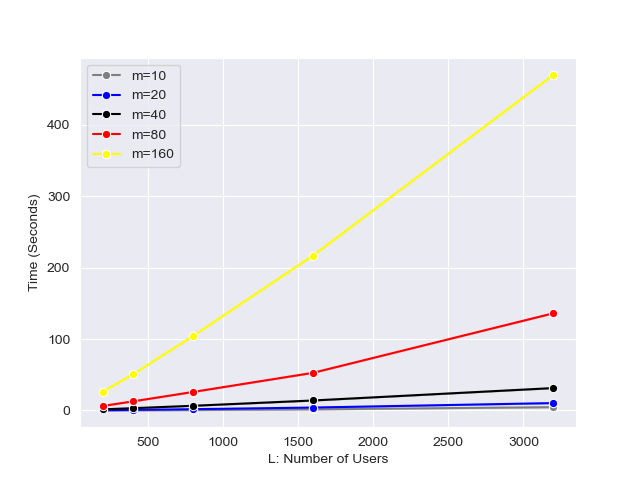}
        \captionof{figure}{The averaged computational times (in seconds) for different values of $(L,m)$.}
        \label{fig:Scalability}
    \end{figure}

\vspace{5mm}
In this section, we provide additional experimental results. Specifically, we include tables summarizing the averaged estimation errors or ranking errors across all scenarios. Before we present the additional results of simulation and the real application,  we conduct experiments to study the scalability of the our framework. We consider cases with $L \in \{100 \times 2^i, i\in [5]\}$ and $m \in \{10,20,40,80,160\}$, and we report the averaged computational time (Seconds) of 30 replicates in boxplots in Figure~\ref{fig:Scalability}. 
    
Clearly, as $L$ doubles, the average computational time of the proposed method roughly doubles for different values of $m$. Similarly, when $m$ doubles, the average computational time approximately quadruples. This observation aligns with our theoretical computational complexity result of $O(m^2L)$. Given a fixed number of items, the computational time appears to be linear with respect to the number of users. In this sense, the proposed method demonstrates scalability.

\begin{table}[h!]
    \centering
    \caption{The averaged estimation errors $m^{-\frac{1}{2}}\Vert\widehat{\bm{\theta}}-\bm{\theta}^\star\Vert_2$ and $\Vert\widehat{\bm{\theta}}-\bm{\theta}^\star\Vert_{\infty}$ under the BTL and the TM models in Scenario I.}
\scriptsize
    \begin{tabular}{l|l|cccc}
        \toprule
    Cases &   $M$ & $L=100$ & $L=200$ & $L=300$ & $L=400$   \\
        \midrule
        \multirow{3}{*}{BTL: $\frac{1}{\sqrt{m}}\Vert\widehat{\bm{\theta}}-\bm{\theta}^\star\Vert_2$}  
&10 & 0.1104 (0.0301) & 0.0770 (0.0193) & 0.0633 (0.0164) &0.0552 (0.0135)   \\
 & 20 & 0.0796 (0.0128) & 0.0551 (0.0091) & 0.0450 (0.0074) & 0.0393 (0.0066)  \\
  &30 & 0.0653 (0.0086) & 0.0455 (0.0058) & 0.0373 (0.0048) & 0.0324 (0.0042)  \\
          \midrule
                \multirow{3}{*}{TM: $\frac{1}{\sqrt{m}}\Vert\widehat{\bm{\theta}}-\bm{\theta}^\star\Vert_2$}  
&10 & 0.0779 (0.0203) &0.0541 (0.0133) &  0.0439 (0.0109) &0.0383 (0.0096)  \\
& 20 &0.0546 (0.0091) &0.0376 (0.0064) & 0.0308 (0.0052)  &0.0267 (0.0044)  \\
 &30 &0.0441 (0.0061) &0.0308 (0.0042) & 0.0252 (0.0033) &0.0219 (0.0026)  \\
       \midrule
\multirow{3}{*}{BTL: $\Vert\widehat{\bm{\theta}}-\bm{\theta}^\star\Vert_{\infty}$}  
  &     10 & 0.2158 (0.0738) & 0.1506 (0.0466) & 0.1226 (0.0363)&0.1052 (0.0294)  \\
  &20 & 0.1736 (0.0375) & 0.1204 (0.0266) &0.0995 (0.0212)  &0.0864 (0.0201)  \\
  &30 & 0.1548 (0.0329) & 0.1064 (0.0186) & 0.0869 (0.0161)&0.0755 (0.0144)  \\
   \midrule
\multirow{3}{*}{TM: $\Vert\widehat{\bm{\theta}}-\bm{\theta}^\star\Vert_{\infty}$} 
& 10&0.1514 (0.0515)&0.1054 (0.0319)&  0.0846 (0.0276 &0.0731 (0.0212) \\
&  20&0.1209 (0.0282)&0.0823 (0.0195)&  0.0678 (0.0166) &0.0591 (0.0138) \\
&  30&0.1056 (0.0226)&0.0726 (0.0147)& 0.0594 (0.0123) &0.0516 (0.0100)  \\
 \bottomrule
    \end{tabular}
    \label{tab:Au1}
\end{table}

\begin{figure}[h!]
        \centering
        \begin{subfigure}[b]{0.495\textwidth}  
            \centering 
            \includegraphics[width=\textwidth]{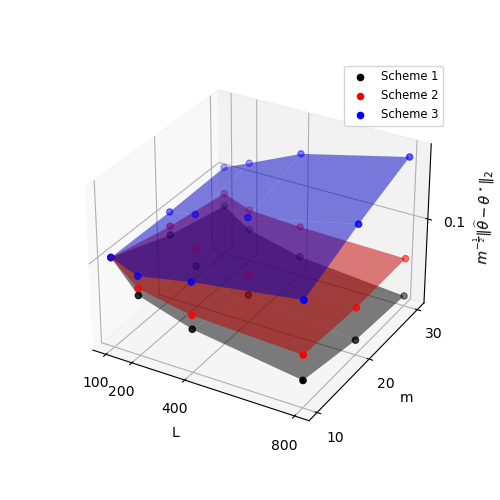}   
        \end{subfigure}
                        \begin{subfigure}[b]{0.495\textwidth}  
            \centering 
            \includegraphics[width=\textwidth]{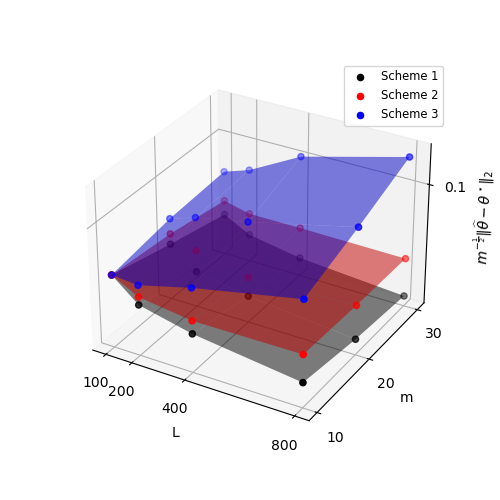}  
        \end{subfigure}
        \caption{The averaged estimation errors measured by $m^{-\frac{1}{2}}\Vert \widehat{\bm{\theta}} - \bm{\theta}^* \Vert_{2}$ in three privacy adaptive schemes in Scenario III under the BTL model (Left) and the TM model (Right).}
        \label{fig:S3}
    \end{figure}

\begin{table}[h!]
    \centering
    \caption{The averaged estimation errors $m^{-\frac{1}{2}}\Vert\widehat{\bm{\theta}}-\bm{\theta}^\star\Vert_2$ under the BTL model and the TM model in Scenario III.}
\scriptsize
    \begin{tabular}{l|l|cccc}
        \toprule
    Cases &   $M$ & $L=100$ & $L=200$ & $L=400$ & $L=800$   \\
        \midrule
        \multirow{3}{*}{BTL - Scheme 1: $\frac{1}{\sqrt{m}}\Vert\widehat{\bm{\theta}}-\bm{\theta}^\star\Vert_2$}  
&10 &   0.1028 (0.0246)&0.0801 (0.0179)&0.0670 (0.0142)&0.0546 (0.0133) \\
 & 20 & 0.0843 (0.0157)&0.0656 (0.0109)&0.0551 (0.0089)  &  0.0450 (0.0067)\\
  &30 &  0.0726 (0.0093)&0.0589 (0.0075)&0.0488 (0.0063)& 0.0414 (0.0051) \\
          \midrule
                \multirow{3}{*}{BTL - Scheme 2: $\frac{1}{\sqrt{m}}\Vert\widehat{\bm{\theta}}-\bm{\theta}^\star\Vert_2$}  
&10 & 0.1028 (0.0246)&0.0853 (0.0182)&0.0774 (0.0168)& 0.0737 (0.0177)\\
& 20 & 0.0911 (0.0168)&0.0795 (0.0128)&0.0705 (0.0113)& 0.0703 (0.0120) \\
 &30 &  0.0827 (0.0101)&0.0750 (0.0097)&0.0728 (0.0095) &0.0712 (0.0100)\\
       \midrule
\multirow{3}{*}{BTL - Scheme 3: $\frac{1}{\sqrt{m}}\Vert\widehat{\bm{\theta}}-\bm{\theta}^\star\Vert_2$}  
  &     10 &  0.1028 (0.0246)&0.0948 (0.0216)&0.1026 (0.0234)& 0.1142 (0.0289)\\
  &20 &0.1023 (0.0193)&0.1060 (0.0174)&0.1151 (0.0180)& 0.1332 (0.0228)  \\
  &30 & 0.1035 (0.0131)&0.1122 (0.0143)&0.1303 (0.0168) &0.1492 (0.0232)  \\
          \midrule
        \multirow{3}{*}{TM - Scheme 1: $\frac{1}{\sqrt{m}}\Vert\widehat{\bm{\theta}}-\bm{\theta}^\star\Vert_2$}  
&10 &  0.0699 (0.0170)&0.0562 (0.0135)&0.0486 (0.0115)  &0.0400 (0.0102)\\
 & 20 &  0.0597 (0.0105)&0.0468 (0.0080)&0.0410 (0.0073) &0.0341 (0.0059)\\
  &30 & 0.0503 (0.0064)&0.0416 (0.0052)&0.0358 (0.0050)& 0.0306 (0.0041)  \\
          \midrule
                \multirow{3}{*}{TM - Scheme 2: $\frac{1}{\sqrt{m}}\Vert\widehat{\bm{\theta}}-\bm{\theta}^\star\Vert_2$}  
&10 & 0.0699 (0.0170)&0.0612 (0.0152)&0.0567 (0.0128) & 0.0573 (0.0150)\\
& 20 &  0.0663 (0.0116)&0.0604 (0.0101)&0.0531 (0.0097)&0.0556 (0.0093) \\
 &30 & 0.0595 (0.0073)&0.0555 (0.0078)&0.0555 (0.0076)& 0.0549 (0.0072)\\
       \midrule
\multirow{3}{*}{TM - Scheme 3: $\frac{1}{\sqrt{m}}\Vert\widehat{\bm{\theta}}-\bm{\theta}^\star\Vert_2$}  
  &     10 & 0.0699 (0.0170)&0.0685 (0.0169)&0.0770 (0.0196)&  0.0906 (0.0228)\\
  &20 &  0.0760 (0.0139)&0.0814 (0.0144)&0.0880 (0.0136) & 0.1037 (0.0205) \\
  &30 &  0.0785 (0.0115)&0.0840 (0.0114)&0.1014 (0.0150) &0.1186 (0.0192)\\
 \bottomrule
    \end{tabular}
    \label{tab:Au3}
\end{table}

\begin{table}[h!]
    \centering
    \caption{The averaged estimation error $\Vert\widehat{\bm{\theta}}-\bm{\theta}^\star\Vert_{\infty}$ under the BTL model and the TM model in Scenario III.}
\scriptsize
    \begin{tabular}{l|l|cccc}
        \toprule
    Cases &   $M$ & $L=100$ & $L=200$ & $L=400$ & $L=800$   \\
        \midrule
        \multirow{3}{*}{BTL - Scheme 1: $\Vert\widehat{\bm{\theta}}-\bm{\theta}^\star\Vert_{\infty}$}  
&10 &  0.2018 (0.0586)&0.1548 (0.0416)&0.1277 (0.0324)&0.1067 (0.0304)  \\
 & 20 & 0.1865 (0.0469)&0.1435 (0.0318)&0.1224 (0.0292)&0.0990 (0.0207)   \\
  &30 &  0.1733 (0.0351)&0.1380 (0.0286)&0.1125 (0.0219) &0.0988 (0.0188)  \\
          \midrule
                \multirow{3}{*}{BTL - Scheme 2: $\Vert\widehat{\bm{\theta}}-\bm{\theta}^\star\Vert_{\infty}$}  
&10 &  0.2018 (0.0586)&0.1656 (0.0445)&0.1478 (0.0369)&0.1416 (0.0390)\\
& 20 & 0.2017 (0.0486)&0.1747 (0.0379)&0.1559 (0.0344)&0.1573 (0.0373)  \\
 &30 & 0.1978 (0.0398)&0.1770 (0.0365)&0.1714 (0.0329)&0.1653 (0.0335) \\
       \midrule
\multirow{3}{*}{BTL - Scheme 3: $\Vert\widehat{\bm{\theta}}-\bm{\theta}^\star\Vert_{\infty}$}  
  &     10 &  0.2018 (0.0586)&0.1818 (0.0498)&0.1977 (0.0503)&0.2179 (0.0666)\\
  &20 & 0.2279 (0.0571)&0.2349 (0.0555)&0.2533 (0.0560)&0.2875 (0.0671) \\
  &30 & 0.2486 (0.0496)&0.2649 (0.0566)&0.3111 (0.0628)&0.3525 (0.0761)  \\
          \midrule
        \multirow{3}{*}{TM - Scheme 1: $\Vert\widehat{\bm{\theta}}-\bm{\theta}^\star\Vert_{\infty}$}  
&10 &  0.1348 (0.0419)&0.1093 (0.0315)&0.0924 (0.0268) &0.0792 (0.0243) \\
 & 20 & 0.1320 (0.0350)&0.1045 (0.0245)&0.0936 (0.0240) &0.0760 (0.0183)  \\
  &30 & 0.1198 (0.0251)&0.0980 (0.0194)&0.0852 (0.0172)&0.0733 (0.0152)  \\
          \midrule
                \multirow{3}{*}{TM - Scheme 2: $\Vert\widehat{\bm{\theta}}-\bm{\theta}^\star\Vert_{\infty}$}  
&10 & 0.1348 (0.0419)&0.1198 (0.0360)&0.1091 (0.0304)& 0.1105 (0.0346)  \\
& 20 & 0.1468 (0.0370)&0.1373 (0.0336)&0.1161 (0.0275)&0.1253 (0.0288)   \\
 &30 &  0.1426 (0.0289)&0.1358 (0.0295)&0.1335 (0.0261)&0.1331 (0.0307) \\
       \midrule
\multirow{3}{*}{TM - Scheme 3: $\Vert\widehat{\bm{\theta}}-\bm{\theta}^\star\Vert_{\infty}$}  
  &     10 & 0.1348 (0.0419)&0.1323 (0.0419)&0.1474 (0.0437)&0.1754 (0.0530) \\
  &20 & 0.1694 (0.0445)&0.1826 (0.0458)&0.1952 (0.0423)&0.2351 (0.0672)  \\
  &30 &  0.1913 (0.0424)&0.2047 (0.0462)&0.2525 (0.0616& 0.2974 (0.0850) \\
 \bottomrule
    \end{tabular}
    \label{tab:Au4}
\end{table}

\begin{figure}[ht!]
        \centering
                \begin{subfigure}[b]{0.43\textwidth}   
            \centering 
            \includegraphics[width=\textwidth]{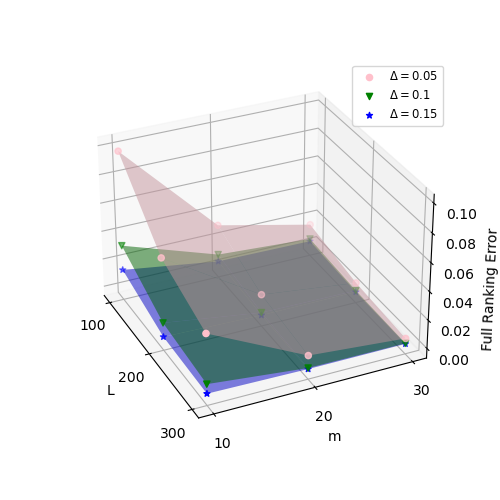}
        \end{subfigure}
        \begin{subfigure}[b]{0.43\textwidth}   
            \centering 
            \includegraphics[width=\textwidth]{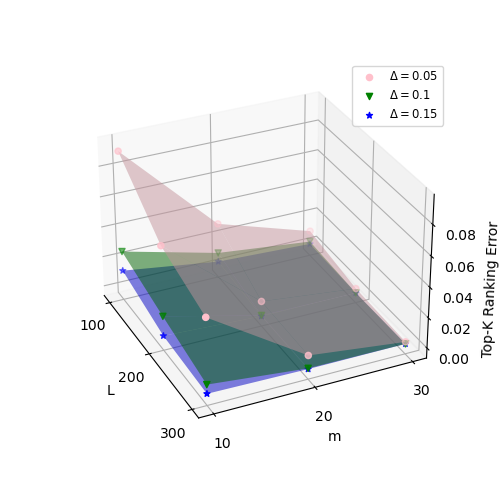}
        \end{subfigure}
        \caption{The averaged full ranking error (Left) and top-$K$ ranking error (Right) of all cases in Scenario IV under the TM model.}
        \label{fig:AUPG4}
    \end{figure}

\begin{table}[h!]
    \centering
    \caption{The averaged full ranking errors under the BTL and the TM models in Scenario IV.}
\scriptsize
    \begin{tabular}{l|l|cccccc}
        \toprule
    $\Delta$ &   $m$ & $L=100,BTL$ & $L=200,BTL$ & $L=300,BTL$ & $L=100,TM$  & $L=200,TM$ & $L=300,TM$  \\
        \midrule
0.05 &  10 & 0.1673 (0.0606) & 0.1144 (0.0484) & 0.0896 (0.0395) & 0.0987 (0.0458) & 0.0582 (0.0323) & 0.0429 (0.0280)  \\
0.1  &  10 & 0.0736 (0.0353) & 0.0416 (0.0267) & 0.0287 (0.0240) & 0.0323 (0.0236) & 0.0128 (0.0153) & 0.0081 (0.0136)  \\
0.15 &  10 & 0.0373 (0.0255) & 0.0167 (0.0175) & 0.0106 (0.0154) & 0.0148 (0.0156) & 0.0028 (0.0074) & 0.0014 (0.0055)  \\
          \midrule 
       0.05 &  20 & 0.0575 (0.0178) & 0.0357 (0.0128) & 0.0257 (0.0101) & 0.0313 (0.0126) & 0.0159 (0.0085) & 0.0099 (0.0063) \\

       0.1  &  20 & 0.0229 (0.0099) & 0.0111 (0.0070) & 0.0059 (0.0048) & 0.0098 (0.0067) & 0.0029 (0.0042) & 0.0011 (0.0024) \\

       0.15 &  20 & 0.0120 (0.0077) & 0.0041 (0.0048) & 0.0018 (0.0030) & 0.0048 (0.0049) & 0.0010 (0.0025) & 0.0003 (0.0013) \\
       \midrule
       0.05 &  30 & 0.0312 (0.0081) & 0.0184 (0.0056) & 0.0130 (0.0048) & 0.0162 (0.0053) & 0.0077 (0.0032) & 0.0042 (0.0029) \\
       0.1  &  30 & 0.0122 (0.0047) & 0.0051 (0.0027) & 0.0027 (0.0024) & 0.0055 (0.0035) & 0.0020 (0.0020) & 0.0007 (0.0013) \\
       0.15 &  30 & 0.0062 (0.0035) & 0.0022 (0.0021) & 0.0008 (0.0014) & 0.0038 (0.0029) & 0.0010 (0.0014) & 0.0003 (0.0008) \\
 \bottomrule
    \end{tabular}
    \label{tab:Au4_fke}
\end{table}

\begin{table}[h!]
    \centering
    \caption{The averaged top-$K$ ranking errors under the BTL and TM models in Scenario IV.}
\scriptsize
    \begin{tabular}{l|l|cccccc}
        \toprule
    $\Delta$ &   $m$ & $L=100,BTL$ & $L=200,BTL$ & $L=300,BTL$ & $L=100,TM$  & $L=200,TM$ & $L=300,TM$  \\
        \midrule
       0.05 &  10 & 0.1520 (0.1207) & 0.1120 (0.1163) & 0.0860 (0.1032) & 0.0920 (0.1058) & 0.0620 (0.0949) & 0.0500 (0.0868) \\
       0.1  &  10 & 0.0690 (0.0994) & 0.0490 (0.0885) & 0.0370 (0.0779) & 0.0260 (0.0674) & 0.0160 (0.0544) & 0.0070 (0.0368) \\
       0.15 &  10 & 0.0360 (0.0770) & 0.0240 (0.0652) & 0.0110 (0.0457) & 0.0130 (0.0494) & 0.0030 (0.0244) & 0.0010 (0.0141) \\
        \midrule
       0.05 &  20 & 0.0535 (0.0539) & 0.0315 (0.0466) & 0.0250 (0.0434) & 0.0300 (0.0459) & 0.0100 (0.0301) & 0.0090 (0.0287) \\
       0.1  &  20 & 0.0220 (0.0415) & 0.0090 (0.0287) & 0.0045 (0.0208) & 0.0100 (0.0301) & 0.0005 (0.0071) & 0.0005 (0.0071) \\
       0.15 &  20 & 0.0090 (0.0287) & 0.0010 (0.0100) & 0.0005 (0.0071) & 0.0040 (0.0196) & 0.0000 (0.0000) & 0.0000 (0.0000) \\
        \midrule
       0.05 &  30 & 0.0240 (0.0328) & 0.0157 (0.0283) & 0.0120 (0.0257) & 0.0103 (0.0242) & 0.0030 (0.0139) & 0.0013 (0.0094) \\
       0.1  &  30 & 0.0063 (0.0196) & 0.0023 (0.0123) & 0.0003 (0.0047) & 0.0030 (0.0139) & 0.0000 (0.0000) & 0.0000 (0.0000) \\
       0.15 &  30 & 0.0033 (0.0146) & 0.0000 (0.0000) & 0.0000 (0.0000) & 0.0013 (0.0094) & 0.0000 (0.0000) & 0.0000 (0.0000) \\
 \bottomrule
    \end{tabular}
    \label{tab:Au4_tke}
\end{table}


\begin{figure}[h!]
        \centering
        \begin{subfigure}[b]{0.495\textwidth}  
            \centering 
            \includegraphics[width=\textwidth]{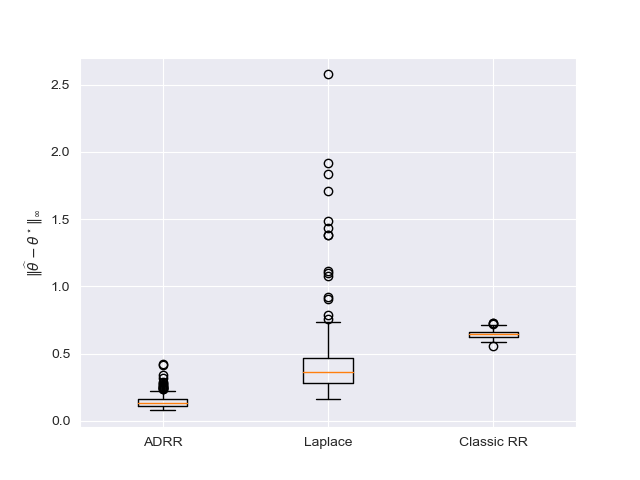}
        \end{subfigure}
        \begin{subfigure}[b]{0.495\textwidth}   
            \centering 
            \includegraphics[width=\textwidth]{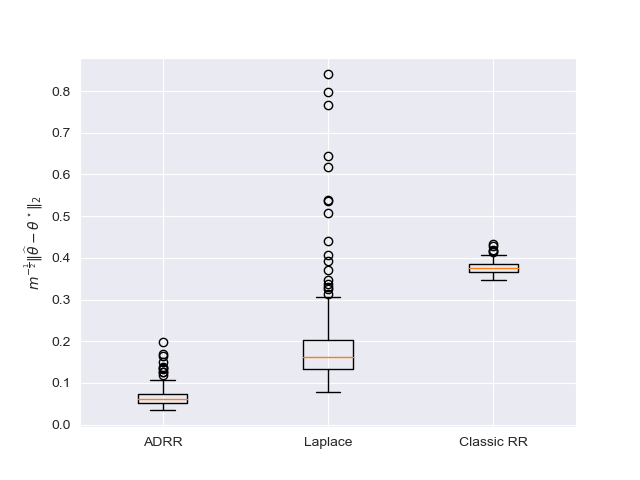}
        \end{subfigure}
        \caption{Comparison of the proposed ADRR method and others under the TM model regarding $\Vert \widehat{\bm{\theta}} - \bm{\theta}^\star\Vert_{\infty}$ (Left) and $m^{-\frac{1}{2}}\Vert \widehat{\bm{\theta}} - \bm{\theta}^\star\Vert_{2}$ (right).}
        \label{fig:AUPG5}
    \end{figure}

\clearpage
    \begin{table}[ht]
    \centering
    \caption{The averaged estimation errors under the BTL and the TM models in Scenario V.}
\scriptsize
    \begin{tabular}{l|c|cccc}
        \toprule
Model & Estimation Error & ADRR & NoisyObj &Laplace & Classic RR \\
        \midrule
        BTL    &$m^{-\frac{1}{2}}\Vert \widehat{\bm{\theta}}-\bm{\theta}\Vert_{2}$     &0.0882 (0.0019)&  0.1327 (0.0045)&0.2221 (0.0057)&0.3545 (0.0012) \\
 BTL    &$\Vert \widehat{\bm{\theta}}-\bm{\theta}\Vert_{\infty}$&0.1889 (0.0042)& 0.2581 (0.0080)&0.4852 (0.0149)&0.6107 (0.0025)  \\
 TM     &$m^{-\frac{1}{2}}\Vert \widehat{\bm{\theta}}-\bm{\theta}\Vert_{2}$     &0.0670 (0.0019)& -&0.1930 (0.0081)&0.3770 (0.0010) \\
TM     &$\Vert \widehat{\bm{\theta}}-\bm{\theta}\Vert_{\infty}$ &0.1478 (0.0038)& - &0.4440 (0.0222)&0.6437 (0.0018) \\
 \bottomrule
    \end{tabular}
    \label{tab:Au5_T1}
\end{table}

    \begin{table}[h!]
    \centering
    \caption{The averaged full ranking error under the BTL and the TM models in Scenario V.}
\scriptsize
    \begin{tabular}{l|l|cc}
        \toprule
Model & Error & ADRR &  Count Method  \\
        \midrule
 BTL    &Full Ranking Error&0.0221 (0.0009)&0.0374 (0.0013)
 \\
 TM     &Full Ranking Error&0.0112 (0.0006)&0.0206 (0.0009) \\
 \bottomrule
    \end{tabular}
    \label{tab:Au5_T2}
\end{table}

\begin{figure}[h!]
        \centering
        \begin{subfigure}[b]{0.495\textwidth}  
            \centering 
            \includegraphics[width=\textwidth]{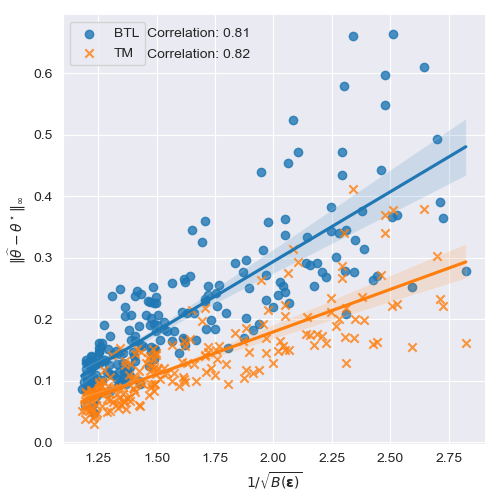}
        \end{subfigure}
        \begin{subfigure}[b]{0.495\textwidth}   
            \centering 
            \includegraphics[width=\textwidth]{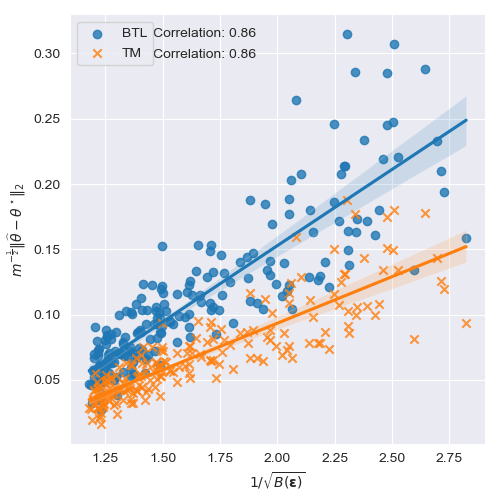}
        \end{subfigure}

        \caption{The linear models between estimation errors and $(B(\bm{\epsilon}))^{-\frac{1}{2}}$ in real application. Here, the dots represent the estimation errors of 200 replicates, and the shaded areas represent the corresponding 95\% prediction intervals.}
        \label{Real_Comp33}
    \end{figure}

\clearpage

\section{Proof}
\label{SuppSec:Thm}
In this section, we provide the proofs for all theoretical results presented in this paper. First, we introduce the necessary notations for the proofs of Lemmas and Theorems in Section \ref{SecLNot}. Next, we present the proofs of all theorems in Section \ref{Sec:ProofThm}. Finally, we present the proofs of all lemmas in Section \ref{SuppSec:Lemmas}.

\subsection{Notations}
\label{SecLNot}

\begin{itemize}
    \item $F(x)$: a cumulative distribution function for a symmetric continuous random variable;
    \item $f(x)$: the derivative of $F(x)$, i.e., $F'(x)=f(x)$;
    \item $g(x)$: the derivative of $\log F(x)$, i.e., $g(x)=f(x)/F(x)$;
    \item $g'(x)$: the derivative of $g(x)$
    \item $C_{F,L}(q)$ and $C_{F,U}(q)$ represent the minimum and maximum of $F(x)$ on the interval $[-\kappa-2q,\kappa+2q]$;
    \item $C_{f,L}(q)$ and $C_{f,U}(q)$ represent the minimum and maximum of $f(x)$ on the interval $[-\kappa-2q,\kappa+2q]$;
    \item $C_{g,L}(q)$ and $C_{g,U}(q)$ represent the minimum and maximum of $g(x)$ on the interval $[-\kappa-2q,\kappa+2q]$;
    \item $C_{g',L}(q)$ and $C_{g',U}(q)$ represent the minimum and maximum of $-g'(x)$ on the interval $[-\kappa-2q,\kappa+2q]$;
    \item When $q=0$, $C_{g,L}(0)$ is abbreviated as $C_{g,L}$ and we apply similar conventions to other notations;
    \item $G(\bm{\epsilon})=\sum_{l=1}^L \left(\frac{e^{\epsilon_l}-1}{e^{\epsilon_l}+1}\right)^2$ and $B(\bm{\epsilon})=\frac{G(\bm{\epsilon})}{L} = \frac{1}{L}\sum_{l=1}^L \left(\frac{e^{\epsilon_l}-1}{e^{\epsilon_l}+1} \right)^2$
\end{itemize}

\subsection{Proof of All Theorems}
\label{Sec:ProofThm}

\begin{theorem}
\label{Thm:ConsisT1}
Conditional on the event $\mathcal{E}(q) = \{\Vert \widehat{\bm{\theta}} - \bm{\theta}^\star \Vert_{\infty} \leq q\}$, under Assumptions \ref{Ass1}--\ref{Ass2}, and choosing $\lambda \leq \frac{m}{\kappa}\delta_{m,L}$, the following holds for any $\delta_{m,L} \gtrsim \frac{C_{g,U}^2(q)}{C_{f,L}^2(q)} \frac{\log(m)}{mLB(\bm{\epsilon})}$:
\begin{align}
\label{Conver:ER}
\mathbb{P}\left(
e(\widehat{\bm\theta},\bm\theta^\star) \ge \delta_{m,L}
\right) \leq 2
\exp\left(-
\frac{Cm^2 LB(\bm{\epsilon})   \delta_{m,L}}{R_F^2(q)}
\right),
\end{align}
for some positive constant $C$ depending on $F(\cdot)$, $\kappa$, and $q$, where the excess risk $e(\bm\theta,\bm\theta^\star)$ is defined as
\begin{align*}
    e(\bm\theta,\bm\theta^\star)= &
    \frac{1}{m^2}\sum_{i<j,l}
    \mathbb{E}\big(\mathcal{L}_{ij}(\bm\theta,\bm{z}^{(l)})-\mathcal{L}_{ij}(\bm\theta^\star,\bm{z}^{(l)})\big) \\
    = &
    \frac{1}{m^2}\sum_{i<j}
    p\left[F(\theta_i^\star-\theta_j^\star) \log \left(\frac{F(\theta_i^\star-\theta_j^\star)}{F(\theta_i-\theta_j)}\right)+(1-F(\theta_i^\star-\theta_j^\star))\log\left(\frac{1-F(\theta_i^\star-\theta_j^\star)}{1-F(\theta_i-\theta_j)}\right)\right].
\end{align*}
\end{theorem}

\noindent\textbf{Proof of Theorem \ref{Thm:ConsisT1}.}  Note that $\mathbb{E}(\widetilde{z}_{ij}^{(l)})=\mathbb{E}(y_{ij}^{(l)})$ for each $i<j$ and $l \in [L]$, hence it holds true that
\begin{align*}
&\sum_{l=1}^L \mathbb{E}\big(\mathcal{L}_{ij}(\bm\theta,\bm{z}^{(l)})-\mathcal{L}_{ij}(\bm\theta^\star,\bm{z}^{(l)})\big) \\
 =&   p\left[F(\theta_i^\star-\theta_j^\star) \log \left(\frac{F(\theta_i^\star-\theta_j^\star)}{F(\theta_i-\theta_j)}\right)+(1-F(\theta_i^\star-\theta_j^\star))\log\left(\frac{1-F(\theta_i^\star-\theta_j^\star)}{1-F(\theta_i-\theta_j)}\right)\right] \geq 0
\end{align*}
where the inequality follows from Jensen's inequality. Moreover, we have $\mathbb{E}\big(\mathcal{L}_{ij}(\bm\theta,\bm{z}^{(l)})-\mathcal{L}_{ij}(\bm\theta^\star,\bm{z}^{(l)})\big)=0$ if and only if $\theta_i-\theta_j=\theta_i^\star-\theta_j^\star$. It can be verified that $e(\bm\theta,\bm\theta^\star)\ge 0$ and the inequality holds if and only if $\bm\theta=\bm\theta^\star$ due to the identifiability condition $\bm{1}^T\bm\theta^\star=0$. 

 Conditional on $\mathcal{E}(q)$, we can ensure $\widehat{\bm{\theta}} =
\argmin_{\bm{\theta} \in \mathbb{R}^m} \mathcal{L}_{\lambda}(\bm{\theta})=
\argmin_{\bm{\theta} \in \bm\Theta_0} \mathcal{L}_{\lambda}(\bm{\theta})$, where $\bm\Theta_0 =\{\bm{\theta} \in \mathbb{R}^m: \bm{1}_m^T \bm{\theta}=0, \bm{\theta} \in \mathcal{C}(q)\}$ and $\mathcal{C}(q)=\{\bm{\theta}: \Vert \bm{\theta} - \bm{\theta}^\star\Vert_{\infty} \leq q \}$. Here it should be noted that if $\bm{\theta} \in \mathcal{C}(q)$, we have $$|\gamma_{ij}|=|\theta_i-\theta_j| \leq |\theta_i-\theta_j^\star|+|\theta_j-\theta_j^\star|+|\theta_j^\star-\theta_i^\star| \leq 2q+\kappa.
$$

To establish the large deviation inequality for $e(\widehat{\bm\theta},\bm\theta^\star)$, we first observe that
\begin{align*}
\Big\{
e(\widehat{\bm\theta},\bm\theta^\star) \geq \delta_{m,L}
\Big \} \subset 
\left \{
\sup_{\{\bm\theta\in\bm\Theta_0:\  e(\bm\theta,\bm\theta^\star)\ge\delta_{m,L}\}}\big(\mathcal{L}_{\lambda}(\bm\theta^\star)-\mathcal{L}_{\lambda}(\bm\theta)\big)\ge 0
\right \}.
\end{align*}
This implies that
\begin{align*}
\mathbb{P}\left(
e(\widehat{\bm\theta},\bm\theta^\star) \geq \delta_{m,L}
\right) \leq 
\mathbb{P}\left(\sup_{\{\bm\theta\in\bm\Theta_0:\  e(\bm\theta,\bm\theta^\star)\ge\delta_{m,L}\}}\big(\mathcal{L}_{\lambda}(\bm\theta^\star)-\mathcal{L}_{\lambda}(\bm\theta)\big)\ge 0\right).
\end{align*}

For each $u\ge 1$, let $S_{u}=\{\bm\theta\in\bm\Theta_0:\ 2^{u-1}\delta_{m,L}\le e(\bm\theta,\bm\theta^\star)\le 2^{u}\delta_{m,L}\}$, then we have $\{\bm\theta\in\bm\Theta_0:\  e(\bm\theta,\bm\theta^\star)\ge\delta_{m,L}\}=\bigcup_{u=1}^{\infty}S_u$. Therefore,
\begin{align*}
&\ \mathbb{P}\left(\sup_{\{\bm\theta\in\bm\Theta_0:\  e(\bm\theta,\bm\theta^\star)\ge\delta_{m,L}\}}\big(\mathcal{L}_{\lambda}(\bm\theta^\star)-\mathcal{L}_{\lambda}(\bm\theta)\big)\ge 0\right)\\
=&\ \mathbb{P}\left(\sup_{\{\bm\theta\in\bm\Theta_0:\  e(\bm\theta,\bm\theta^\star)\ge\delta_{m,L}\}}\frac{1}{m^2}\sum_{i<j}\sum_{l=1}^L\big(\mathcal{L}_{ij}(\bm\theta^\star,\bm{z}^{(l)})-\mathcal{L}_{ij}(\bm\theta,\bm{z}^{(l)})\big)+\frac{\lambda}{m^2}\big(J(\bm\theta^\star)-J(\bm\theta)\big)\ge 0\right)\\
\le&\  \sum_{u=1}^{\infty}\mathbb{P}\left(\sup_{\bm{\theta} \in S_u}\frac{1}{m^2}\sum_{i<j}\sum_{l=1}^L\big(\mathcal{L}_{ij}(\bm\theta^\star,\bm{z}^{(l)})-\mathcal{L}_{ij}(\bm\theta,\bm{z}^{(l)})\big)+\frac{\lambda}{m^2}\big(J(\bm\theta^\star)-J(\bm\theta)\big)\ge 0\right)\triangleq \sum_{u=1}^{\infty}I_u,
\end{align*}
where $J(\bm{\theta}) = \Vert\bm{\theta} \Vert_2^2$. 

Then, we can bound each $I_u$ separately. Let $\nu(\bm\theta,\bm{z})=\frac{1}{m^2}\sum_{i<j}\sum_{l=1}^L\big(\mathcal{L}_{ij}(\bm\theta^\star,\bm{z}^{(l)})-\mathcal{L}_{ij}(\bm\theta,\bm{z}^{(l)})\big)-\mathbb{E}\big(\mathcal{L}_{ij}(\bm\theta^\star,\bm{z}^{(l)})-\mathcal{L}_{ij}(\bm\theta,\bm{z}^{(l)})\big)$. Then we have
\begin{align*}
I_{u}&=\mathbb{P}\left(\sup_{\bm{\theta} \in S_u}\frac{1}{m^2}\sum_{i<j,l}\big(\mathcal{L}_{ij}(\bm\theta^\star,\bm{z}^{(l)})-\mathcal{L}_{ij}(\bm\theta,\bm{z}^{(l)})\big)+\frac{\lambda}{m^2}\big(J(\bm\theta^\star)-J(\bm\theta)\big)\ge 0\right)\\
&\le \mathbb{P}\left(\sup_{\bm{\theta} \in S_u}\nu(\bm\theta,\bm{z})\ge \inf_{\bm{\theta} \in S_u}\big(e(\bm\theta,\bm\theta^\star)+\frac{\lambda}{m^2}J(\bm\theta)-\frac{\lambda}{m^2}J(\bm\theta^\star)\big)\right).
\end{align*}
Following from the definition of $S_{u}$, we have
$$ \inf_{\bm{\theta} \in S_u}\left(e(\bm\theta,\bm\theta^\star)+\frac{\lambda}{m^2}J(\bm\theta)-\frac{\lambda}{m^2}J(\bm\theta^\star)\right)\ge 2^{u-1}\delta_{m,L}-\frac{1}{2}\delta_{m,L}\ge 2^{u-2}\delta_{m,L}\triangleq M_{u},
$$
where the inequality holds by setting $\lambda/(m^2) J(\bm{\theta}^\star) \leq \frac{1}{2}\delta_{m,L}$. Moreover, it follows from Lemma \ref{lemma: Esup} that $\mathbb{E}\big(\sup_{\bm{\theta} \in S_u}\nu(\bm\theta,\bm{z})\big)\le \frac{M_{u}}{2}$. Thus, 
\begin{align}
\label{Main_INequ}
I_{u} \le \mathbb{P}\left(\sup_{\bm{\theta} \in S_u} \nu(\bm\theta,\bm{z}) \ge M_{u}\right) \le \mathbb{P}\left(\sup_{\bm{\theta} \in S_u}\nu(\bm\theta,\bm{z}) \ge \mathbb{E}\Big(\sup_{\bm{\theta} \in S_u}\nu(\bm\theta,\bm{z})\Big)+\frac{M_{u}}{2}\right).
\end{align}
By Lemma \ref{lemma:inequ}, we have
\begin{align*}
&\sup_{\bm{\theta} \in S_u}\sum_{i<j}\sum_{l=1}^L\text{Var}\left(\mathcal{L}_{ij}(\bm\theta^\star,\bm{z}^{(l)})-\mathcal{L}_{ij}(\bm\theta,\bm{z}^{(l)})\right)
\le 
\frac{2C_{g,U}^2(q)}{C_{f,L}^2(q) LB(\bm{\epsilon})}\sum_{i<j}\sum_{l=1}^L\mathbb{E}\big(\mathcal{L}_{ij}(\bm\theta,\bm{z}^{(l)})-\mathcal{L}_{ij}(\bm\theta^\star,\bm{z}^{(l)})\big) \\
= &
\frac{2m^2 C_{g,U}^2(q) }{C_{f,L}^2(q) LB(\bm{\epsilon})}
\sup_{\bm{\theta} \in S_u}
\frac{1}{m^2}
\sum_{i<j}\sum_{l=1}^L\mathbb{E}\big(\mathcal{L}_{ij}(\bm\theta,\bm{z}^{(l)})-\mathcal{L}_{ij}(\bm\theta^\star,\bm{z}^{(l)})\big)  \leq 
\frac{2m^2 C_{g,U}^2(q) 2^u \delta_{m,L}}{C_{f,L}^2(q) LB(\bm{\epsilon})} \\
=&
\frac{2m^2 R_F^2(q) 2^u \delta_{m,L}}{LB(\bm{\epsilon})}
=
\frac{8m^2 R_F^2(q) M_u}{LB(\bm{\epsilon})},
\end{align*}
where $R_F(q) = C_{g,U}(q)/C_{f,L}(q)$.

Next, we intend to use the concentration inequality to prove the convergence $I_u$. We first note that, $\nu(\bm\theta;\bm{z})$ can be written as
\begin{align*}
&\nu(\bm\theta,\bm{z})=
\frac{1}{m^2 LB(\bm{\epsilon})}
   \sum_{i<j}\sum_{l=1}^L\left[f_{ijl}(\bm\theta,\bm{z})-\mathbb{E}(f_{ijl}(\bm\theta,\bm{z})\big)\right], \\
&f_{ijl}(\bm{\theta},\bm{z}) =  a_{ij}^{(l)}\widetilde{w}_l
    \left(
    \widetilde{z}_{ij}^{(l)}
\log \frac{F(\gamma_{ij})}{F(\gamma_{ij}^\star)}
+(1-\widetilde{z}_{ij}^{(l)})
\log  \frac{F(-\gamma_{ij})}{F(-\gamma_{ij}^\star)}
    \right),
\end{align*}
where $\widetilde{w}_l = w_l G(\bm{\epsilon})=\left(\frac{e^{\epsilon_l}-1}{e^{\epsilon_l}+1}\right)^2$ and $\gamma_{ij}=\theta_i-\theta_j$. For each $i<j$ and $l\in [L]$ and any $\bm{\theta} \in \mathcal{E}(q)$,
\begin{align*}
\left|
f_{ijl}(\bm\theta,\bm{z})-\mathbb{E}(f_{ijl}(\bm\theta,\bm{z})\big)\right| 
\leq  \left(\frac{e^{\epsilon_l}-1}{e^{\epsilon_l}+1}\right)^2 C_{g,U}(q) \cdot |\gamma_{ij}-\gamma_{ij}^\star| \leq 2q C_{g,U}(q),
\end{align*}
where the last inequality follows from the facts that $\frac{e^{\epsilon_l}-1}{e^{\epsilon_l}+1}\leq 1$ and $|\gamma_{ij}-\gamma_{ij}^\star| \leq 2q$ conditional on the event $\mathcal{E}(q)$.

We let $C_3 = \frac{m^2 LB(\bm{\epsilon})}{2q C_{g,U}(q)}$ be a scaling constant for applying Theorem 1.1 of \cite{klein2005concentration} to bound (\ref{Main_INequ}).
\begin{align*}
&\mathbb{P}\left(
\sup_{\bm{\theta} \in S_u} \nu(\bm{\theta},\bm{z}) \geq \mathbb{E}
\big(
\sup_{\bm{\theta} \in S_u} \nu(\bm{\theta},\bm{z}) 
\big)+M_u/2
\right) \\
=&
\mathbb{P}\left(
\sup_{\bm{\theta} \in S_u} C_3 \nu(\bm{\theta},\bm{z}) \geq \mathbb{E}
\big(
\sup_{\bm{\theta} \in S_u} C_3 \nu(\bm{\theta},\bm{z}) 
\big)+C_3M_u/2
\right) 
\\
\leq &
\exp\left(-
\frac{m^2 LB(\bm{\epsilon})  M_u}{64 R_F(q)^2+28 q C_{g,U}(q) }
\right).
\end{align*}
To sum up, it holds that
\begin{align*}
    \sum_{u=1}^{\infty} I_u \leq  &
    \sum_{u=1}^{\infty} 
    \exp\left(-
\frac{m^2 LB(\bm{\epsilon})  2^{u-2}\delta_{m,L}}{64 R_F^2(q)+28 q C_{g,U}(q) }
\right) \leq 
\sum_{u=1}^{\infty} 
\exp\left(-
\frac{m^2 LB(\bm{\epsilon})  u \delta_{m,L}}{128 R_F^2(q)+56 q C_{g,U}(q) }
\right) \\
\leq &
\frac{\exp\left(-
\frac{m^2 LB(\bm{\epsilon})   \delta_{m,L}}{128 R_F^2(q)+56 q C_{g,U}(q) }
\right)}{1-\exp\left(-
\frac{m^2 LB(\bm{\epsilon})   \delta_{m,L}}{128 R_F^2(q)+56 q C_{g,U}(q) }
\right)} \leq 2\exp\left(-
\frac{m^2 LB(\bm{\epsilon})   \delta_{m,L}}{128 R_F^2(q)+56 q C_{g,U}(q) }
\right).
\end{align*}
where the last inequality holds asymptotically if $\delta_{m,L} \gtrsim \frac{R_F^2(q) \log(m)}{mLB(\bm{\epsilon})}$.
\qed \\

\noindent
\textbf{Proof of Theorem \ref{Thm:Estimation}.} In this theorem, we first establish a general result for any $F(x)$ that satisfies Assumptions \ref{Ass1} and \ref{Ass2}. Following this, we derive the desired result for the BTL model by substituting $F(x)$ with $1/(1+e^{-x})$ since the BTL model satisfies Assumptions \ref{Ass1} and \ref{Ass2}.

By applying a similar step as (\ref{Gasms}) in Lemma \ref{lemma:inequ}, we get
\begin{align}
\label{Eq1}
  &  p\left[ \left(\sqrt{F(\widehat{\theta}_i-\widehat{\theta}_j)}-\sqrt{F(\theta_i^\star-\theta_j^\star)}\right)^2+\left(\sqrt{1-F(\widehat{\theta}_i-\widehat{\theta}_j)}-\sqrt{1-F(\theta_i^\star-\theta_j^\star)}\right)^2
\right] \notag\\
\le & \sum_{l=1}^L \mathbb{E}\big(\mathcal{L}_{ij}(\widehat{\bm\theta},\bm{z}^{(l)})-\mathcal{L}_{ij}(\bm\theta^\star,\bm{z}^{(l)})\big),
\end{align}
 for each $i<j$ and $l\in[L]$. 

Conditional on the event $\mathcal{E}(q)= \{\Vert \widehat{\bm{\theta}} - \bm{\theta}^\star \Vert_{\infty} \leq q\}$, 
\begin{align*}
e(\widehat{\bm\theta},\bm\theta^\star)
\geq &
\frac{p}{m^2}\sum_{i<j}\left[ \left(\sqrt{F(\widehat{\theta}_i-\widehat{\theta}_j)}-\sqrt{F(\theta_i^\star-\theta_j^\star)}\right)^2+\left(\sqrt{1-F(\widehat{\theta}_i-\widehat{\theta}_j)}-\sqrt{1-F(\theta_i^\star-\theta_j^\star)}\right)^2
\right]\\
\geq &\frac{2p}{m^2}\min_{x \in [-\kappa-2q,\kappa+2q]}\frac{f^2(x)}{F(x)}
\sum_{i<j}\left(
\widehat{\theta}_i-\theta_i^\star-(\widehat{\theta}_j-
\theta_j^\star)
\right)^2 \\
=&\frac{p}{m^2}\min_{x \in [-\kappa-2q,\kappa+2q]}\frac{f^2(x)}{F(x)}\sum_{i,j\in[m],i\neq j}\left[(\widehat{\theta}_{i}-\theta_i^\star)^2+(\widehat{\theta}_{j}-\theta_j^\star)^2-2(\widehat{\theta}_{i}-\theta_i^\star)(\widehat{\theta}_{j}-\theta_j^\star)\right]\\
=&\frac{p(m-1)}{m^2}\min_{x \in [-\kappa-2q,\kappa+2q]}\frac{f^2(x)}{F(x)}\|\widehat{\bm\theta}-\bm\theta^\star\|_2^2 
\geq 
\frac{p}{2m}C_{f,L}^2(q)\|\widehat{\bm\theta}-\bm\theta^\star\|_2^2 
\end{align*}
where the last equality follows from $\bm{1}_m^T\widehat{\bm\theta}=\bm{1}_m^T\bm\theta^\star=0$. Then we can conclude that 
\begin{align}
\label{Use}
&\ \mathbb{P}\left(\frac{1}{\sqrt{m}}\|\widehat{\bm\theta}-\bm\theta^\star\|_2 
\ge 
\sqrt{2p^{-1}C_{f,L}^{-2}(q)\delta_{m,L}}\right)\le \mathbb{P}\left(e(\widehat{\bm\theta},\bm\theta^\star)\ge \delta_{m,L} \right)\notag \\
\le&\ \mathbb{P}\left(\sup_{\{\bm\theta\in\bm\Theta:\  e(\bm\theta,\bm\theta^\star)\ge\delta_{m,L}\}}\big(\mathcal{L}_{\lambda}(\bm\theta^\star)-\mathcal{L}_{\lambda}(\bm\theta)\big)\ge 0\right)
\leq 2
\exp\left(-
\frac{Cm^2 LB(\bm{\epsilon})   \delta_{m,L}}{R_F^2(q)}
\right),
\end{align}
where $C$ is as defined in Theorem \ref{Thm:ConsisT1}. The first desired result follows immediately by Lemma \ref{Lemma:Boundedness} that $\mathbb{P}(\mathcal{E}^c(q)) \lesssim m^{-8}$. Therefore, we have
\begin{align*}
\mathbb{P}\left(\frac{1}{\sqrt{m}}\|\widehat{\bm\theta}-\bm\theta^\star\|_2 
\ge 
\frac{C_{g,U}(q)}{C_{f,L}^2(q)}
\sqrt{ \frac{2\log(m)}{mLpB(\bm{\epsilon})}}\right) 
\leq  2
\exp\left(-Cm \log(mL)
\right)+\frac{C}{m^{8}}.
\end{align*}
Here $C$ represents some universal constants. Using the fact that $\frac{C_{g,U}(q)}{C_{f,L}^2(q)} \asymp \frac{C_{g,U}}{C_{f,L}^2}$ as $q$ approaches 0. It then follows that
\begin{align}
\label{Eqq1}
\frac{1}{\sqrt{m}}\|\widehat{\bm\theta}-\bm\theta^\star\|_2 
\lesssim \frac{C_{g,U}}{C_{f,L}^2} 
\sqrt{ \frac{\log(m)}{mLpB(\bm{\epsilon})}},
\end{align}
with probability at least $1-O(m^{-8})$. This completes the proof of the first result.

Next, we proceed to prove the second result. The proof for the second result is similar as above. First, by combining Lemma \ref{Temp_Lemma} and (\ref{Eqq1}), we obtain
\begin{align*}
\Vert \widehat{\bm{\theta}}-\bm{\theta}^\star\Vert_{\infty}
\leq 
    m^{-\frac{1}{2}} \Vert \widehat{\bm{\theta}} - 
    \bm{\theta}^\star\Vert_2 + 16\sqrt{\frac{\log(m)}{mLpB(\bm{\epsilon})}}+ \frac{4\lambda \kappa}{mp}
    \lesssim \left(
16 + \frac{C_{g,U}}{C_{f,L}^2} 
    \right)\sqrt{\frac{\log(m)}{mLpB(\bm{\epsilon})}}+\frac{4\lambda \kappa}{mp},
\end{align*}
with probability at least $1-O(m^{-8})-O(m^{-3})$. Here the probability is obtained by applying a union bound. In Theorem \ref{Thm:ConsisT1}, we have $\frac{\lambda}{mp} \lesssim \frac{\log(m)}{mLpB(\bm{\epsilon})}$. This then gives the desired result as follows:
\begin{align*}
    \Vert \widehat{\bm{\theta}}-\bm{\theta}^\star\Vert_{\infty}
\lesssim \sqrt{\frac{\log(m)}{mLpB(\bm{\epsilon})}},
\end{align*}
with probability at least $1-O(m^{-3})$. This completes the proof of Theorem \ref{Thm:Estimation}. \qed \\

\noindent
\textbf{Proof of Theorem \ref{Thm:Estimation_General}.} Note that the proof of Theorem \ref{Thm:Estimation} considers a general $F(x)$. Applying similar steps as in Theorem \ref{Thm:Estimation} will complete this proof. First, we have
\begin{align*}
    \mathbb{P}\left(\frac{1}{\sqrt{m}}\|\widehat{\bm\theta}-\bm\theta^\star\|_2 
\ge 
\frac{C_{g,U}(q)}{C_{f,L}^2(q)}
\sqrt{ \frac{2\log(m)}{mLpB(\bm{\epsilon})}}\right) 
\leq  2
\exp\left(-Cm \log(m)
\right)+\frac{C}{m^{8}},
\end{align*}
for some universal constants $C$. Next, by Lemma \ref{Temp_Lemma}, we have
\begin{align*}
\Vert \widehat{\bm{\theta}}-\bm{\theta}^\star\Vert_{\infty}
\leq &\frac{C_{g',U}(q)}{C_{g',L}(q)}
    m^{-\frac{1}{2}} \Vert \widehat{\bm{\theta}} - 
    \bm{\theta}^\star\Vert_2 + 8\frac{C_{g',U}(q)C_{g,U}(q)}{C_{g',L}(q)}\sqrt{\frac{\log(m)}{mLpB(\bm{\epsilon})}}+ \frac{2\lambda \kappa}{mpC_{g',L}(q)}, \\
    \lesssim &
\frac{C_{g',U}(q)C_{g,U}(q)}{C_{g',L}(q)C_{f,L}^2(q)}
    \sqrt{\frac{\log(m)}{mLpB(\bm{\epsilon})}}
\end{align*}
with probability at least $1-O(m^{-3})$. As $m$ and $L$ converge to zero, we have $q \rightarrow 0$ and 
\begin{align*}
    \Vert \widehat{\bm{\theta}}-\bm{\theta}^\star\Vert_{\infty} \lesssim 
    \frac{C_{g',U}C_{g,U}}{C_{g',L}C_{f,L}^2}
    \sqrt{\frac{\log(m)}{mLpB(\bm{\epsilon})}} 
    \asymp \sqrt{\frac{\log(m)}{mLpB(\bm{\epsilon})}.
    } 
\end{align*}
This completes the proof.\qed \\

\noindent
\textbf{Proof of Theorem \ref{Thm_1-Minimax}.} We first note that $z_{ij}^{(l)}$ is obtained as a linear transformation of $\widetilde{y}_{ij}^{(l)}$. Therefore, the amount of information stays unchanged and it is equivalently to consider the randomness of $\widetilde{y}_{ij}^{(l)}$'s. By the assumption on the missing pattern, the dataset $\{\widetilde{\bm{y}}^{(l)}\}_{l=1}^L$ can be illustrated by the following procedure:
\begin{align*}
  &  a_{ij}^{(l)} \sim \text{Bern}(p) 
    \stackrel{a_{ij}^{(l)}=1}{\Longrightarrow} y_{ij}^{(l)} \sim \text{Bernoulli}(F(\theta_i^\star-\theta_j^\star)) 
    \Longrightarrow \widetilde{y}_{ij}^{(l)} \in \{0,1\}, \\
  &  a_{ij}^{(l)} \sim \text{Bern}(p) 
    \stackrel{a_{ij}^{(l)}=0}{\Longrightarrow} y_{ij}^{(l)} \mbox{ is unobserved}
    \Longrightarrow \widetilde{y}_{ij}^{(l)} \mbox{ is unobserved}.
\end{align*}
For the case $a_{ij}^{(l)}=0$, we let $\widetilde{y}_{ij}^{(l)}=\text{``NaN"}$. In other words, given that $a_{ij}^{(l)}=0$, the distribution of $\widetilde{y}_{ij}^{(l)}$ is fixed for any $\bm{\theta}^\star$.

In what follows, we proceed to bound the KL divergence between two different $\bm{\theta}^\star$. Given that $y_{ij}^{(l)}\sim \text{Bern}(F(\theta_i^\star-\theta_j^\star))$ and $a_{ij}^{(l)}=1$, $\widetilde{y}_{ij}^{(l)}$ follows the distribution 
\begin{align*}
\widetilde{y}_{ij}^{(l)}= 
\begin{cases}
1 \mbox{ with probability } \left(\frac{1}{2}+
(F(\theta_i^\star-\theta_j^\star)-1/2)(1-2p_{\epsilon_l})\right), \\
0 \mbox{ with probability } \left(\frac{1}{2}-
(F(\theta_i^\star-\theta_j^\star)-1/2)(1-2p_{\epsilon_l})\right).
\end{cases}
\end{align*}

Let $\bm{\theta}_1^\star$ and $\bm{\theta}_2^\star$ be two elements in $\bm{\Theta}$ and let $D_{ij}^{(l)}(\bm{\theta}_1^\star)$ and $D_{ij}^{(l)}(\bm{\theta}_2^\star)$ denote the distributions of $\widetilde{Y}_{ij}^{(l)}$ under the true parameters $\bm{\theta}_1^\star$ and $\bm{\theta}_2^\star$, respectively. Then we turn to bound the KL divergence between $D_{ij}(\bm{\theta}_1^\star)$ and $D_{ij}(\bm{\theta}_2^\star)$. First, we have
\begin{align*}
   &\text{KL}\left(D_{ij}^{(l)}(\bm{\theta}_1^\star) \Vert D_{ij}^{(l)}(\bm{\theta}_2^\star)\right) \\
   =&
   p\text{KL}\left(D_{ij}^{(l)}(\bm{\theta}_1^\star) \Vert D_{ij}^{(l)}(\bm{\theta}_2^\star)\big|a_{ij}^{(l)}=1\right)+
   (1-p)\text{KL}\left(D_{ij}^{(l)}(\bm{\theta}_1^\star) \Vert D_{ij}^{(l)}(\bm{\theta}_2^\star)\big|a_{ij}^{(l)}=0\right) \\
   =&p\text{KL}\left(D_{ij}^{(l)}(\bm{\theta}_1^\star) \Vert D_{ij}^{(l)}(\bm{\theta}_2^\star)\big|a_{ij}^{(l)}=1\right).
\end{align*}
By the relation between KL divergence and $\chi^2$ divergence \citep{van2014renyi}, ons has
\begin{align}
\label{KL_Bound}
\text{KL}\left(D_{ij}^{(l)}(\bm{\theta}_1^\star) \Vert D_{ij}^{(l)}(\bm{\theta}_2^\star)\big|a_{ij}^{(l)}=1\right)
\leq 
\chi^2 \left(D_{ij}^{(l)}(\bm{\theta}_1^\star) \Vert D_{ij}^{(l)}(\bm{\theta}_2^\star)\big|a_{ij}^{(l)}=1\right)
=
\frac{(\widetilde{p}_{ij}^{(l)}-\widetilde{q}_{ij}^{(l)})^2}{\widetilde{q}_{ij}^{(l)}(1-\widetilde{q}_{ij}^{(l)})},
\end{align}
where $\widetilde{p}_{ij}^{(l)} = \mathbb{P}(\widetilde{Y}^{(l)}_{ij}=1|\bm{\theta}_1^\star)$ and $\widetilde{q}_{ij}^{(l)} = \mathbb{P}(\widetilde{Y}^{(l)}_{ij}=1|\bm{\theta}_2^\star)$. Their difference is given as
\begin{align*}
\widetilde{p}_{ij}^{(l)}-\widetilde{q}_{ij}^{(l)} = 
\left(
\frac{e^{\epsilon_l}-1}{e^{\epsilon_l}+1}\right)
\left(
F(\theta_{1,i}^\star-\theta_{1,j}^\star)-
F(\theta_{2,i}^\star-\theta_{2,j}^\star)
\right).
\end{align*}
Notice that
\begin{align*}
\Big|
F(\theta_{1,i}^\star-\theta_{1,j}^\star)-
F(\theta_{2,i}^\star-\theta_{2,j}^\star)
\Big|
\leq C_{f,U}
| \theta_{1,i}^\star-\theta_{1,j}^\star+\theta_{2,j}^\star - \theta_{2,i}^\star|,
\end{align*}
where $C_{f,U}=\max\limits_{x \in [-\kappa,\kappa]}f(x)$. Then we have
\begin{align*}
\left(\widetilde{p}_{ij}^{(l)}-\widetilde{q}_{ij}^{(l)}\right)^2 \leq 
C_{f,U}^2
\left(
\frac{e^{\epsilon_l}-1}{e^{\epsilon_l}+1}\right)^2 \left[
(\theta_{1,i}^\star-\theta_{2,i}^\star)^2 +(\theta_{1,j}^\star-\theta_{2,j}^\star)^2
\right].
\end{align*}
Next, we proceed to bound $\left(\widetilde{q}_{ij}(1-\widetilde{q}_{ij})
\right)^{-1}$. By the definition of $\widetilde{q}_{ij}$, one has
\begin{align*}
\left|
\widetilde{q}_{ij}^{(l)}-1/2\right|=\left|
F(\theta_{2,i}^\star-\theta_{2,j}^\star)-\frac{1}{2}
\right|
\frac{e^{\epsilon_l}-1}{e^{\epsilon_l}+1}
\leq\left|
F(\theta_{2,i}^\star-\theta_{2,j}^\star)-\frac{1}{2}
\right|,
\end{align*}
Clearly, $\widetilde{q}_{ij}^{(l)}$ is closer to $1/2$ than $F(\theta_{2,i}^\star-\theta_{2,j}^\star)$. Therefore, we have
\begin{align}
\label{Kappa_bound}
\widetilde{q}_{ij}^{(l)}(1-\widetilde{q}_{ij}^{(l)}) \geq &F(\theta_{2,i}^\star-\theta_{2,j}^\star)(1-F(\theta_{2,i}^\star-\theta_{2,j}^\star)) \notag \\
\geq &
\min\left\{
\frac{F(\theta_{2,i}^\star-\theta_{2,j}^\star)}{2},
\frac{1-F(\theta_{2,i}^\star-\theta_{2,j}^\star)}{2}
\right\},
\end{align}
where the last inequality is based on $x(1-x) \geq \min\{x/2,(1-x)/2\}$ for any $x\in [0,1]$. Plugging (\ref{Kappa_bound}) into (\ref{KL_Bound}) gives
\begin{align*}
\text{KL}\left(D_{ij}^{(l)}(\bm{\theta}_1^\star) \Vert D_{ij}^{(l)}(\bm{\theta}_2^\star)\right) 
\leq 
\frac{2pC_{f,U}^2}{F(-\kappa)}
\left(
\frac{e^{\epsilon_l}-1}{e^{\epsilon_l}+1}\right)^2 \left[
(\theta_{1,i}^\star-\theta_{2,i}^\star)^2 +(\theta_{1,j}^\star-\theta_{2,j}^\star)^2
\right].
\end{align*}
Let $D^{(l)}(\bm{\theta}_1^\star)$ be the joint distribution of all pairwise comparisons of the $l$-th user. By the independence assumption, it holds that 
\begin{align*}
\text{KL}\left(D^{(l)}(\bm{\theta}_1^\star) \Vert D^{(l)}(\bm{\theta}_2^\star)\right) 
\leq 
\sum_{i<j}
\text{KL}\left(D_{ij}^{(l)}(\bm{\theta}_1^\star) \Vert D_{ij}^{(l)}(\bm{\theta}_2^\star)\right) 
\leq  
\frac{p(m-1)C_{f,U}^2}{F(-\kappa)}
\left(
\frac{e^{\epsilon_l}-1}{e^{\epsilon_l}+1}\right)^2 
\Vert \bm{\theta}_1^\star - \bm{\theta}_2^\star \Vert_2^2.
\end{align*}

Next, we apply the Fano's inequality to derive the minimax lower bound. Recall that $\bm{\Theta}$ is defined as 
\begin{align*}
\bm{\Theta}=\left\{
\bm{\theta} \in \mathbb{R}^m:\bm{1}_m^T \bm{\theta}=0,
\max_{i \neq j}|\theta_i - \theta_j| \leq \kappa
\right\}.
\end{align*}
Notice that $\max_{i \neq j}|\theta_i - \theta_j|  \leq \max_{i}2|\theta_i|
\leq 2
\sqrt{\sum_{i=1}^n \theta_i^2} =
2\Vert \bm{\theta}\Vert_2$. Therefore, $2\Vert \bm{\theta}\Vert_2 \leq \kappa$ implies $\max_{i \neq j}|\theta_i - \theta_j| \leq \kappa$. With this, we define $
\bm{\Theta}'(c) = 
\left\{
\bm{\theta} \in \mathbb{R}^n:\bm{1}_n^T \bm{\theta}=0,
\Vert \bm{\theta} \Vert_2 \leq c
\right\}$.

We denote $r = \min\{\kappa/2,1\}$ and let $\mathcal{S}=\{\widetilde{\bm{\theta}}^{(1)},\ldots,\widetilde{\bm{\theta}}^{(M)}\}$ denote the maximum $1/2$-packing set of $\bm{\Theta}'(1)$ with respect to the metrics $\Vert \cdot \Vert_2$. Then for any two $\widetilde{\bm{\theta}}^{(i)},\widetilde{\bm{\theta}}^{(j)} \in \mathcal{S}$, we have $\Vert \widetilde{\bm{\theta}}^{(i)}-\widetilde{\bm{\theta}}^{(j)} \Vert_2 \geq 1/2$. Next, we define a new set $\mathcal{S}_{r}$ with $\delta \leq 1$ as $\mathcal{S}_{r} = 
\left\{
 \bm{\theta}^{(i)}= 
\delta r
\widetilde{\bm{\theta}}^{(i)}:
\widetilde{\bm{\theta}}^{(i)} \in \mathcal{S}
\right\}$. We can verify that $\Vert \bm{\theta}^{(i)} \Vert_2 \leq \delta r  \leq r$ and 
\begin{align*}
 \Vert \bm{\theta}^{(i)} - \bm{\theta}^{(j)}\Vert_2=\delta
r \Vert \widetilde{\bm{\theta}}^{(i)} - \widetilde{\bm{\theta}}^{(j)}\Vert_2 \geq \delta r/2.
\end{align*}
Note that $\Vert \widetilde{\bm{\theta}}^{(i)} \Vert_2 \leq 1$ for $i \in [M]$, we have $ \Vert \bm{\theta}^{(i)} - \bm{\theta}^{(j)}\Vert_2 \leq 2\delta r$ and $\Vert \bm{\theta}^{(i)} \Vert_2 \leq \delta r$. Therefore, $\mathcal{S}_{r}$ is the maximal $\delta r/2$-packing set of $\bm{\Theta}'(\delta r)$. With this, we then have
\begin{align*}
\sup_{\bm{\theta}^\star \in \bm{\Theta}}
\mathbb{E}
\left[
\Vert  \widehat{\bm{\theta}} - \bm{\theta}^\star  \Vert_2
\right] \geq 
\sup_{\bm{\theta}^\star \in\bm{\Theta}'(\delta r)}
\mathbb{E}
\left[
\Vert  \widehat{\bm{\theta}} - \bm{\theta}^\star  \Vert_2
\right] \geq 
\sup_{\bm{\theta}^\star \in \mathcal{S}_{r}}
\mathbb{E}
\left[
\Vert  \widehat{\bm{\theta}} - \bm{\theta}^\star  \Vert_2
\right] .
\end{align*}

Next, we consider the decision rule $\omega(\widehat{\bm{\theta}}) \in \{\bm{\theta}^{(i)}, i\in [M]\}$ for any estimator $\widehat{\bm{\theta}}$, which is defined as $\omega(\widehat{\bm{\theta}}) =\argmin\limits_{\bm{\theta} \in  \{\bm{\theta}^{(i)}, i\in [M]\}}\Vert \widehat{\bm{\theta}}  - \bm{\theta}\Vert_2$. Notice that when $\Vert \widehat{\bm{\theta}}  - \bm{\theta}^{(i)}\Vert_2<\delta r/4$, we have $\Vert \widehat{\bm{\theta}}  - \bm{\theta}^{(j)}\Vert_2>\delta r/4$ for any $j \neq i$ by the triangle inequality. Therefore, if $\bm\theta^\star\in\mathcal{S}_r$, then $\omega(\widehat{\bm{\theta}}) \neq \bm{\theta}^\star$ indicates that $\Vert \widehat{\bm{\theta}} - \bm{\theta}^\star  \Vert_2>\delta r/4$. This leads to a lower bound as
\begin{align*}
\sup_{\bm{\theta}^\star \in \mathcal{S}_{r}}
\mathbb{E}
\left[
\Vert  \widehat{\bm{\theta}} - \bm{\theta}^\star  \Vert_2
\right] \geq &
\sup_{\bm{\theta}^\star \in \mathcal{S}_{r}}
\mathbb{E}
\left[
\Vert  \widehat{\bm{\theta}} - \bm{\theta}^\star  \Vert_2 I\left(\Vert \widehat{\bm{\theta}} - \bm{\theta}^\star  \Vert_2>\delta r/4\right)
\right] \\
 \geq &  \sup_{\bm{\theta}^\star \in \mathcal{S}_r} \frac{\delta r}{4} \mathbb{P}_{\bm{\theta}^\star}\left(\Vert \widehat{\bm{\theta}} - \bm{\theta}^\star  \Vert_2>\delta r/4 \right) \geq 
\frac{\delta r}{4} \sup_{\bm{\theta}^\star \in \mathcal{S}_r} 
\mathbb{P}_{\bm{\theta}^\star}\left(\omega(\widehat{\bm{\theta}}) \neq \bm{\theta}^\star\right),
\end{align*}
where $\mathbb{P}_{\bm{\theta}^\star}$ denotes the probability function of the ranking dataset $\widetilde{\mathcal{D}}$ and $\bm{\theta}^\star$ is the parameter vector of the underlying model. Furthermore,
\begin{align*}
\frac{\delta r}{4} \sup_{\bm{\theta}^\star \in \mathcal{S}_r} 
\mathbb{P}_{\bm{\theta}^\star}\left(\omega(\widehat{\bm{\theta}}) \neq \bm{\theta}^\star\right)
\geq \frac{\delta r}{4} \frac{1}{M}\sum_{i=1}^M 
\mathbb{P}_{\bm{\theta}^{(i)}}\left(\omega(\widehat{\bm{\theta}}) \neq \bm{\theta}^{(i)}\right)
=\frac{\delta r}{4} \mathbb{P}\left(\omega(\widehat{\bm{\theta}}) \neq \bm{\theta}'\right),
\end{align*}
where $\bm{\theta}'$ is a random variable uniformly distributed over $\mathcal{S}_r$. Applying Fano's inequality, we obtain
\begin{align}
\label{Fano}
\sup_{\bm{\theta}^\star \in \bm{\Theta}}
\mathbb{E}
\left[
\Vert  \widehat{\bm{\theta}} - \bm{\theta}^\star  \Vert_2
\right] \geq \frac{\delta r}{4} 
\left(
1-
\frac{I(\bm{\theta}',\{\widetilde{\bm{y}}^{(l)}\}_{l=1}^L)+\log 2}{\log(M)}
\right),
\end{align}
where $I(\bm{\theta}',\{\widetilde{\bm{y}}^{(l)}\}_{l=1}^L)$ denotes the mutual information between two random variables $\bm{\theta}'$ and $\widetilde{\bm{y}}^{(l)}$, which can be bounded as
\begin{align*}
I(\bm{\theta}',\{\widetilde{\bm{y}}^{(l)}\}_{l=1}^L) 
\leq &
\frac{p(m-1)C_{f,U}^2}{F(-\kappa)M^2}
\sum_{i=1}^M \sum_{j=1}^M \sum_{l=1}^L
\left(
\frac{e^{\epsilon_l}-1}{e^{\epsilon_l}+1}\right)^2 
\Vert \bm{\theta}^{(i)} - \bm{\theta}^{(j)} \Vert_2^2\\
=& 
\frac{p(m-1)L C_{f,U}^2}{F(-\kappa)M^2}
\frac{1}{L}\sum_{l=1}^L 
\left(
\frac{e^{\epsilon_l}-1}{e^{\epsilon_l}+1}\right)^2 
\sum_{i=1}^M \sum_{j=1}^M
\Vert \bm{\theta}^{(i)}- \bm{\theta}^{(j)} \Vert_2^2 \\
\leq & 
\frac{4p(m-1)L C_{f,U}^2}{F(-\kappa)}
\frac{1}{L}\sum_{l=1}^L 
\left(
\frac{e^{\epsilon_l}-1}{e^{\epsilon_l}+1}\right)^2 
\delta^2r^2.
\end{align*}
Plugging this into the lower bound in (\ref{Fano}) yields
\begin{align}
\label{fano2}
\sup_{\bm{\theta}^\star \in \bm{\Theta}}
\mathbb{E}
\left[
\frac{1}{m^{\frac{1}{2}}}
\Vert  \widehat{\bm{\theta}} - \bm{\theta}^\star  \Vert_2
\right] \geq & 
\frac{\delta r}{4m^{\frac{1}{2}}} 
\left(
1-\frac{
\frac{4pm(m-1) L C_{f,U}^2}{F(-\kappa)}
\frac{1}{L}\sum_{l=1}^L 
\left(
\frac{e^{\epsilon_l}-1}{e^{\epsilon_l}+1}\right)^2 
\frac{\delta^2r^2}{m}
+\log(2)}{\log(M)}
\right).
\end{align}
Here we also need to provide an lower bound for $\log(M)$. Recall that the definition of $\bm{\Theta}'(c)$ is given as $ \bm{\Theta}'(c) = \left\{\bm{\theta}=(\theta_1,\ldots,\theta_m): \bm{1}_m^T \bm{\theta}=0, \Vert \bm{\theta}\Vert_2 \leq c
\right\}$. Next, suppose $\bm{U} \in \mathbb{R}^{m\times m}$ is an orthogonal matrix with the last column being $\bm{U}_{.m}=(m^{-1/2},\ldots,m^{-1/2})$. With this, we define a new set as
\begin{align*}
\widetilde{\bm{\Theta}}'(c)=
\left\{
\widetilde{\bm{\theta}} = \bm{\theta}^T\bm{U}:\bm{\theta} \in \bm{\Theta}'(c) 
\right\}.
\end{align*}
We can verify that the last element of $\widetilde{\bm{\theta}}$ is zero for any $\widetilde{\bm{\theta}}\in \widetilde{\bm{\Theta}}'(c)$ since $\bm{1}_m^T\bm{\theta}=0$ for any $\bm{\theta} \in \bm{\Theta}'(c)$. Therefore, $\widetilde{\bm{\Theta}}'(c)$ can be reformulated as
\begin{align*}
\widetilde{\bm{\Theta}}'(c) = \left\{
\widetilde{\bm{\theta}}=(\widetilde{\theta}_1,\ldots,\widetilde{\theta}_{m-1},0):
\sqrt{\sum_{i=1}^{m-1}\widetilde{\theta}_i^2} \leq c
\right\}.
\end{align*}
Since orthogonal matrix possesses the property of preservation of length, we get $\Vert \bm{\theta}_1^T \bm{U}-\bm{\theta}_2^T \bm{U}\Vert_2 = 
\Vert \bm{\theta}_1^T-\bm{\theta}_2^T \Vert_2$ for any $\bm{\theta}_1,\bm{\theta}_2$. This equality indicates that $\{ \bar{\bm{\theta}}^{(i)}=(\widetilde{\bm{\theta}}^{(i)})^T \bm{U},\widetilde{\bm{\theta}}^{(i)} \in \mathcal{S}\}$ is the maximal $1/2$-packing set of $\widetilde{\bm{\Theta}}'(1)$. Therefore, we have $\log(M) \geq (m-1)\log(2)$. With this, (\ref{fano2}) becomes
\begin{align}
\label{fano3}
\sup_{\bm{\theta}^\star \in \bm{\Theta}}
\mathbb{E}
\left[
\frac{1}{\sqrt{m}}
\Vert  \widehat{\bm{\theta}} - \bm{\theta}^\star  \Vert_2
\right] \geq & 
\frac{\delta r}{4m^{\frac{1}{2}}} 
\left(
1-\frac{
\frac{4pm(m-1) L C_{f,U}^2}{F(-\kappa)}
B(\bm{\epsilon})
\frac{\delta^2r^2}{m}
+\log(2)}{(m-1)\log(2)}
\right).
\end{align}
where $B(\bm{\epsilon}) = \frac{1}{L}\sum_{l=1}^L 
\left(\frac{e^{\epsilon_l}-1}{e^{\epsilon_l}+1}\right)^2$.
If we let $\frac{\delta^2r^2}{m} = \frac{F(-\kappa)}{16mLpC_{f,U}^2B(\bm{\epsilon})}$ and suppose $\frac{F(-\kappa)}{16LpC_{f,U}^2B(\bm{\epsilon})} \leq 1$, it follows that
\begin{align*}
\sup_{\bm{\theta}^\star \in \bm{\Theta}}
\mathbb{E}
\left[
\frac{1}{\sqrt{m}}
\Vert  \widehat{\bm{\theta}} - \bm{\theta}^\star  \Vert_2
\right] \gtrsim 
\frac{F(-\kappa)}{C_{f,U}}\sqrt{
\frac{1}{m p L B(\bm{\epsilon})}}.
\end{align*}
By the fact that $\Vert \widehat{\bm{\theta}} - \bm{\theta}^\star \Vert_{\infty} \geq \frac{1}{\sqrt{m}}\Vert \widehat{\bm{\theta}} - \bm{\theta}^\star \Vert_{2}$, we further have
\begin{align*}
    \sup_{\bm{\theta}^\star \in \bm{\Theta}}
\mathbb{E}
\left[
\Vert \widehat{\bm{\theta}} - \bm{\theta}^\star \Vert_{\infty}
\right] \geq 
    \sup_{\bm{\theta}^\star \in \bm{\Theta}}
\mathbb{E}
\left[
\frac{1}{\sqrt{m}}
\Vert  \widehat{\bm{\theta}} - \bm{\theta}^\star  \Vert_2
\right] \gtrsim 
\frac{F(-\kappa)}{C_{f,U}}\sqrt{
\frac{1}{m p LB(\bm{\epsilon})}}.
\end{align*}
This completes the proof. \qed \\

\noindent
\textbf{Proof of Theorem \ref{Thm: partial}}: Following Lemma \ref{Lemma:REB}, we have
\begin{align*}
\mathbb{E}\big(H_K(\widehat{\bm{\theta}},\bm{\theta}^\star)\big)  
&\le \frac{1}{2K}\left[\sum_{i:\sigma(\theta_i^\star) \leq K}
\mathbb{P}\left(\widehat{\theta}_i \leq \frac{\theta_{(K)}^\star+\theta_{(K+1)}^\star}{2}\right) +\sum_{i:\sigma(\theta_i^\star)> K}\mathbb{P}\left(\widehat{\theta}_i \geq  \frac{\theta_{(K)}^\star+\theta_{(K+1)}^\star}{2}\right)\right].
\end{align*}
Thus, it suffices to bound $\mathbb{P}\left(\widehat{\theta}_i \leq \frac{\theta_{(K)}^\star+\theta_{(K+1)}^\star}{2}\right)$ and $\mathbb{P}\left(\widehat{\theta}_i \geq \frac{\theta_{(K)}^\star+\theta_{(K+1)}^\star}{2}\right)$ separately. For each $i$ with $\sigma(\theta_i^\star) \leq K$, we have $\theta_i^\star\ge \theta_{(K)}^\star$. Then, it holds true that
\begin{align*}
&\mathbb{P}\left(\widehat{\theta}_i \leq \frac{\theta_{(K)}^\star+\theta_{(K+1)}^\star}{2}\right)=\mathbb{P}\left(\widehat{\theta}_i -\theta_i^\star\leq \frac{\theta_{(K)}^\star+\theta_{(K+1)}^\star}{2}-\theta_i^\star\right)\\
 \le &\mathbb{P}\left(\widehat{\theta}_i -\theta_i^\star\leq \frac{\theta_{(K)}^\star+\theta_{(K+1)}^\star}{2}-\theta_{(K)}^\star\right) \le  \mathbb{P}\left(|\widehat{\theta}_i -\theta_i^\star|\geq \frac{\Delta_K}{2}\right).
\end{align*}
Similarly, for each $i$ with $\sigma(\theta_i^\star) > K$, we have $\theta_i^\star\le \theta_{(K+1)}^\star$. Therefore, 
\begin{align*}
&\mathbb{P}\left(\widehat{\theta}_i \geq \frac{\theta_{(K)}^\star+\theta_{(K+1)}^\star}{2}\right)=\mathbb{P}\left(\widehat{\theta}_i -\theta_i^\star\geq \frac{\theta_{(K)}^\star+\theta_{(K+1)}^\star}{2}-\theta_i^\star\right)\\
 \le &\mathbb{P}\left(\widehat{\theta}_i -\theta_i^\star\geq \frac{\theta_{(K)}^\star+\theta_{(K+1)}^\star}{2}-\theta_{(K+1)}^\star\right)
 \le  \mathbb{P}\left(|\widehat{\theta}_i -\theta_i^\star|\geq \frac{\Delta_K}{2}\right).
\end{align*}
To sum up, we get
\begin{align*}
\mathbb{E}\big(H_K(\widehat{\bm{\theta}},\bm{\theta}^\star)\big)  
\leq &
\frac{1}{K}
\sum_{i=1}^K
\mathbb{P}\left(|\widehat{\theta}_i -\theta_i^\star|\geq \frac{\Delta_K}{2}\right)
\leq 
\mathbb{P}\left(
\Vert \widehat{\bm{\theta}} - \bm{\theta}^\star \Vert_{2} \geq 
\frac{\Delta_K}{2}
\right).
\end{align*}
Next, we proceed to provide an upper bound for $\mathbb{P}\left(
\Vert \widehat{\bm{\theta}} - \bm{\theta}^\star \Vert_{2} \geq 
\frac{\Delta_K}{2}
\right)$. If $\Delta_K/(2\sqrt{m})$ is larger than $q_0$ specified in Lemma \ref{Lemma:Boundedness}, we have $\mathbb{P}\left(
\Vert \widehat{\bm{\theta}} - \bm{\theta}^\star \Vert_{2} \geq 
\frac{\Delta_K}{2}\right) \lesssim m^{-8}$. Next, we consider the case that $\Delta_K/2$ is extremely small.
\begin{align*}
& \mathbb{P}\left(
\Vert \widehat{\bm{\theta}} - \bm{\theta}^\star \Vert_{2} \geq 
\frac{\Delta_K}{2}
\right) \\
=&  \mathbb{P}\left(
\Vert \widehat{\bm{\theta}} - \bm{\theta}^\star \Vert_{2} \geq 
\frac{\Delta_K}{2} \Big| 
\widehat{\bm{\theta}} \in \mathcal{E}(q_0)
\right) \mathbb{P}\left(\widehat{\bm{\theta}} \in \mathcal{E}(q_0)\right) +   \mathbb{P}\left(
\Vert \widehat{\bm{\theta}} - \bm{\theta}^\star \Vert_{2} \geq 
\frac{\Delta_K}{2} \Big| 
\widehat{\bm{\theta}} \in \mathcal{E}^c(q_0)
\right) \mathbb{P}\left(\widehat{\bm{\theta}} \in \mathcal{E}^c(q_0)\right) \\
\leq &  \mathbb{P}\left(
\Vert \widehat{\bm{\theta}} - \bm{\theta}^\star \Vert_{2} \geq 
\frac{\Delta_K}{2} \Big| 
\widehat{\bm{\theta}} \in \mathcal{E}(q_0)
\right)  +   \mathbb{P}\left(\widehat{\bm{\theta}} \in \mathcal{E}^c(q_0)\right) \\
\lesssim &\mathbb{P}\left(
\Vert \widehat{\bm{\theta}} - \bm{\theta}^\star \Vert_{2} \geq 
\frac{\Delta_K}{2} \Big| 
\widehat{\bm{\theta}} \in \mathcal{E}(q_0)
\right)+ Cm^{-8}.
\end{align*}
Next, by (\ref{Use}), one has
\begin{align*}
 &\mathbb{P}\left(\|\widehat{\bm\theta}-\bm\theta^\star\|_2 
\ge 
\frac{\Delta_K}{2}
\right) 
\leq 
2
\exp\left(-C
\frac{ C_{f,L}^4(q)m L p B(\bm{\epsilon})\Delta_K^2}{C_{g,U}^2(q)}
\right)+Cm^{-8}.
\end{align*}
Recall that $H_K(\widehat{\bm{\theta}},\bm{\theta}^\star)$ takes values in $\{0,\frac{1}{K},\frac{2}{K},\ldots,1\}$. Therefore, $O(m^{-8})$ is extremely small compared with $1/K$. If the second term dominates the first term, then we have $\mathbb{E}\big(H_K(\widehat{\bm{\theta}},\bm{\theta}^\star)\big)\lesssim  m^{-8}$, indicating that $H_K(\widehat{\bm{\theta}},\bm{\theta}^\star)=0$ with probability at least $1-O(m^{-7})$. Then by the Markov's inequality, we have
\begin{align*}
    H_K(\widehat{\bm{\theta}},\bm{\theta}^\star) \lesssim 
    \exp\left(-C m L p B(\bm{\epsilon})\Delta_K^2
\right),
\end{align*}
with probability $1-o(1)$, where $C$ denotes some constant depending on $F(\cdot)$.
\qed \\

\noindent
\textbf{Proof of Theorem \ref{Thm:FRE}.} Without loss of generality, we assume that $\bm{\theta}^\star$ satisfies $\theta_i^\star>\theta_j^\star$ for $i<j$, i.e., $\sigma(\theta_i^\star)=i$ and $\theta_{(i)}^\star=\theta_i^\star$. First, we establish the relation between $K(\widehat{\bm{\theta}},\bm{\theta}^\star)$ and $S(\widehat{\bm{\theta}},\bm{\theta}^\star)$ based on the Diaconis–Graham inequality \citep{diaconis1977spearman}. Specifically, we utilize the fact $|\sigma(\widehat{\theta}_i)-\sigma(\theta_i^\star)| \leq \sum_{j=1}^m I\left((\sigma(\widehat{\theta}_i)-\sigma(\widehat{\theta}_j))(\sigma(\theta_i^\star)-\sigma(\theta_j^\star))<0\right)$. If $|\sigma(\widehat{\theta}_i)-\sigma(\theta_i^\star)|=k$, it implies that at least $k$ relative positions between item $i$ and other items are incorrect. Therefore, it follows that
\begin{align*}
    \sum_{i=1}^m |\sigma(\widehat{\theta}_i)-\sigma(\theta_i^\star)|
    \leq & 
    \sum_{i=1}^m
    \sum_{j=1}^m I\left((\sigma(\widehat{\theta}_i)-\sigma(\widehat{\theta}_j))(\sigma(\theta_i^\star)-\sigma(\theta_j^\star))<0\right) \\
    = &2\sum_{1 \leq i<j\leq m} I\left((\sigma(\widehat{\theta}_i)-\sigma(\widehat{\theta}_j))(\sigma(\theta_i^\star)-\sigma(\theta_j^\star))<0\right) 
    \leq  m(m-1) K(\widehat{\bm{\theta}},\bm{\theta}^\star).
\end{align*}
Therefore, we have
\begin{align*}
    S(\widehat{\bm{\theta}},\bm{\theta}^\star) \leq \frac{2m(m-1)}{m^2}K(\widehat{\bm{\theta}},\bm{\theta}^\star) \leq 2K(\widehat{\bm{\theta}},\bm{\theta}^\star).
\end{align*}

Note that $\sigma(\widehat{\theta}_i)-\sigma(\widehat{\theta}_j)<0$ is equivalent to $\widehat{\theta}_i>\widehat{\theta}_j$. Therefore, $\mathbb{E}\left(
K(\widehat{\bm{\theta}},\bm{\theta}^\star)
\right)$ can be rewritten as
\begin{align}
\label{MainBound}
\mathbb{E}\left(
K(\widehat{\bm{\theta}},\bm{\theta}^\star)
\right)  = 
\frac{2}{m(m-1)}
\sum_{1 \leq i<j \leq m}
\mathbb{P}
\left(
\widehat{\theta}_{i}<
\widehat{\theta}_{j}
\right).
\end{align}
We first note that, for any $i<j$ with $\theta_i^\star>\theta_j^\star$, 
\begin{align*}
    \{\widehat{\theta}_{i}<\widehat{\theta}_{j}\}
    \subset 
    \{|\theta_i^\star-\widehat{\theta}_{i}|>(\theta_i^\star-\theta_j^\star)/2\} \cup 
    \{|\widehat{\theta}_{j}-\theta_j^\star|>(\theta_i^\star-\theta_j^\star)/2\}.
\end{align*}
Therefore, we have
\begin{align}
\label{Use1}
\mathbb{P}
\left(
\widehat{\theta}_{i}<
\widehat{\theta}_{j}
\right)
\leq &
\mathbb{P}\left(
\left|\widehat{\theta}_{i}- \theta_i^\star\right|>\frac{\theta_i^\star-\theta_j^\star}{2}
\right)+ 
\mathbb{P}\left(
\left|\widehat{\theta}_{j}-\theta_j^\star\right|>\frac{\theta_i^\star-\theta_j^\star}{2}
\right)
\end{align}
Plugging (\ref{Use1}) into the right-hand side of (\ref{MainBound}) yields that
\begin{align*}
&\mathbb{E}\left(
K(\widehat{\bm{\theta}},\bm{\theta}^\star)
\right)
\leq 
\frac{2}{m(m-1)}
\sum_{1 \leq i<j \leq m}
\left[
\mathbb{P}\left(
\left|\widehat{\theta}_{i}- \theta_i^\star\right|>\frac{\theta_i^\star-\theta_j^\star}{2}
\right)+ 
\mathbb{P}\left(
\left|\widehat{\theta}_{j}-\theta_j^\star\right|>\frac{\theta_i^\star-\theta_j^\star}{2}
\right)
\right] \\
\leq &
\frac{2}{m-1}
\sum_{i=1}^{m-1}
\mathbb{P}\left(
\Vert \widehat{\bm{\theta}} - \bm{\theta}^\star \Vert_{2}>\frac{\theta_{(i)}^\star-\theta_{(i+1)}^\star}{2}
\right)=
\frac{2}{m-1}
\sum_{i=1}^{m-1}
\mathbb{P}\left(
\Vert \widehat{\bm{\theta}} - \bm{\theta}^\star \Vert_{2}>\frac{\Delta_i}{2}
\right)
\end{align*}
where the last inequality follows from the facts that $|\widehat{\theta}_{i}- \theta_i^\star|\leq \Vert \widehat{\bm{\theta}} - \bm{\theta}^\star \Vert_{2}$ for any $i\in [m]$ and that $\theta_{(i)}^\star-\theta_{(i+1)}^\star \leq  \theta_{(i)}^\star-\theta_{(j)}^\star$ for $j \geq i+1$. In what follows, we proceed to bound $\mathbb{P}\big(\Vert \widehat{\bm{\theta}} - \bm{\theta}^\star \Vert_{2}>\frac{\theta_{(i)}^\star-\theta_{(i+1)}^\star}{2}\big)$. Applying similar steps as in the proof of Theorem \ref{Thm: partial}, we get
\begin{align*}
\mathbb{E}
\left(
S(\widehat{\bm{\theta}},\bm{\theta}^\star)
\right) \leq & 2
\mathbb{E}\left(
K(\widehat{\bm{\theta}},\bm{\theta}^\star)
\right)  
\lesssim  
\frac{4}{m-1}
\sum_{i=1}^{m-1}
\exp\left(
- CmLp\Delta_i^2 B(\bm{\epsilon})
\right).
\end{align*}
The desired results immediately follows by applying the Markov inequality. This completes the proof.\qed \\

\subsection{Proof of All Lemmas}
\label{SuppSec:Lemmas}

\noindent
\textbf{Proof of Lemma \ref{Lemma:Consis}.} By the definition of $Y_{ij}$, we have
\begin{align*}
\mathbb{P}\big(
\widetilde{Y}_{ij}=1
\big) = &
\mathbb{P}\left(
Y_{ij}=1
\right)  (1-p_{\epsilon}) + 
\mathbb{P}\left(
Y_{ij}=0
\right) p_{\epsilon} \\
=&
\frac{e^{\theta_i^\star}}{e^{\theta_i^\star}+e^{\theta_j^\star}}(1-p_{\epsilon})+
\frac{e^{\theta_j^\star}}{e^{\theta_i^\star}+e^{\theta_j^\star}}p_{\epsilon} \\
=&
\frac{1}{2}+
\frac{e^{\theta_i^\star} - e^{\theta_j^\star}}{e^{\theta_i^\star}+e^{\theta_j^\star}}(1/2-p_{\epsilon}).
\end{align*}
This completes the proof. \qed \\

\noindent
\textbf{Proof of Lemma \ref{Lemma:Debias}.} For each $l \in [L]$ and $i\neq j$, we have
\begin{align*}
&\mathbb{E}\left(
\widetilde{y}_{ij}^{(l)}
\right) =
F(\theta_i^\star-\theta_j^\star)(1-p_{\epsilon_l})+
(1-F(\theta_i^\star-\theta_j^\star))p_{\epsilon_l} \\
=&
\frac{1}{2}+
F(\theta_i^\star-\theta_j^\star)\left(\frac{1}{2}-p_{\epsilon_l}\right)+
(1-F(\theta_i^\star-\theta_j^\star))\left(p_{\epsilon_l}-\frac{1}{2}\right) \\
=& \frac{1}{2}+\left(\frac{1}{2}-p_{\epsilon_l}\right)(2F(\theta_i^\star-\theta_j^\star)-1) = 
\frac{1}{2}+\left(\frac{e^{\epsilon_l}-1}{e^{\epsilon_l}+1}\right)
\left(F(\theta_i^\star-\theta_j^\star)-\frac{1}{2}\right).
\end{align*}
Therefore, we have
\begin{align*}
    \mathbb{E}\left(
\frac{(e^{\epsilon_l}+1)\widetilde{y}_{ij}^{(l)}-1}{e^{\epsilon_l}-1} 
\right) = &
\frac{e^{\epsilon_l}-1}{2(e^{\epsilon_l}-1)} 
+\frac{e^{\epsilon_l}+1}{e^{\epsilon_l}-1} \left(\frac{e^{\epsilon_l}-1}{e^{\epsilon_l}+1}\right)
\left(F(\theta_i^\star-\theta_j^\star)-\frac{1}{2}\right) 
=F(\theta_i^\star-\theta_j^\star).
\end{align*}
As for the variance,
\begin{align*}
\text{Var}\left(\widetilde{z}_{ij}^{(l)}\right) = 
\left(
\frac{e^{\epsilon_l}+1}{e^{\epsilon_l}-1}\right)^2
\left[
 \frac{1}{4}- \frac{1}{4}\left( \frac{e^{\epsilon_l}-1}{e^{\epsilon_l}+1}
 \right)^2 
 \left(
2F(\theta_i^\star-\theta_j^\star)-1
\right)^2\right].
\end{align*}
This completes the proof. \qed \\

\noindent
\textbf{Proof of Lemma \ref{Lemma:Restricted}.} Note that $\mathcal{L}_{\lambda}(\bm{\theta}) = \mathcal{L}_0(\bm{\theta})+\lambda \Vert \bm{\theta}\Vert_2^2$. For any $\bm{g}=(g_1,\ldots,g_{m-1}) \in \mathbb{R}^{m-1}$, we define
\begin{align*}
\bm{\Theta}(\bm{g}) = \big\{
\bm{\theta} \in \mathbb{R}^m : 
\theta_i - \theta_{i+1} = g_i, \text{ for } i \in [m-1]
\big\}.
\end{align*}
Notice that $\mathcal{L}_0(\bm{\theta})$ stays invariant to translation for any fixed $\bm{\theta}$. In other words, $\mathcal{L}_0(\bm{\theta}) = \mathcal{L}_0(\bm{\theta}+C\bm{1}_m )$ for any constant $C$. Therefore, for any $\bm{\theta}_1,\bm{\theta}_2\in \bm{\Theta}(\bm{g})$, we have $\mathcal{L}_0(\bm{\theta}_1)=\mathcal{L}_0(\bm{\theta}_2)$. With this, for any $\bm{\theta}+C\bm{1}_m \in \bm{\Theta}(\bm{g})$,
\begin{align}
\label{L4_Eq1}
\Vert \bm{\theta}+C \bm{1} \Vert_2^2 = 
\Vert \bm{\theta} \Vert_2^2+2C\sum_{i=1}^m \theta_i +mC^2.
\end{align}
Clearly, (\ref{L4_Eq1}) is minimized when $C = -\frac{1}{m}\sum_{i=1}^m \theta_i$. Therefore, for any $\bm{g}\in \mathbb{R}^{m-1}$, we have
\begin{align*}
\bm{\theta}_0 - \overline{\theta}_0 \bm{1}=
\argmin_{\bm{\theta} \in \bm{\Theta}(\bm{g})}
\left\{
 \mathcal{L}_0(\bm{\theta})+\lambda \Vert \bm{\theta}\Vert_2^2
\right\}
,
\end{align*}
where $\bm{\theta}_0$ is any vector in $\bm{\Theta}(\bm{g})$ and $\overline{\theta}_0 =\frac{1}{m}\sum_{i=1}^m \theta_{0i}$. This result holds for any $\bm{g}\in \mathbb{R}^{m-1}$. It then follows that $\bm{1}_m^T\widehat{\bm{\theta}}=0$. This completes the proof.\qed \\

\noindent \textbf{Proof of Lemma \ref{Lemma:Hessian}.}  \textbf{(Strongly Convexity of $\mathbb{E}\left(\mathcal{L}_{\lambda}(\bm{\theta})\right)$).} We first show that $\mathbb{E}\left(\mathcal{L}_{\lambda}(\bm{\theta})\right)$ is strongly convex. It suffices to show that $\nabla^2 \mathbb{E}\left(\mathcal{L}_{\lambda}(\bm{\theta})\right)$ is positive definite. Note that the $(i,j)$-th element of $\nabla^2 \mathbb{E}\left(\mathcal{L}_{\lambda}(\bm{\theta})\right)$, for $i\neq j$, is given as
\begin{align*}
   \left( \nabla^2 \mathbb{E}\left(\mathcal{L}_{\lambda}(\bm{\theta})\right)\right)_{ij}
    =  
    p\left[(F(\theta_{i}^\star-\theta_j^\star) g'(\theta_i-\theta_j)+
    F(\theta_{j}^\star-\theta_i^\star) g'(\theta_j-\theta_i)\right] \leq 0.
\end{align*}
Here the inequality follows from Assumption \ref{Ass1}. Further, the diagonal values of $\nabla^2 \mathbb{E}\left(\mathcal{L}_{\lambda}(\bm{\theta})\right)$ are given as
\begin{align*}
    \left( \nabla^2 \mathbb{E}\left(\mathcal{L}_{\lambda}(\bm{\theta})\right)\right)_{ii}
    = &  -\sum_{j \in [m]\setminus \{i\}}
    p\left[(F(\theta_{i}^\star-\theta_j^\star) g'(\theta_i-\theta_j)+
    F(\theta_{j}^\star-\theta_i^\star) g'(\theta_j-\theta_i)\right]
    + 2 \lambda \\
    = & -\sum_{j \in [m]\setminus \{i\}}
    \left( \nabla^2 \mathbb{E}\left(\mathcal{L}_{\lambda}(\bm{\theta})\right)\right)_{ij} + 2\lambda>0,
\end{align*}
for $i  \in [m]$. Finally, by the Gershgorin circle theorem, it follows that $\Lambda_i\left(
\nabla^2 \mathbb{E}\left(\mathcal{L}_{\lambda}(\bm{\theta})\right)
    \right) \geq 2\lambda$, where $\Lambda_i(\cdot)$ denotes the $i$-th largest eigenvalue of a matrix. Hence, $\mathbb{E}\left(\mathcal{L}_{\lambda}(\bm{\theta})\right)$ is strongly convex with respect to $\bm{\theta}$.

\noindent
\textbf{(Tail Bound for Smallest Non-zero Eigenvalue).} We consider the decomposition of $\nabla^2\mathcal{L}_{0}(\bm\theta)$ as $\nabla^2\mathcal{L}_{0}(\bm\theta) = \bm{D}(\bm\theta) + \bm{A}(\bm\theta)$, where $\bm{D}(\bm\theta)$ and $\bm{A}(\bm\theta)$ being the diagonal and off-diagonal parts of $\nabla^2\mathcal{L}_{0}(\bm\theta)$, respectively. Specifically, $A_{ij}(\bm\theta)$ and $D_{ii}(\bm{\theta})$ can be written as
\begin{align*}
 &   A_{ij}(\bm\theta) = \sum_{l=1}^L w_l
    a_{ij}^{(l)}\left(
\widetilde{z}_{ij}^{(l)}g'(\gamma_{ij})+(1-\widetilde{z}_{ij}^{(l)})
g'(-\gamma_{ij})
\right), \\
&D_{ii}(\bm{\theta})=-
    \sum_{l=1}^L \left\{
\sum_{j \in [m]\setminus \{i\}}a_{ij}^{(l)}w_l\left(\widetilde{z}_{ij}^{(l)}g'(\gamma_{ij})+(1-\widetilde{z}_{ij}^{(l)})
g'(-\gamma_{ij})
\right)
\right\} = -\sum_{j \neq i}A_{ij}(\bm\theta),
\end{align*}
where $\gamma_{ij}=\theta_i - \theta_j$. Next, we define a matrix $Q_{ijl}(\bm{\theta})$ as
\begin{align*}
    \big(Q_{ijl}(\bm{\theta})\big)_{(k,m)} =  
    \begin{cases}
-a_{ij}^{(l)}w_l\left(\widetilde{z}_{ij}^{(l)}g'(\gamma_{ij})+(1-\widetilde{z}_{ij}^{(l)})
g'(-\gamma_{ij})
\right), k=m=i \mbox{ or } k=m=j,\\
a_{ij}^{(l)}w_l\left(\widetilde{z}_{ij}^{(l)}g'(\gamma_{ij})+(1-\widetilde{z}_{ij}^{(l)})
g'(-\gamma_{ij})
\right), (k,m)=(i,j) \mbox{ or } (k,m)=(j,i),\\
0, \mbox{otherwise}.\\
    \end{cases}
\end{align*}
With this, $\nabla^2\mathcal{L}_{0}(\bm\theta)$ can be represented as $\nabla^2\mathcal{L}_{0}(\bm\theta)= \sum_{i < j}\sum_{l=1}^L Q_{ijl}(\bm{\theta})$. Here $Q_{ijl}(\bm{\theta})$ is a hermitian matrix. Next, we turn to bound the smallest non-zero eigenvalue of $\nabla^2\mathcal{L}_{0}(\bm\theta)$. Here the intuition is that $\mathbb{E}(\nabla^2\mathcal{L}_{0}(\bm\theta))$ is a Laplacian matrix with smallest eigenvalue being zero. However, $\nabla^2\mathcal{L}_{0}(\bm\theta)$ can be a non-positive semi-definite matrix since $\widetilde{z}_{ij}^{(l)}$ can takes negative values and $g'(x)$ is always negative due to the log-concave assumption on $F(x)$ (Assumption \ref{Ass1}). Therefore, for any positive constant $v>0$,
\begin{align*}
   & \mathbb{P}
    \left(
\Lambda_{min,\perp}
\left(
\nabla^2\mathcal{L}_{0}(\bm\theta)
\right)>v
    \right)=\mathbb{P}
    \left(
\Lambda_{max,\perp}
\left(-
\nabla^2\mathcal{L}_{0}(\bm\theta)
\right) <-v
    \right) 
    = 
    1-\mathbb{P}
    \left(
\Lambda_{max,\perp}
\left(-
\nabla^2\mathcal{L}_{0}(\bm\theta)
\right) >-v
    \right).
\end{align*}
Next, we turn to show that $\mathbb{P}\left(\Lambda_{max,\perp}\left(-\nabla^2\mathcal{L}_{0}(\bm\theta)\right) >-v\right)$ is exponentially small. By the assumption that $F(\cdot)$ is log-concave, we have $g'(x)<0$ for $x \in \mathbb{R}$. Therefore, 
\begin{align*}
    \mathbb{E}
    \left(-A_{ij}(\bm{\theta})\right) = -p
    \left(
F(\theta_i^\star-\theta_j^\star)g'(\gamma_{ij})+
F(\theta_j^\star-\theta_i^\star)g'(-\gamma_{ij})
    \right)>0.
\end{align*}
Therefore, $\mathbb{E}\left(\nabla^2\mathcal{L}_{0}(\bm\theta)\right)$ is a positive semi-definite matrix regardless of the value of  $\bm{\theta}$. Furthermore,
\begin{align*}
    &\mathbb{P}
    \left(
\Lambda_{max,\perp}
\left(-
\nabla^2\mathcal{L}_{0}(\bm\theta)
\right) >-v
    \right)\\
    =&
        \mathbb{P}
    \left(
\Lambda_{max,\perp}
\left(-
\nabla^2\mathcal{L}_{0}(\bm\theta)+\mathbb{E}\left(\nabla^2\mathcal{L}_{0}(\bm\theta)\right)-\mathbb{E}\left(\nabla^2\mathcal{L}_{0}(\bm\theta)\right)
\right) >-v
    \right) \\
    \leq &
            \mathbb{P}
    \left(
\Lambda_{max}
\left(-
\nabla^2\mathcal{L}_{0}(\bm\theta)+\mathbb{E}\left(\nabla^2\mathcal{L}_{0}(\bm\theta)\right) \right)+
\Lambda_{max,\perp} \left(
-\mathbb{E}\left(\nabla^2\mathcal{L}_{0}(\bm\theta)\right)
\right) >-v
    \right) \\
    =& \mathbb{P}
    \left(
\Lambda_{max}
\left(-
\nabla^2\mathcal{L}_{0}(\bm\theta)+\mathbb{E}\left(\nabla^2\mathcal{L}_{0}(\bm\theta)\right) \right) >-v-
\Lambda_{max,\perp} \left(
-\mathbb{E}\left(\nabla^2\mathcal{L}_{0}(\bm\theta)\right)
\right)
    \right) \\
    =&
    \mathbb{P}
    \left(
\Lambda_{max}
\left(-
\nabla^2\mathcal{L}_{0}(\bm\theta)+\mathbb{E}\left(\nabla^2\mathcal{L}_{0}(\bm\theta)\right) \right) >-v+
\Lambda_{min,\perp} \left(
\mathbb{E}\left(\nabla^2\mathcal{L}_{0}(\bm\theta)\right)
\right)
    \right),
\end{align*}
where the inequality follows from the Weyl's inequality \citep{franklin2012matrix}, that is $\Lambda_{max,\perp}(\bm A+\bm B)\leq \Lambda_{max,\perp}(\bm A)+\Lambda_{max}(\bm B)$ for any squared matrices $\bm A$ and $\bm B$ satisfying $\bm{A} \bm{1}_m =\bm{B} \bm{1}_m= \bm{0}_m$.

To use the matrix Bernstein's inequality \citep{tropp2012user}, we first verify the corresponding conditions. The boundedness condition is satisfied as
\begin{align*}
\max_{l \in [L]}  \left|  \big(Q_{ijl}(\bm{\theta})\big)_{(i,j)} - \mathbb{E}
    \left( \big(Q_{ijl}(\bm{\theta})\big)_{(i,j)} \right) \right| 
     \leq \frac{C_{g',U}(q)}{G(\bm{\epsilon})} ,
\end{align*}
where $\big(Q_{ijl}(\bm{\theta})\big)_{(i,j)}$ denotes the $(i,j)$-th element of $Q_{ijl}(\bm{\theta})$. 

Note that $Q_{ijl}(\bm{\theta})$ can be written as
\begin{align*}
    Q_{ijl}(\bm{\theta}) = 
    -a_{ij}^{(l)}w_l\left(\widetilde{z}_{ij}^{(l)}g'(\gamma_{ij})+(1-\widetilde{z}_{ij}^{(l)})
g'(-\gamma_{ij})
\right) (\bm{e}_i-\bm{e}_j) (\bm{e}_i-\bm{e}_j)^T,
\end{align*}
where $\bm{e}_i$ is a zero vector except for the $i$-th element, which is 1. Therefore, we have
\begin{align*}
    \Lambda_{max}\left(
Q_{ijl}(\bm{\theta})-\mathbb{E}
    \left( Q_{ijl}(\bm{\theta}) \right)
    \right) \leq 
    \frac{2C_{g',U}(q) }{G(\bm{\epsilon})},
\end{align*}
for each $i,j \in [m]$ and $l \in [L]$. Furthermore,
\begin{align*}
&\left(\mathbb{E}
\left[Q_{ijl}(\bm{\theta})-\mathbb{E}
    \left( Q_{ijl}(\bm{\theta}) \right)\right]^T
    \left[Q_{ijl}(\bm{\theta})-\mathbb{E}
    \left( Q_{ijl}(\bm{\theta}) \right)\right]
\right)_{(i,i)} \\
\leq & \left(\mathbb{E}
\left[Q_{ijl}^2(\bm{\theta})\right]
\right)_{(i,i)}
 \leq  \frac{p
C_{g',U}^2(q) }{G^2(\bm{\epsilon})} \left(\frac{e^{\epsilon_l}-1}{e^{\epsilon_l}+1}\right)^2.
\end{align*}
Therefore, it follows that, for each $k \in [m]$,
\begin{align*}
&\mathbb{E}
\left(
    \sum_{i < j}\sum_{l=1}^L \left[Q_{ijl}(\bm{\theta})-\mathbb{E}
    \left( Q_{ijl}(\bm{\theta}) \right)\right]^T
    \left[Q_{ijl}(\bm{\theta})-\mathbb{E}
    \left( Q_{ijl}(\bm{\theta}) \right)\right]
    \right)_{(k,k)}  
    \leq 
    \frac{2 p m C^2_{g',U}(q)  }{G(\bm{\epsilon})} .
\end{align*}
Using the Gershgorin circle theorem, we have
\begin{align*}
    \Lambda_{max}\left(
\mathbb{E}
\left(
    \sum_{i < j}\sum_{l=1}^L \left[Q_{ijl}(\bm{\theta})-\mathbb{E}
    \left( Q_{ijl}(\bm{\theta}) \right)\right]^T
    \left[Q_{ijl}(\bm{\theta})-\mathbb{E}
    \left( Q_{ijl}(\bm{\theta}) \right)\right]
    \right)
    \right)\leq \frac{4 p m C^2_{g',U}(q)  }{G(\bm{\epsilon})}.
\end{align*}
For $0<t \leq  2 p m C_{g',U}(q) $, we have
\begin{align}
\label{Eigen_Bound}
    \mathbb{P}
    \left\{
\Lambda_{max}
\left(-
\nabla^2\mathcal{L}_{0}(\bm\theta)+\mathbb{E}\left(\nabla^2\mathcal{L}_{0}(\bm\theta)\right)
\right) >t
    \right\} \leq m \exp\left(-\frac{3t^2G(\bm{\epsilon})}{32 p m C_{g',U}^2(q)  }\right).
\end{align}
Next, we turn to provide a lower bound for $\Lambda_{min,\perp} \left(
\mathbb{E}\left(\nabla^2\mathcal{L}_{0}(\bm\theta)\right)\right)$. For any $\bm{u}\in \mathbb{R}^m$ such that $\bm{1}_m^T \bm{u}=0$ and $\Vert\bm{u}\Vert_2 =1$, 
\begin{align*}
  & \min_{\bm{\theta} \in \mathcal{C}(q) }\bm{u}^T   \mathbb{E}\left(\nabla^2\mathcal{L}_{0}(\bm\theta)\right)\bm{u} 
  = \min_{\bm{\theta} \in \mathcal{C}(q) } -p\sum_{i<j}
  \left[F(\theta_i^\star-\theta_j^\star)g'(\gamma_{ij})+(1-F(\theta_i^\star-\theta_j^\star))
g'(-\gamma_{ij}) 
\right](u_i-u_j)^2 \\
=&\min_{\bm{\theta} \in \mathcal{C} }
-\frac{p}{2}\sum_{i \neq j}
  \left[F(\theta_i^\star-\theta_j^\star)g'(\gamma_{ij})+(1-F(\theta_i^\star-\theta_j^\star))
g'(-\gamma_{ij}) 
\right](u_i^2-2u_iu_j+u_j^2) \\
\geq &
C_{g',L}(q) \frac{p}{2}\sum_{i \neq j}(u_i^2-2u_iu_j+u_j^2)  =C_{g',L}(q) (m-1)p \geq \frac{C_{g',L} (q)mp}{2}.
\end{align*}
This implies that $\Lambda_{min,\perp} \left(
\mathbb{E}\left(\nabla^2\mathcal{L}_{0}(\bm\theta)\right)\right) \geq \frac{C_{g',L}(q) mp}{2}$ for each $\bm{\theta} \in \mathcal{C}(q) $. This combined with (\ref{Eigen_Bound}) yields that
\begin{align*}
&\mathbb{P}
    \left\{
\Lambda_{max}
\left(-
\nabla^2\mathcal{L}_{0}(\bm\theta)+\mathbb{E}\left(\nabla^2\mathcal{L}_{0}(\bm\theta)\right) \right) >-v+
\Lambda_{min,\perp} \left(
\mathbb{E}\left(\nabla^2\mathcal{L}_{0}(\bm\theta)\right)
\right)
    \right\} \\
    \leq &
    m \exp\left(-\frac{3(C_{g',L}(q) mp/2-v)^2LB(\bm{\epsilon})}{32 p m  C_{g',U}^2(q)  }\right).
\end{align*}
where $v$ needs to satisfy
$
0<C_{g',L}(q) Lpm/2-v \leq 2 p m C_{g',U}(q) $. Therefore, we let $v = C_{g',L}(q) mp/4$. It then follows that
\begin{align*}
    \mathbb{P}
    \left(
\Lambda_{min,\perp}
\left(
\nabla^2\mathcal{L}_{0}(\bm\theta)
\right)>C_{g',L}(q) mp/4
    \right)
    \geq 1-
    m \exp\left(-\frac{3C_{g',L}^2(q) mLpB(\bm{\epsilon})}{128  C_{g',U}^2(q)  }\right).
\end{align*}
This completes the proof.
\qed \\

\begin{lemma}
	\label{lemma:inequ}
 Define $\mathcal{C}(q)=\{\bm{\theta}: \Vert \bm{\theta} - \bm{\theta}^\star\Vert_{\infty} \leq q \}$. Under Assumptions \ref{Ass1} and \ref{Ass2}, for any $\bm{\theta}\in \mathcal{C}(q)$, it holds true that
	\begin{align*}
 \sum_{l=1}^L
\mathrm{Var}\big(\mathcal{L}_{ij}(\bm\theta,\bm{z}^{(l)})-\mathcal{L}_{ij}(\bm\theta^\star,\bm{z}^{(l)})\big) 
\le  \sum_{l=1}^L\frac{2C_{g,U}^2(q)}{C_{f,L}^2(q) G(\bm{\epsilon})} \mathbb{E}\big(\mathcal{L}_{ij}(\bm\theta,\bm{z}^{(l)})-\mathcal{L}_{ij}(\bm\theta^\star,\bm{z}^{(l)})\big).
  \end{align*}
  for each $i<j$ and $l\in[L]$, where $C_{g,U}(q)=\max_{x \in [-\kappa-2q,\kappa+2q]} g(x)$ and $C_{f,L}(q) = \min_{x \in [-\kappa-2q,\kappa+2q]}f(x)$.
\end{lemma}

\noindent \textbf{Proof of Lemma \ref{lemma:inequ}.} Note that  
\begin{align*}
&\ \sum_{l=1}^L\mathbb{E}\big(\mathcal{L}_{ij}(\bm\theta,\bm{z}^{(l)})-\mathcal{L}_{ij}(\bm\theta^\star,\bm{z}^{(l)})\big) 
=  p
\mathbb{E}\left\{ 
\left[\overline{z}_{ij} \log \left(\frac{F(\theta_i^\star-\theta_j^\star)}{F(\theta_i-\theta_j)}\right)+(1-\overline{z}_{ij})\log\left(\frac{1-F(\theta_i^\star-\theta_j^\star)}{1-F(\theta_i-\theta_j)}\right)\right]
\right\} \notag \\
=&
p\left[F(\theta_i^\star-\theta_j^\star) \log \left(\frac{F(\theta_i^\star-\theta_j^\star)}{F(\theta_i-\theta_j)}\right)+(1-F(\theta_i^\star-\theta_j^\star))\log\left(\frac{1-F(\theta_i^\star-\theta_j^\star)}{1-F(\theta_i-\theta_j)}\right)\right] \notag \\
=&
-2p\left[F(\theta_i^\star-\theta_j^\star) \log \left(\sqrt{\frac{F(\theta_i-\theta_j)}{F(\theta_i^\star-\theta_j^\star)}}\right)+(1-F(\theta_i^\star-\theta_j^\star))\log\left(\sqrt{\frac{1-F(\theta_i-\theta_j)}{1-F(\theta_i^\star-\theta_j^\star)}}\right)\right],
\end{align*}
where $\overline{z}_{ij} = \sum_{l=1}^L w_l z_{ij}^{(l)}$. 

Subsequently, we provide a lower bound for $\sum_{l=1}^L\mathbb{E}\big(\mathcal{L}_{ij}(\bm\theta,\bm{z}^{(l)})-\mathcal{L}_{ij}(\bm\theta^\star,\bm{z}^{(l)})\big)$. Next, using the fact that $-\log(x)\geq -(x-1)$ for any $x>0$, we have
\begin{align}
    \label{Gasms}
&\sum_{l=1}^L \mathbb{E}\big(\mathcal{L}_{ij}(\bm\theta,\bm{z}^{(l)})-\mathcal{L}_{ij}(\bm\theta^\star,\bm{z}^{(l)})\big)\notag \\
    \ge & -
2p\left[
F(\theta_i^\star-\theta_j^\star)
\left(\sqrt{\frac{F(\theta_i-\theta_j)}{F(\theta_i^\star-\theta_j^\star)}}-1\right)+
(1-F(\theta_i^\star-\theta_j^\star))
\left(
\sqrt{\frac{1-F(\theta_i-\theta_j)}{1-F(\theta_i^\star-\theta_j^\star)}}-1
\right)
\right]
\notag \\
=&\ 2p-2p
\sqrt{F(\theta_i-\theta_j)F(\theta_i^\star-\theta_j^\star)}-
2p\sqrt{(1-F(\theta_i-\theta_j))(1-F(\theta_i^\star-\theta_j^\star))}
\notag \\
=&p\left[ \left(\sqrt{F(\theta_i-\theta_j)}-\sqrt{F(\theta_i^\star-\theta_j^\star)}\right)^2+\left(\sqrt{1-F(\theta_i-\theta_j)}-\sqrt{1-F(\theta_i^\star-\theta_j^\star)}\right)^2
\right]
,
\end{align}

Next, let $\bm\gamma=(\gamma_{ij})_{i<j}$ with $\gamma_{ij}=\theta_i-\theta_j$ for each $i<j$. Then the derivative of $\frac{\partial \mathcal{L}_{ij}(\bm\theta,\bm{z}^{(l)})}{\partial\gamma_{ij}}$ with respect to $\gamma_{ij}$ is given as
\begin{align*}
    \frac{\partial \mathcal{L}_{ij}(\bm\theta,\bm{z}^{(l)})}{\partial \gamma_{ij}}  
    =&-a_{ij}^{(l)}  w_l
\left(\widetilde{z}_{ij}^{(l)} \frac{f(\gamma_{ij})}{F(\gamma_{ij})}-(1-\widetilde{z}_{ij}^{(l)})\frac{f(\gamma_{ij})}{1-F(\gamma_{ij})}\right),
\end{align*}
where $w_l = \frac{\left(\frac{e^{\epsilon_l}-1}{e^{\epsilon_l}+1}\right)^2}{\sum_{l=1}^L \left(\frac{e^{\epsilon_l}-1}{e^{\epsilon_l}+1}\right)^2}$. Next, by the mean value theorem, it follows that
\begin{align*}
    &|\mathcal{L}_{ij}(\bm\theta,\bm{z}^{(l)})-\mathcal{L}_{ij}(\bm\theta^\star,\bm{z}^{(l)})|  \leq 
    \max_{\bm{\theta} \in \mathcal{C}(q)}
    w_l
    \left|
    \widetilde{z}_{ij}^{(l)}
\frac{f(\gamma_{ij})}{F(\gamma_{ij})}
-(1-\widetilde{z}_{ij}^{(l)})
\frac{f(\gamma_{ij})}{1-F(\gamma_{ij})}
    \right| \cdot |\gamma_{ij}-\gamma_{ij}^\star| \\
    \leq & 
    \frac{ \left(\frac{e^{\epsilon_l}-1}{e^{\epsilon_l}+1}\right)C_{g,U}(q)}{\sum_{l=1}^L \left(\frac{e^{\epsilon_l}-1}{e^{\epsilon_l}+1}\right)^2}|\gamma_{ij}-\gamma_{ij}^\star|
    \triangleq \frac{\frac{e^{\epsilon_l}-1}{e^{\epsilon_l}+1} C_{g,U}(q) |\gamma_{ij}-\gamma_{ij}^\star|}{LB(\bm{\epsilon})},
\end{align*}
where the first inequality follows from the fact that $\widetilde{z}_{ij}^{(l)} \leq \frac{e^{\epsilon_l}+1}{e^{\epsilon_l}-1}$ for any $i\neq j$ and $j\in [L]$ and $B(\bm{\epsilon}) = \frac{1}{L}\sum_{l=1}^L \left(\frac{e^{\epsilon_l}-1}{e^{\epsilon_l}+1}\right)^2$. Therefore, we have $$\sum_{l=1}^L \mathbb{E}\big(\mathcal{L}_{ij}(\bm\theta,\bm{z}^{(l)})-\mathcal{L}_{ij}(\bm\theta^\star,\bm{z}^{(l)})\big)^2 \le \frac{pC_{g,U}^2(q)(\gamma_{ij}-\gamma_{ij}^\star)^2}{LB(\bm{\epsilon})}.$$ Next, by the assumption that $F(\cdot)$ is the CDF of a zero-symmetric random variable, we have
\begin{align}
\label{Gmssa}
\big|\gamma_{ij}-\gamma_{ij}^\star\big|= &
\frac{1}{2}\left|
F^{-1}(F(\gamma_{ij}))-F^{-1}(F(\gamma_{ij}^\star))\right|+
\frac{1}{2}\left|
F^{-1}(F(-\gamma_{ij}))-F^{-1}(F(-\gamma_{ij}^\star))\right| \notag \\
\leq &
\frac{1}{2C_{f,L}(q)}
\left[
\left|
F(\gamma_{ij}) - F(\gamma_{ij}^\star)\right|+
\left|1-F(\gamma_{ij}) - (1-F(\gamma_{ij}^\star))
\right|\right] \notag \\
\leq &
\frac{1}{C_{f,L}(q)}
\left[
\left|
\sqrt{F(\gamma_{ij})} - \sqrt{F(\gamma_{ij}^\star)}\right|+
\left|\sqrt{1-F(\gamma_{ij})} - \sqrt{1-F(\gamma_{ij}^\star)}
\right|\right],
\end{align}
where the first inequality follows from the fact that $\frac{dF^{-1}(x)}{dx}=\frac{1}{f(F^{-1}(x))}$, the second inequality follows from the fact that $\max_{x \in \mathbb{R}}F(\gamma) \leq 1$ for any $x \in \mathbb{R}$, and $C_{f,L}(q) = \min \limits_{x \in [-\kappa-2q,\kappa+2q]}f(x)$. Subsequently, combining (\ref{Gasms}) and (\ref{Gmssa}) yields that
\begin{align*}
   & \sum_{l=1}^L \mathbb{E}\big(\mathcal{L}_{ij}(\bm\theta,\bm{z}^{(l)})-\mathcal{L}_{ij}(\bm\theta^\star,\bm{z}^{(l)})\big)^2 \\
    \leq &\frac{2pC_{g,U}^2(q)}{C_{f,L}^2(q)LB(\bm{\epsilon})}
    \left\{
\left(
\sqrt{F(\gamma_{ij})} - \sqrt{F(\gamma_{ij}^\star)}\right)^2+
\left(\sqrt{1-F(\gamma_{ij})} - \sqrt{1-F(\gamma_{ij}^\star)}
\right)^2\right\} \\
\leq & \frac{2C_{g,U}^2(q)}{C_{f,L}^2(q) LB(\bm{\epsilon})} \sum_{l=1}^L\mathbb{E}\big(\mathcal{L}_{ij}(\bm\theta,\bm{z}^{(l)})-\mathcal{L}_{ij}(\bm\theta^\star,\bm{z}^{(l)})\big).
\end{align*}
Finally, we have the relation between the variance and the mean of $\mathcal{L}_{ij}(\bm\theta,\bm{z})-\mathcal{L}_{ij}(\bm\theta^\star,\bm{z})$ as
\begin{align}
\label{inequ: Var}
&\sum_{l=1}^L\text{Var}\big(\mathcal{L}_{ij}(\bm\theta,\bm{z}^{(l)})-\mathcal{L}_{ij}(\bm\theta^\star,\bm{z}^{(l)})\big) \le \sum_{l=1}^L\mathbb{E}\big(\mathcal{L}_{ij}(\bm\theta,\bm{z})-\mathcal{L}_{ij}(\bm\theta^\star,\bm{z})\big)^2\notag\\
\le& \sum_{l=1}^L \frac{2C_{g,U}^2(q)}{C_{f,L}^2(q) LB(\bm{\epsilon})} \mathbb{E}\big(\mathcal{L}_{ij}(\bm\theta,\bm{z}^{(l)})-\mathcal{L}_{ij}(\bm\theta^\star,\bm{z}^{(l)})\big).
\end{align}
This completes the proof. \qed \\

\begin{lemma}
\label{Lemma:BTLHessian}
    If $F(x)=1/(1+e^{-x})$, for any $\bm{\theta}\in \mathcal{C}(q)$, it holds true that
    \begin{align}
    \label{Eqn:ConvBTL}
    \mathbb{P}
    \left(
\Lambda_{min,\perp}
\left(
\nabla^2\mathcal{L}_{0}(\bm\theta)
\right)> \frac{C_0(q)mp}{2}
    \right)
    \geq 1-
    m \exp\left(-\frac{C_{0}^2(q) mLpB(\bm{\epsilon})}{32}\right),
    \end{align}
    where $\mathcal{C}(q)= \{\bm{\theta}:\Vert\bm{\theta} - \bm{\theta}^\star \Vert_{\infty} \leq q\}$, $\Lambda_{min,\perp}(\cdot)$ denotes the smallest non-zero eigenvalue, $C_0(q) = \frac{e^{\kappa+2q}}{4(1+e^{\kappa+2q})^2}$, and $B(\bm \epsilon)=L^{-1}\sum_{l=1}^L \left(\frac{e^{\epsilon_l}-1}{e^{\epsilon_l}+1} \right)^2$.
\end{lemma}

\noindent \textbf{Proof of Lemma \ref{Lemma:BTLHessian}.}  Similar to the proof of Lemma \ref{Lemma:Hessian}, we consider the decomposition of $\nabla^2\mathcal{L}_{0}(\bm\theta)$ as $\nabla^2\mathcal{L}_{0}(\bm\theta) = \bm{D}(\bm\theta) + \bm{A}(\bm\theta)$ under the BTL model, where $\bm{D}(\bm\theta)$ and $\bm{A}(\bm\theta)$ being the diagonal and off-diagonal parts of $\nabla^2\mathcal{L}_{0}(\bm\theta)$, respectively. Specifically, when $F(x)=1/(1+e^{-x})$, $A_{ij}(\bm\theta)$ and $D_{ii}(\bm{\theta})$ can be written as
\begin{align*}
 &   A_{ij}(\bm\theta) = \sum_{l=1}^L w_l
    a_{ij}^{(l)}\left(
\widetilde{z}_{ij}^{(l)}g'(\gamma_{ij})+(1-\widetilde{z}_{ij}^{(l)})
g'(-\gamma_{ij})
\right) = -\sum_{l=1}^L w_l
    a_{ij}^{(l)} \frac{e^{\gamma_{ij}}}{(1+e^{\gamma_{ij}})^2}, \\
&D_{ii}(\bm{\theta})=-
    \sum_{l=1}^L \left\{
\sum_{j \in [m]\setminus \{j\}}a_{ij}^{(l)}w_l\left(\widetilde{z}_{ij}^{(l)}g'(\gamma_{ij})+(1-\widetilde{z}_{ij}^{(l)})
g'(-\gamma_{ij})
\right)
\right\} = -\sum_{j \neq i}A_{ij}(\bm\theta),
\end{align*}
where $\gamma_{ij}=\theta_i - \theta_j$. Next, we define a matrix $Q_{ijl}(\bm{\theta})$ as
\begin{align*}
    \big(Q_{ijl}(\bm{\theta})\big)_{km} =  
    \begin{cases}
-w_l a_{ij}^{(l)}\frac{e^{\gamma_{ij}}}{(1+e^{\gamma_{ij}})^2}, k=m=i \mbox{ or } k=m=j,\\
w_l a_{ij}^{(l)}\frac{e^{\gamma_{ij}}}{(1+e^{\gamma_{ij}})^2}, (k,m)=(i,j) \mbox{ or } (k,m)=(j,i),\\
0, \mbox{otherwise}.\\
    \end{cases}
\end{align*}
With this, $\nabla^2\mathcal{L}_{0}(\bm\theta)$ can be represented as $\nabla^2\mathcal{L}_{0}(\bm\theta)= \sum_{i < j}\sum_{l=1}^L Q_{ijl}(\bm{\theta})$, where $Q_{ijl}(\bm{\theta})$ is a hermitian matrix. Next, we turn to bound the smallest non-zero eigenvalue of $\nabla^2\mathcal{L}_{0}(\bm\theta)$. Here the intuition is that $\mathbb{E}(\nabla^2\mathcal{L}_{0}(\bm\theta))$ is a Laplacian matrix with smallest eigenvalue being zero. For any positive constant $v>0$,
\begin{align*}
   & \mathbb{P}
    \left(
\Lambda_{min,\perp}
\left(
\nabla^2\mathcal{L}_{0}(\bm\theta)
\right)>v
    \right)=\mathbb{P}
    \left(
\Lambda_{max,\perp}
\left(-
\nabla^2\mathcal{L}_{0}(\bm\theta)
\right) <-v
    \right) 
    = 
    1-\mathbb{P}
    \left(
\Lambda_{max,\perp}
\left(-
\nabla^2\mathcal{L}_{0}(\bm\theta)
\right) >-v
    \right).
\end{align*}
Next, we turn to show that $\mathbb{P}\left(\Lambda_{max,\perp}\left(-\nabla^2\mathcal{L}_{0}(\bm\theta)\right) >-v\right)$ is exponentially small. We first note that $\mathbb{E}
    \left(-A_{ij}(\bm{\theta})\right) = \frac{pe^{\gamma_{ij}}}{(1+e^{\gamma_{ij}})^2} > 0$. Therefore, $\mathbb{E}\left(\nabla^2\mathcal{L}_{0}(\bm\theta)\right)$ is a positive semi-definite matrix regardless of the value of  $\bm{\theta}$. Furthermore,
\begin{align*}
    &\mathbb{P}
    \left(
\Lambda_{max,\perp}
\left(-
\nabla^2\mathcal{L}_{0}(\bm\theta)
\right) >-v
    \right)\\
    =&
        \mathbb{P}
    \left(
\Lambda_{max,\perp}
\left(-
\nabla^2\mathcal{L}_{0}(\bm\theta)+\mathbb{E}\left(\nabla^2\mathcal{L}_{0}(\bm\theta)\right)-\mathbb{E}\left(\nabla^2\mathcal{L}_{0}(\bm\theta)\right)
\right) >-v
    \right) \\
    \leq &
            \mathbb{P}
    \left(
\Lambda_{max}
\left(-
\nabla^2\mathcal{L}_{0}(\bm\theta)+\mathbb{E}\left(\nabla^2\mathcal{L}_{0}(\bm\theta)\right) \right)+
\Lambda_{max,\perp} \left(
-\mathbb{E}\left(\nabla^2\mathcal{L}_{0}(\bm\theta)\right)
\right) >-v
    \right) \\
    =& \mathbb{P}
    \left(
\Lambda_{max}
\left(-
\nabla^2\mathcal{L}_{0}(\bm\theta)+\mathbb{E}\left(\nabla^2\mathcal{L}_{0}(\bm\theta)\right) \right) >-v-
\Lambda_{max,\perp} \left(
-\mathbb{E}\left(\nabla^2\mathcal{L}_{0}(\bm\theta)\right)
\right)
    \right) \\
    =&
    \mathbb{P}
    \left(
\Lambda_{max}
\left(-
\nabla^2\mathcal{L}_{0}(\bm\theta)+\mathbb{E}\left(\nabla^2\mathcal{L}_{0}(\bm\theta)\right) \right) >-v+
\Lambda_{min,\perp} \left(
\mathbb{E}\left(\nabla^2\mathcal{L}_{0}(\bm\theta)\right)
\right)
    \right),
\end{align*}
where the inequality follows from the Weyl's inequality \citep{franklin2012matrix}, that is $\Lambda_{max,\perp}(\bm A+\bm B)\leq \Lambda_{max,\perp}(\bm A)+\Lambda_{max}(\bm B)$ for any squared matrices $\bm A$ and $\bm B$ such that $\bm{A} \bm{1}_m =\bm{B} \bm{1}_m= \bm{0}_m$.

To use the matrix Bernstein's inequality \citep{tropp2012user}, we first verify the corresponding conditions. The boundedness condition is satisfied as
\begin{align*}
\max_{\bm{\theta} \in \mathcal{C}(q)}\max_{l \in [L]}  \left|  \big(Q_{ijl}(\bm{\theta})\big)_{(i,j)} - \mathbb{E}
    \left( \big(Q_{ijl}(\bm{\theta})\big)_{(i,j)} \right) \right| 
     \leq \frac{1}{4G(\bm{\epsilon})},
\end{align*}
where $\big(Q_{ijl}(\bm{\theta})\big)_{(i,j)}$ denotes the $(i,j)$-th element of $Q_{ijl}(\bm{\theta})$.

Note that $Q_{ijl}(\bm{\theta})$ can be rewritten as
\begin{align*}
    Q_{ijl}(\bm{\theta}) = 
    \frac{a_{ij}^{(l)}w_le^{\gamma_{ij}}}{(1+e^{\gamma_{ij}})^2} (\bm{e}_i-\bm{e}_j) (\bm{e}_i-\bm{e}_j)^T,
\end{align*}
where $\bm{e}_i$ is a zero vector except for the $i$-th element, which is 1. Therefore, we have
\begin{align*}
    \Lambda_{max}\left(
Q_{ijl}(\bm{\theta})-\mathbb{E}
    \left( Q_{ijl}(\bm{\theta}) \right)
    \right) \leq 
    \frac{1}{2G(\bm{\epsilon})},
\end{align*}
for each $i,j \in [m]$ and $l \in [L]$. Furthermore,
\begin{align*}
\left(\mathbb{E}
\left[Q_{ijl}(\bm{\theta})-\mathbb{E}
    \left( Q_{ijl}(\bm{\theta}) \right)\right]^T
    \left[Q_{ijl}(\bm{\theta})-\mathbb{E}
    \left( Q_{ijl}(\bm{\theta}) \right)\right]
\right)_{(i,i)} 
 \leq  \frac{p}{G^2(\bm{\epsilon})} \left(\frac{e^{\epsilon_l}-1}{e^{\epsilon_l}+1}\right)^2.
\end{align*}
Therefore, it follows that, for each $k \in [m]$,
\begin{align*}
\mathbb{E}
\left(
    \sum_{i < j}\sum_{l=1}^L \left[Q_{ijl}(\bm{\theta})-\mathbb{E}
    \left( Q_{ijl}(\bm{\theta}) \right)\right]^T
    \left[Q_{ijl}(\bm{\theta})-\mathbb{E}
    \left( Q_{ijl}(\bm{\theta}) \right)\right]
    \right)_{(k,k)}  
    \leq 
    \frac{mp}{G(\bm{\epsilon})}. 
\end{align*}
Using the Gershgorin circle theorem \citep{horn2012matrix}, we have
\begin{align*}
    \Lambda_{max}\left(
\mathbb{E}
\left(
    \sum_{i < j}\sum_{l=1}^L \left[Q_{ijl}(\bm{\theta})-\mathbb{E}
    \left( Q_{ijl}(\bm{\theta}) \right)\right]^T
    \left[Q_{ijl}(\bm{\theta})-\mathbb{E}
    \left( Q_{ijl}(\bm{\theta}) \right)\right]
    \right)
    \right)\leq \frac{2mp}{G(\bm{\epsilon})}.
\end{align*}

If $0<t \leq  2mp $, we have
\begin{align}
\label{Eigen_Bound_1}
    \mathbb{P}
    \left\{
\Lambda_{max}
\left(-
\nabla^2\mathcal{L}_{0}(\bm\theta)+\mathbb{E}\left(\nabla^2\mathcal{L}_{0}(\bm\theta)\right)
\right) >t
    \right\} \leq m \exp\left(-\frac{3t^2 G(\bm{\epsilon})}{8 mp}\right).
\end{align}
Next, we turn to provide a lower bound for $\Lambda_{min,\perp} \left(
\mathbb{E}\left(\nabla^2\mathcal{L}_{0}(\bm\theta)\right)\right)$. For any $\bm{u}\in \mathbb{R}^m$ such that $\bm{1}_m^T \bm{u}=0$ and $\Vert\bm{u}\Vert_2 =1$, 
\begin{align*}
  & \min_{\bm{\theta} \in \mathcal{C}(q)}\bm{u}^T   \mathbb{E}\left(\nabla^2\mathcal{L}_{0}(\bm\theta)\right)\bm{u} 
  = \min_{\bm{\theta} \in \mathcal{C}(q)} -p\sum_{i<j} 
  \left[F(\theta_i^\star-\theta_j^\star)g'(\gamma_{ij})+(1-F(\theta_i^\star-\theta_j^\star))
g'(-\gamma_{ij}) 
\right](u_i-u_j)^2 \\
=&
\frac{p}{2}\sum_{i \neq j} 
 \frac{e^{\kappa+2q}}{(1+e^{\kappa+2q})^2}
 (u_i^2-2u_iu_j+u_j^2) 
\geq 
\frac{pe^{\kappa+2q}}{2(1+e^{\kappa+2q})^2}\sum_{i \neq j}(u_i^2-2u_iu_j+u_j^2) \geq
\frac{mpe^{\kappa+2q}}{4(1+e^{\kappa+2q})^2},
\end{align*}
where the last inequality follows from $m \geq 2$.
This implies that $\Lambda_{min,\perp} \left(
\mathbb{E}\left(\nabla^2\mathcal{L}_{0}(\bm\theta)\right)\right) \geq C_0(q) mp$ for any $\bm{\theta} \in \mathcal{C}(q)$ with $C_0(q) = \frac{e^{\kappa+2q}}{4(1+e^{\kappa+2q})^2}$. Setting $v = C_0(q)mp/2$, we have $C_0(q) mp-v \leq 2mp$. This then leads to
\begin{align*}
&\mathbb{P}
    \left\{
\Lambda_{max}
\left(-
\nabla^2\mathcal{L}_{0}(\bm\theta)+\mathbb{E}\left(\nabla^2\mathcal{L}_{0}(\bm\theta)\right) \right) >-v+
\Lambda_{min,\perp} \left(
\mathbb{E}\left(\nabla^2\mathcal{L}_{0}(\bm\theta)\right)
\right)
    \right\} \\
    \leq &
    m \exp\left(-\frac{3(C_0(q)mp-v)^2 G(\bm{\epsilon})}{8 mp}\right).
\end{align*}
In conclusion, we have the following results:
\begin{align*}
    \mathbb{P}
    \left(
\Lambda_{min,\perp}
\left(
\nabla^2\mathcal{L}_{0}(\bm\theta)
\right)> \frac{e^{\kappa+2q}mp}{8(1+e^{\kappa+2q})^2}
    \right)
    \geq 1-
    m \exp\left(-\frac{3C_0(q)^2mpLB(\bm{\epsilon})}{32}\right).
\end{align*}
 This completes the proof. \qed \\ 

\begin{lemma}
\label{lemma: Esup}
Let $\bm{z}=\{\bm{z}^{(l)}\}_{l=1}^L$. Define
\begin{align*}
&\nu(\bm\theta,\bm{z})=
\frac{1}{m^2G(\bm{\epsilon})}
   \sum_{i<j}\sum_{l=1}^L\big(f_{ijl}(\bm\theta,\bm{z})-\mathbb{E}(f_{ijl}(\bm\theta,\bm{z})\big), \\
&f_{ijl}(\bm{\theta},\bm{z}) =  a_{ij}^{(l)}w_lG(\bm{\epsilon})
    \left(
    \widetilde{z}_{ij}^{(l)}
\log \frac{F(\gamma_{ij})}{F(\gamma_{ij}^\star)}
+(1-\widetilde{z}_{ij}^{(l)})
\log  \frac{F(-\gamma_{ij})}{F(-\gamma_{ij}^\star)}
    \right),
\end{align*}
where $G(\bm{\epsilon})=\sum_{l=1}^L \left(\frac{e^{\epsilon_l}-1}{e^{\epsilon_l}+1}\right)^2$. Under the assumptions of Theorem \ref{Thm:ConsisT1}, it holds true that 
$$\mathbb{E}\left(\sup_{\bm{\theta} \in S_{u}}\nu(\bm\theta,\bm{z})\right)\le \frac{M_{u}}{2},$$ 
for any $u \geq 1$, where $S_u$ is as defined in the proof of Theorem \ref{Thm:ConsisT1}.
\end{lemma}

\noindent\textbf{Proof of Lemma \ref{lemma: Esup}.} We first note that
\begin{align*}
    &\nu(\bm\theta,\bm{z}) = \frac{1}{m^2 }\sum_{i<j} \sum_{l=1}^L
    \left\{
    \mathcal{L}_{ij}(\bm{\theta}^\star,\bm{z}^{(l)}) -\mathcal{L}_{ij}(\bm{\theta},\bm{z}^{(l)})
    \right\}-
 \frac{1}{m^2}\sum_{i<j}\sum_{l=1}^L
   \mathbb{E} \left\{
    \mathcal{L}_{ij}(\bm{\theta}^\star,\bm{z}^{(l)}) -
    \mathcal{L}_{ij}(\bm{\theta},\bm{z}^{(l)})
    \right\}, \\
    &\sum_{l=1}^L \mathcal{L}_{ij}(\bm{\theta}^\star,\bm{z}^{(l)}) -\mathcal{L}_{ij}(\bm{\theta},\bm{z}^{(l)})= \frac{1}{G(\bm{\epsilon})}\sum_{l=1}^L f_{ijl}(\bm{\theta},\bm{z}).
\end{align*}

Let $\bm{z}'$ be an independent copy of $\bm{z}$ and $(\tau_{ijl})_{i<j,l}$ be independent Rademacher random variables. Then we have
\begin{align*}
\mathbb{E} \left(\sup_{\bm{\theta} \in S_{u}}\nu(\bm\theta,\bm{z})\right)
&=\frac{1}{m^2G(\bm{\epsilon})}\mathbb{E} \bigg(\sup_{\bm{\theta} \in S_{u}}\mathbb{E}\Big(\sum_{i<j}\sum_{l=1}^L\big( f_{ijl}(\bm\theta,\bm{z})- f_{ijl}(\bm\theta,\bm{z}')\big)\Big|\bm{z}\Big)\bigg)\\
&\le \frac{1}{m^2G(\bm{\epsilon})}\mathbb{E}\Big(\sup_{\bm{\theta} \in S_{u}}\sum_{i<j}\sum_{l=1}^L\big( f_{ijl}(\bm\theta,\bm{z})- f_{ijl}(\bm\theta,\bm{z}')\big)\Big)\\
&=\frac{1}{m^2G(\bm{\epsilon})}\mathbb{E}\Big(\sup_{\bm{\theta} \in S_{u}}\sum_{i<j}\sum_{l=1}^L\tau_{ijl}\big( f_{ijl}(\bm\theta,\bm{z})- f_{ijl}(\bm\theta_0,\bm{z})+ f_{ijl}(\bm\theta_0,\bm{z}')- f_{ijl}(\bm\theta,\bm{z}')\big)\Big)\\
&\le \frac{2}{m^2G(\bm{\epsilon})}\mathbb{E} \Big(\sup_{\bm{\theta} \in S_{u}}\Big|\sum_{i<j}\sum_{l=1}^L\tau_{ijl}\big( f_{ijl}(\bm\theta,\bm{z})- f_{ijl}(\bm\theta_0,\bm{z})\big)\Big|\Big),
\end{align*}
where $\bm\theta_0\in S_u$, the first inequality follows from the Jensen's inequality, and the second equality follows from the standard symmetrization argument \citep{koltchinskii2011oracle}.

Note that conditional on $\bm{z}$, $\frac{1}{m\sqrt{L}}\sum_{i<j,l}\tau_{ijl} f_{ijl}(\bm\theta,\bm{z})$ is a sub-Gaussian process with respect to $d$, where 
\begin{align*}
d^2\big(\bm{\theta}^{(1)},\bm{\theta}^{(2)}\big)=\frac{1}{m^2L}\sum_{i<j}\sum_{l=1}^L\big( f_{ijl}(\bm\theta^{(1)},\bm{z})- f_{ijl}(\bm\theta^{(2)},\bm{z})\big)^2,
\end{align*}
for any $\bm{\theta}^{(1)},\bm{\theta}^{(2)} \in S_{u}$. It then follows from Theorem 3.1 of \cite{koltchinskii2011oracle} that 
\begin{align*}
\frac{1}{m\sqrt{L}}\mathbb{E}\Big(\sup_{\bm{\theta} \in S_{u}}\Big|\sum_{i<j}\sum_{l=1}^L\tau_{ijl} f_{ijl}\big((\bm\theta,\bm{z})- f_{ijl}(\bm\theta_0,\bm{z})\big)\Big|\Big)\lesssim \mathbb{E}\Big(\int_0^{D(S_{u})}H^{1/2}\big(S_{u}, d, \eta\big)d\eta\Big),
\end{align*}
where $D(S_{u})$ is the diameter of $S_{u}$ with respect to $d$, and $H(S_{u}, d, \eta)$ is the $\eta$-entropy of $(S_{u},d)$. Next, we establish the quantitative relationship between $d^2\big(\bm{\theta}^{(1)},\bm{\theta}^{(2)}\big)$ and $\Vert \bm{\theta}^{(1)}-\bm{\theta}^{(2)}\Vert_2^2$ and bounding the diameter $\sup_{\bm{\theta}^{(1)},\bm{\theta}^{(2)}\in S_u}\mathbb{E}d^2\big(\bm{\theta}^{(1)},\bm{\theta}^{(2)}\big)$. It is worth noting that using (\ref{Gasms}), we can derive that
\begin{align*}
    \sup_{\bm{\theta}^{(1)},\bm{\theta}^{(2)}\in S_u}\mathbb{E}d^2\big(\bm{\theta}^{(1)},\bm{\theta}^{(2)}\big) \asymp
    \mathbb{E} \left[ \sup_{\bm{\theta}^{(1)},\bm{\theta}^{(2)}\in S_u} d^2\big(\bm{\theta}^{(1)},\bm{\theta}^{(2)}\big)\right].
\end{align*}

\noindent
\textbf{Step 1.} We first bound the diameter. 
\begin{align*}
\mathbb{E}d^2\big(\bm{\theta}^{(1)},\bm{\theta}^{(2)}\big)= &
\frac{p}{m^2L}\sum_{i<j}\sum_{l=1}^L
\mathbb{E}\left\{
    \left(\frac{e^{\epsilon_l}-1}{e^{\epsilon_l}+1}\right)^4
    \left(
    \widetilde{z}_{ij}^{(l)}
\log \frac{F(\gamma_{ij}^{(1)})}{F(\gamma_{ij}^{(2)})}
+(1-\widetilde{z}_{ij}^{(l)})
\log  \frac{F(-\gamma_{ij}^{(1)})}{F(-\gamma_{ij}^{(2)})}
    \right)^2\right\}  \\
    \leq &
    \sup_{\bm{\theta} \in S_u}
    \frac{4pC_{g,U}^2(q)}{m^2L}
    \sum_{l=1}^L 
    \left(\frac{e^{\epsilon_l}-1}{e^{\epsilon_l}+1}\right)^2
    \sum_{i<j}
    \left(
    \gamma_{ij}-\gamma_{ij}^\star
    \right)^2 \\
\leq &  \sup_{\bm{\theta} \in S_u} \frac{8C_{g,U}^2(q) B(\bm{\epsilon})}{C_{f,L}^2(q) m^2}
\sum_{i<j}\mathbb{E}\big(\mathcal{L}_{ij}(\bm\theta,\bm{z})-\mathcal{L}_{ij}(\bm\theta^\star,\bm{z})\big)
\\
  \leq &  
   \frac{C_{g,U}^2(q) B(\bm{\epsilon})2^{u+3}\delta_{m,L}}{C_{f,L}^2(q) }.
\end{align*}
Here the first inequality follows from the facts that $\log F(x)-\log F(y) \leq \max_{ z} g(z) \cdot |x-y|$ and $\widetilde{z}_{ij}^{(l)} \leq \frac{e^{\epsilon_l}+1}{e^{\epsilon_l}-1}$ for any $i \neq j$, and the second inequality follows by combining (\ref{Gasms}) and (\ref{Gmssa}). The last inequality follows from the definition of $S_u$.

\noindent
\textbf{Step 2.} Next, we bound the $d^2\big(\bm{\theta}^{(1)},\bm{\theta}^{(2)}\big)$ using $\Vert \bm{\theta}^{(1)} - \bm{\theta}^{(2)}\Vert_2^2$.
\begin{align*}
   d^2\big(\bm{\theta}^{(1)},\bm{\theta}^{(2)}\big) 
   \leq  &
   \frac{1}{m^2L}\sum_{i<j}\sum_{l=1}^L
       \left(\frac{e^{\epsilon_l}-1}{e^{\epsilon_l}+1}\right)^4
    \left(
    \widetilde{z}_{ij}^{(l)}
\log \frac{F(\gamma_{ij}^{(1)})}{F(\gamma_{ij}^{(2)})}
+(1-\widetilde{z}_{ij}^{(l)})
\log  \frac{F(-\gamma_{ij}^{(1)})}{F(-\gamma_{ij}^{(2)})}
    \right)^2 \\
    \leq &
   \frac{2C_{g,U}^2(q) }{m^2 L }\sum_{i<j}
   \sum_{l=1}^L
       \left(\frac{e^{\epsilon_l}-1}{e^{\epsilon_l}+1}\right)^2
    \left[ \left(\theta_i^{(1)}-\theta_i^{(2)}\right)^2+\left(\theta_j^{(1)}-\theta_j^{(2)}\right)^2 \right] \\
    \leq & \frac{2C_{g,U}^2(q) B(\bm{\epsilon})}{m} \Vert \bm{\theta}^{(1)} - \bm{\theta}^{(2)}\Vert_2^2,
\end{align*}
Here $\bm{\theta}^{(1)}$ and $\bm{\theta}^{(2)}$ both belong to $\mathcal{C}(q)$. Thus, one has $d^2\big(\bm{\theta}^{(1)},\bm{\theta}^{(2)}\big)\le \eta^2$ if $\|\bm\theta^{(1)}-\bm\theta^{(2)}\|_2^2\le \frac{m  \eta^2}{2C_{g,U}^2(q) B(\bm{\epsilon})}$. Note that for any $\bm\theta\in \mathcal{C}(q)$, we have $\|\bm\theta\|_2\le \sqrt{m}M$ for some $M>0$. This then leads to
\begin{align*}
H(S_{u}, d, \eta)
\leq H\left(B_2(m), \|\cdot\|_2, 
\frac{\eta}{\sqrt{2B(\bm{\epsilon})}C_{g,U}(q)M}
\right)
\leq m\log\left(\frac{3\sqrt{2B(\bm{\epsilon})}C_{g,U}(q)M}{\eta}\right),
\end{align*}
where $B_2(m)$ is the unit $l_2$-ball in $\mathbb{R}^{m}$. Then, applying the Dudley’s integral entropy bound \citep{koltchinskii2011oracle}, we have
\begin{align}
\label{Last}
\mathbb{E}\left(\sup_{S_{u}}\nu(\bm\theta;\bm{z})\right)&\lesssim \frac{\sqrt{L}}{mG(\bm{\epsilon})}\mathbb{E}\left(\int_0^{D(S_{u})}H^{1/2}\big(S_{u}, d, \eta\big)d\eta\right
)\notag \\
&\le 
\frac{\sqrt{L}}{m G(\bm{\epsilon})}\mathbb{E}\left(
\int_0^{D(S_{u})\wedge \sqrt{2B(\bm{\epsilon})} C_{g,U}M}\sqrt{m\log\left(\frac{3\sqrt{2B(\bm{\epsilon})}C_{g,U}(q)M}{\eta}\right)} 
 d\eta
 \right)
\notag \\
&\le 
\frac{\sqrt{L}}{mG(\bm{\epsilon})}
\int_0^{\mathbb{E}[D(S_{u})]\wedge 
\sqrt{2B(\bm{\epsilon})} C_{g,U}(q)M
}\sqrt{m\log\left(\frac{3\sqrt{2B(\bm{\epsilon})}C_{g,U}(q)M}{\eta}\right)} 
 d\eta \notag \\
 & \leq 
 \frac{ 3\sqrt{2}C_{g,U}(q)M}{\sqrt{mG(\bm{\epsilon})}}
\int_0^{
\frac{\sqrt{2^{u} \delta_{m,L}}}{C_{f,L}(q)M}
\wedge 
\frac{1}{3}
}\sqrt{\log\left(\frac{1}{t}\right)} 
 dt,
\end{align} 
where $x\wedge y=\min\{x,y\}$ for any $x,y\in\mathbb{R}$ and the third inequality follows from Jensen's inequality and the fact that $\int_{0}^{x}(\log(\frac{1}{\eta}))^{1/2}d\eta$ is concave with respect to $x$.

Next, we proceed to provide an upper bound for (\ref{Last}). For ease of notation, we denote that $C_1 =\frac{ 3\sqrt{2}C_{g,U}(q)M}{\sqrt{mG(\bm{\epsilon})}}$ and $C_2 =\frac{\sqrt{2^u \delta_{m,L}}}{C_{f,L}(q)M}$ . Then (\ref{Last}) becomes
\begin{align*}
C_1 \int_{0}^{C_2 \wedge 1/3} \sqrt{\log(1/t)}dt =
C_1 \int_{C_2^{-1} \vee 3}^{\infty} \frac{\sqrt{\log(s)}}{s^2}ds 
\leq 
\frac{C_1}{\sqrt{\log(C_2^{-1} \vee 3)}} \int_{C_2^{-1} \vee 3}^{\infty} \frac{\log(s)}{s^2}ds.
\end{align*}
By the fact that $\int_{a}^{\infty} \log(x)/x^2 dx = (\log(a)+1)/a$, we further have
\begin{align}
\label{Integral}
\frac{C_1}{\sqrt{\log(C_2^{-1} \vee 3)}} \int_{C_2^{-1} \vee 1}^{\infty} \frac{\log(s)}{s^2}ds 
=
\frac{C_1}{C_2^{-1} \vee 3}
\frac{\log(C_2^{-1} \vee 3)+1}{\sqrt{\log(C_2^{-1} \vee 3)}} 
\leq \frac{C_1 \sqrt{\log(C_2^{-1} \vee 3)}}{C_2^{-1} \vee 3},
\end{align}
where the last inequality follows from the fact that $1/x+x \leq 2x$ for $x \geq 1$. Substituting $C_1$ and $C_2$ into (\ref{Integral}) yields that
\begin{align*}
\frac{C_1}{C_2^{-1} \vee 3} \leq C_1 C_2 =3
\sqrt{\frac{2^{u+1}R_F^2(q) \delta_{m,L}}{m G(\bm{\epsilon})}}.
\end{align*}
where $R_F(q)=C_{g,U}(q)/C_{f,L}(q)$. Given that $\delta_{m,L} = o(1)$, one has $C_2 = o(1)$. Therefore, (\ref{Last}) is upper bounded as
\begin{align}
\label{Final}
(\ref{Last}) \lesssim
\sqrt{\frac{2^{u+1}R_F^2(q) \delta_{m,L}}{m G(\bm{\epsilon})}\log\left(
\delta_{m,L}^{-1}
\right)}.
\end{align}
By setting $\delta_{m,L} \gtrsim \frac{R_F^2(q) \log(m)}{mLB(\bm{\epsilon})}$, we have
\begin{align*}
\sqrt{\frac{2^{u+3}R_F^2(q) \delta_{m,L}}{ mLB(\bm{\epsilon})}\log\left(
\delta_{m,L}^{-1}
\right)} \lesssim 2^{u-3}\delta_{m,L}.
\end{align*}
The desired upper bound follows immediately. \qed \\

 \begin{lemma}
 \label{Lemma:REB}
 For any $\bm{\theta}^\star$, the following inequality holds for any $\widehat{\bm{\theta}}$ and $K \in [m-1]$:
 \begin{align*}
\mathbb{E}\big(H_K(\widehat{\bm{\theta}},\bm{\theta}^\star)\big)
 \leq 
 \frac{1}{K}
 \left[
\sum_{i:\sigma(\theta_i^\star) \leq K}
\mathbb{P}\left(
\widehat{\theta}_i \leq \frac{\theta_{(K)}^\star+\theta_{(K+1)}^\star}{2}
\right) +
\sum_{i:\sigma(\theta_i^\star)> K}
\mathbb{P}\left(
\widehat{\theta}_i \geq 
\frac{\theta_{(K)}^\star+\theta_{(K+1)}^\star}{2}
\right)
\right].
 \end{align*}
 \end{lemma}

\noindent
\textbf{Proof of Lemma \ref{Lemma:REB}.} Lemma \ref{Lemma:REB} offers a way to quantify the error associated with utilizing $\widehat{\bm{\theta}}$ for identifying the top-$K$ item set. The proof is a direct application of Lemma 3.1 of \citet{AndersonZhang}. Using Lemma 3.1 of \citet{AndersonZhang}, we have
\begin{align*}
H_K(\widehat{\bm{\theta}},\bm{\theta}^\star) 
\leq 
\frac{1}{2K}
\min_{t \in \mathbb{R}}
\left[
\sum_{i:\sigma(\theta_i^\star) \leq K}
I\left(
\widehat{\theta}_i \leq t
\right) +
\sum_{i:\sigma(\theta_i^\star)> K}
I\left(
\widehat{\theta}_i \geq t
\right)
\right].
\end{align*}
Taking the expectation of both sides, we get
\begin{align*}
\mathbb{E}\left( 
H_K(\widehat{\bm{\theta}},\bm{\theta}^\star) 
\right)
\leq 
\frac{1}{2K}
\min_{t \in \mathbb{R}}
\left[
\sum_{i:\sigma(\theta_i^\star) \leq K}
\mathbb{P}\left(
\widehat{\theta}_i \leq t
\right) +
\sum_{i:\sigma(\theta_i^\star)> K}
\mathbb{P}\left(
\widehat{\theta}_i \geq t
\right)
\right].
\end{align*}
The desired result immediately follows by taking $t = \frac{\theta_{(K)}^\star+\theta_{(K+1)}^\star}{2}$. \qed \\

\begin{lemma}
\label{Lemma:BTL_Infnorm}
Let $\widehat{\bm{\theta}} = \arg\min_{\bm{\theta}} \mathcal{L}_{\lambda}(\bm{\theta})$ be the optimal minimizer under the BTL model. Conditional on the event $\mathcal{E}(q)=\{\Vert \widehat{\bm{\theta}}-\bm{\theta}^\star\Vert_{\infty} \leq q\}$, it holds that 
\begin{align*}
\Vert \widehat{\bm{\theta}}-\bm{\theta}^\star\Vert_{\infty}
\leq  \frac{m^{-\frac{1}{2}}}{C_0(q)}
     \Vert \widehat{\bm{\theta}} - 
    \bm{\theta}^\star\Vert_2 + \frac{16}{C_0(q)}\sqrt{\frac{\log(m)}{mLpB(\bm{\epsilon})}}+ \frac{4\lambda \kappa}{mpC_0(q)},
\end{align*}
with probability at least $1-\frac{C}{(m)^3}$ for some positive constant $C$.
\end{lemma}

\noindent
\textbf{Proof of Lemma \ref{Lemma:BTL_Infnorm}.} By the fact that $\widehat{\bm\theta}$ is optimal minimizer of $\mathcal{L}_{\lambda}(\bm\theta)$, we have
\begin{align}
\label{Main_2}
 \nabla\mathcal{L}_{\lambda}(\widehat{\bm\theta})+ \nabla\mathcal{L}_{\lambda}(\bm\theta^\star)- \nabla\mathcal{L}_{\lambda}(\bm\theta^\star)
= \nabla^2\mathcal{L}_{\lambda}(\bm\theta_0)(\widehat{\bm\theta}-\bm\theta^\star)- \nabla\mathcal{L}_{\lambda}(\bm\theta^\star)=0,
\end{align}
for some $\bm\theta_0=(\theta_{0,1},\ldots,\theta_{0,m})$, where $\nabla^2\mathcal{L}_{\lambda}(\bm\theta)=\left(\frac{\partial^2 \mathcal{L}_{\lambda}(\bm\theta)}{\partial \theta_i \partial \theta
_j}\right)_{i,j \in [m]}$ and $\nabla\mathcal{L}_{\lambda}(\bm\theta)=\left(\frac{\partial \mathcal{L}_{\lambda}(\bm\theta)}{\partial \theta_i}\right)_{i \in [m]}$ are the Hessian matrix and the gradient of $\mathcal{L}_{\lambda}(\bm\theta)$, respectively.

Let $\nabla^2\mathcal{L}_{\lambda}(\bm\theta_0)=\bm{D}(\bm\theta_0)+\bm{A}(\bm\theta_0)$ with $\bm{D}(\bm\theta_0)$ and $\bm{A}(\bm\theta_0)$ being the diagonal and off-diagonal parts of $\nabla^2\mathcal{L}_{\lambda}(\bm\theta_0)$, respectively. For each $i\neq j$, $A_{ij}(\bm\theta_0)$ can be written as
\begin{align*}
    A_{ij}(\bm\theta_0) = -\sum_{l=1}^L w_l
    a_{ij}^{(l)}\left(
\widetilde{z}_{ij}^{(l)}g'(\gamma_{ij}^0)+(1-\widetilde{z}_{ij}^{(l)})
g'(-\gamma_{ij}^0)
\right)=-\sum_{l=1}^L w_l
    a_{ij}^{(l)} \frac{e^{\gamma_{ij}^0}}{(1+e^{\gamma_{ij}^0})^2},
\end{align*}
where $\gamma_{ij}^0 = \theta_{0,i}-\theta_{0,j}$. Furthermore, we can write $D_{ii}(\bm\theta_0)$ as
\begin{align*}
    D_{ii}(\bm\theta_0) = 
     \sum_{l=1}^L w_l\left\{
\sum_{j \in [m] \setminus \{i\}}a_{ij}^{(l)}\frac{e^{\gamma_{ij}^0}}{(1+e^{\gamma_{ij}^0})^2}
\right\}+
    2\lambda 
    =  -\sum_{j \in [m] \setminus \{i\}}A_{ij}(\bm\theta_0)+2\lambda,
\end{align*}
where the first equality holds by the facts that $\widetilde{z}_{ij}^{(l)}=1-\widetilde{z}_{ji}^{(l)}$ and $g'(\gamma_{ij})=g'(-\gamma_{ji})$ for any $i \neq j$. Given that $F(x)=1/(1+e^{-x})$, we have $g'(x)<0$. Using this property, we can show that $\mathcal{L}_{\lambda}(\bm\theta)$ is uniformly convex on $\mathcal{C}(q)=\{\bm{\theta}:\Vert \bm{\theta} - \bm{\theta}^\star\Vert_{\infty} \leq q\}$, or equivalently, $\nabla^2\mathcal{L}_{\lambda}(\bm\theta)$ is a positive definite matrix for any $\bm{\theta} \in  \mathcal{C}(q)$. Rearranging the inequality in (\ref{Main_2}), we have
\begin{align*}
    D_{ii}(\bm\theta_0)
    |\widehat{\theta}_i - \theta_i^\star| \leq   \sum_{j \in [m]\setminus \{i\}} |A_{ij}(\bm{\theta}_0)| \cdot|\widehat{\theta}_j - \theta_j^
    \star|+ |\xi_i|
    \leq \frac{1}{4}
    \sum_{j \in [m]\setminus \{i\}} \left(\sum_{l=1}^L w_l
    a_{ij}^{(l)} \right) \cdot|\widehat{\theta}_j - \theta_j^
    \star|+ |\xi_i|.
\end{align*}
Denote that $C_0(q)=\frac{e^{\kappa+2q}}{(1+e^{\kappa+2q})^2}$. The following inequality holds for any $\bm{\theta} \in \mathcal{C}(q)$.
\begin{align*}
C_0(q)\underbrace{
\left(
\sum_{l=1}^L \sum_{j \in [m]\setminus \{i\}} w_l a_{ij}^{(l)}\right)}_{K_L} |\widehat{\theta}_i - \theta_i^\star|
    \leq D_{ii}(\bm\theta_0) |\widehat{\theta}_i - \theta_i^\star| \leq \frac{1}{4}\underbrace{
    \sum_{j \in [m]\setminus \{i\}} \left(\sum_{l=1}^L w_l
    a_{ij}^{(l)} \right) \cdot|\widehat{\theta}_j - \theta_j^
    \star|}_{K_U}+ |\xi_i|.
\end{align*}
In what follows, we provide probabilistic lower bound and upper bound for $K_L$ and $K_U$, respectively.

\noindent
\textbf{Step 1 Bounding $K_L$.}
For any $v>0$,
\begin{align*}
    &\mathbb{P}
    \left(
K_L>v
    \right) = 
        \mathbb{P}
    \left(
K_L-\mathbb{E}(K_L)>v - (m-1)p
    \right) \leq 
            \mathbb{P}
    \left(
K_L-\mathbb{E}(K_L)>v - mp
    \right).
\end{align*}
To apply the Bernstein's inequality, we first verify the following conditions.
\begin{align*}
  w_l a_{ij}^{(l)}   
\leq \frac{1}{G(\bm{\epsilon})}
 \mbox{ and }
\text{Var}
\left(
\sum_{l=1}^L \sum_{j \in [m]\setminus \{i\}} w_l a_{ij}^{(l)}
\right) 
 \leq \frac{mp(1-p)}{G(\bm{\epsilon})} \leq \frac{mp}{G(\bm{\epsilon})}.
\end{align*}
Let $v = mp/4$. It then follows that
\begin{align*}
\mathbb{P}
    \left(
K_L>v
    \right) \leq & \mathbb{P}
    \left(
K_L-\mathbb{E}(K_L)>-mp/2
    \right) = 1-
    \mathbb{P}
    \left(
K_L-\mathbb{E}(K_L)<-mp/2
    \right) \\
    \leq &
        1-\exp\left(
-mLpB(\bm{\epsilon})/4
    \right).
\end{align*}
Therefore, we have $K_L > mp/2$ with probability at least $1- \exp\left(-mLpB(\bm{\epsilon})/4\right)$.

\noindent
\textbf{Step 2 Bounding $K_U$.} By the Cauchy–Schwarz inequality, we have
\begin{align*}
   & \sum_{j \in [m]\setminus \{i\}} \left(\sum_{l=1}^L w_l
    a_{ij}^{(l)} \right) \cdot|\widehat{\theta}_j - \theta_j^
    \star|
    \leq 
    \sqrt{\sum_{j \in [m]\setminus \{i\}} \left(\sum_{l=1}^L w_l
    a_{ij}^{(l)} \right)^2} \cdot 
    \sqrt{\sum_{j=1}^m
    (\widehat{\theta}_j - \theta_j^\star)^2 }
    \\
    = &
    \sqrt{\sum_{j \in [m]\setminus \{i\}} \left(\sum_{l=1}^L w_l
    a_{ij}^{(l)} \right)^2}\cdot 
\Vert \widehat{\bm{\theta}} - \bm{\theta}^\star\Vert_2
\triangleq \sqrt{\sum_{j \in [m]\setminus\{i\}}q_j^2} \cdot \Vert \widehat{\bm{\theta}} - \bm{\theta}^\star\Vert_2\\
=&
\sqrt{\frac{1}{m}\sum_{j \in [m]\setminus\{i\}}q_j^2} \cdot \sqrt{m}\Vert \widehat{\bm{\theta}} - \bm{\theta}^\star\Vert_2,
\end{align*}
where $q_j = \sum_{l=1}^L w_l a_{ij}^{(l)}$. Here $q_j$'s are independent and identically distributed. Additionally, $\mathbb{E}(q_j) = p$ and $\text{Var}(q_j) = \sum_{l=1}^L w_l^2 p(1-p)$. By the law of large number, $\frac{1}{m}\sum_{j \in [m]\setminus\{i\}}q_j^2$ converges to $\mathbb{E}(q_j^2)$ with large probability, which is given as $\mathbb{E}(q_j^2) = 
    \sum_{l=1}^L w_l^2 p(1-p)+p^2 \lesssim p^2$. To sum up, we have $C_0(q)mp|\widehat{\theta}_i - \theta_i^\star|
     \leq p
    \sqrt{m}\Vert \widehat{\bm{\theta}} - \bm{\theta}^\star\Vert_2+ 4|\xi_i|$ with probability at least $1- \exp\left(-mLpB(\bm{\epsilon})/4\right)$.

\noindent
\textbf{Step 3. Bounding $|\xi_i|$.} Here $\xi_i$ is the $i$-the element of $\nabla\mathcal{L}_{\lambda}(\bm\theta^\star)$ and given as
\begin{align*}
\xi_{i} \triangleq \frac{\partial \mathcal{L}_{\lambda}(\bm\theta)}{\partial \theta_i}\Big|_{\bm{\theta}=\bm{\theta}^\star}
=&
  \sum_{l=1}^L \left\{-
\sum_{j:i<j}w_l a_{ij}^{(l)}\left(\widetilde{z}_{ij}^{(l)}\frac{f(\gamma_{ij}^\star)}{F(\gamma_{ij}^\star)}-(1-\widetilde{z}_{ij}^{(l)})
\frac{f(\gamma_{ij}^\star)}{1-F(\gamma_{ij}^\star)}
\right)+ \right.\\
&\left.
\sum_{j:i>j}w_l a_{ji}^{(l)}\left(\widetilde{z}_{ji}^{(l)}\frac{f(\gamma_{ji}^\star)}{F(\gamma_{ji}^\star)}-(1-\widetilde{z}_{ji}^{(l)})
\frac{f(\gamma_{ji}^\star)}{1-F(\gamma_{ji}^\star)}
\right)
\right\}+2\lambda \theta_i^\star \\
=&
  \sum_{l=1}^L \left\{-\sum_{j:i<j}w_l a_{ij}^{(l)}\left(\widetilde{z}_{ij}^{(l)}\frac{f(\gamma_{ij}^\star)}{F(\gamma_{ij}^\star)}-(1-\widetilde{z}_{ij}^{(l)})
\frac{f(\gamma_{ij}^\star)}{1-F(\gamma_{ij}^\star)}
\right)+\right.\\
&\left.
\sum_{j:i>j}w_l a_{ji}^{(l)}\left(\widetilde{z}_{ji}^{(l)}\frac{f(\gamma_{ij}^\star)}{1-F(\gamma_{ij}^\star)}-(1-\widetilde{z}_{ji}^{(l)})
\frac{f(\gamma_{ij}^\star)}{F(\gamma_{ij}^\star)}
\right)
\right\}+2\lambda \theta_i^\star,
\end{align*}
where the last equality follows from the fact that $f(x)=f(-x)$ and $F(x)=1-F(-x)$ for any $x\in \mathbb{R}$. Next, we turn to provide probabilistic upper bounds for $\sum_{j \in [m]\setminus \{i\}} |A_{ij}(\bm{\theta}_0)| \cdot|\widehat{\theta}_j - \theta_j^
    \star|$ and $\xi_i$, respectively.
 We notice the following facts for $F(x)=1/(1+e^{-x})$:
\begin{align*}
&    \mathbb{E}
    \left(
\widetilde{z}_{ij}^{(l)}\frac{f(\gamma_{ij}^\star)}{F(\gamma_{ij}^\star)}-(1-\widetilde{z}_{ij}^{(l)})
\frac{f(\gamma_{ij}^\star)}{1-F(\gamma_{ij}^\star)}
    \right)=
    F(\gamma_{ij}^\star)\frac{f(\gamma_{ij}^\star)}{F(\gamma_{ij}^\star)}-(1-F(\gamma_{ij}^\star))
\frac{f(\gamma_{ij}^\star)}{1-F(\gamma_{ij}^\star)}=0, \\
&\text{Var}
\left[
w_l
a_{ij}^{(l)}
\left(
\widetilde{z}_{ij}^{(l)}\frac{f(\gamma_{ij}^\star)}{F(\gamma_{ij}^\star)}-(1-\widetilde{z}_{ij}^{(l)})
\frac{f(\gamma_{ij}^\star)}{1-F(\gamma_{ij}^\star)}
\right)\right] \leq \frac{p\left(
\frac{e^{\epsilon_l}-1}{e^{\epsilon_l}+1}\right)^2}{4G^2(\bm{\epsilon})} , \\
& \max_{l \in [L]}w_l
\left|
    \widetilde{z}_{ij}^{(l)}\frac{f(\gamma_{ij}^\star)}{F(\gamma_{ij}^\star)}-(1-z_{ij}^{(l)})
\frac{f(\gamma_{ij}^\star)}{1-F(\gamma_{ij}^\star)}\right|
\leq  \frac{1}{G(\bm{\epsilon})}.
\end{align*}
Then applying the Bernstein's inequality yields that
\begin{align*}
  &  \mathbb{P}
    \left(
\left|\frac{1}{mp}
    \sum_{l \in [L]} \sum_{j \in [m] \setminus \{i\}  }w_l
a_{ij}^{(l)}\left(\widetilde{z}_{ij}^{(l)}\frac{f(\gamma_{ij}^\star)}{F(\gamma_{ij}^\star)}-(1-\widetilde{z}_{ij}^{(l)})
\frac{f(\gamma_{ij}^\star)}{1-F(\gamma_{ij}^\star)}
\right)\right|>t
    \right)  \\
    \leq &\exp\left(
-t^2 mLpB(\bm{\epsilon})/4
    \right).
\end{align*}
Choosing $t = 4\sqrt{\frac{\log(m)}{mLpB(\bm{\epsilon})}}$ yields that
\begin{align*}
    |\widehat{\theta}_i-\theta_i^\star|
\leq &
    \frac{m^{-\frac{1}{2}}}{C_0(q)} \Vert \widehat{\bm{\theta}} - 
    \bm{\theta}^\star\Vert_2 + \frac{16}{C_0(q)}\sqrt{\frac{\log(m)}{mLpB(\bm{\epsilon})}}+ \frac{4\lambda \kappa}{mp},
\end{align*}
with probability at least $1-O(m^{-4})$. Applying the above argument for $i \in [m]$, we obtain the following union bound
\begin{align*}
  \Vert \widehat{\bm{\theta}} - \bm{\theta}^\star\Vert_{\infty} =    \max_{i \in [m]}   |\widehat{\theta}_i-\theta_i^\star|
\leq & \frac{m^{-\frac{1}{2}}}{C_0(q)}
     \Vert \widehat{\bm{\theta}} - 
    \bm{\theta}^\star\Vert_2 + \frac{16}{C_0(q)}\sqrt{\frac{\log(m)}{mLpB(\bm{\epsilon})}}+ \frac{4\lambda \kappa}{mpC_0(q)},
\end{align*}
with probability at least $1-O(m^{-3})$. This completes the proof. \qed \\

\begin{lemma}
\label{Temp_Lemma}
Under the assumptions of Theorem \ref{Thm:Estimation_General}, conditional on the event $\mathcal{E}(q)$, it holds that
\begin{align*}
\Vert \widehat{\bm{\theta}}-\bm{\theta}^\star\Vert_{\infty}
\leq &\frac{C_{g',U}(q)}{C_{g',L}(q)}
    m^{-\frac{1}{2}} \Vert \widehat{\bm{\theta}} - 
    \bm{\theta}^\star\Vert_2 + 8\frac{C_{g',U}(q)C_{g,U}(q)}{C_{g',L}(q)}\sqrt{\frac{\log(m)}{mLpB(\bm{\epsilon})}}+ \frac{2\lambda \kappa}{mpC_{g',L}(q)},
\end{align*}
with probability at least $1-O(m^{-3})$.
\end{lemma}

\noindent
\textbf{Proof of Lemma \ref{Temp_Lemma}.} By the fact that $\widehat{\bm\theta}$ is the optimal minimizer of $\nabla\mathcal{L}_{\lambda}(\widehat{\bm\theta})$, we have
\begin{align}
\label{Main}
 \nabla\mathcal{L}_{\lambda}(\widehat{\bm\theta})+ \nabla\mathcal{L}_{\lambda}(\bm\theta^\star)- \nabla\mathcal{L}_{\lambda}(\bm\theta^\star)
= \nabla^2\mathcal{L}_{\lambda}(\bm\theta_0)(\widehat{\bm\theta}-\bm\theta^\star)- \nabla\mathcal{L}_{\lambda}(\bm\theta^\star)=0,
\end{align}
for some $\bm\theta_0=(\theta_{0,1},\ldots,\theta_{0,m})$, where $\nabla^2\mathcal{L}_{\lambda}(\bm\theta)=\left(\frac{\partial^2 \mathcal{L}_{\lambda}(\bm\theta)}{\partial \theta_i \partial \theta
_j}\right)_{i,j \in [m]}$ and $\nabla\mathcal{L}_{\lambda}(\bm\theta)=\left(\frac{\partial \mathcal{L}_{\lambda}(\bm\theta)}{\partial \theta_i}\right)_{i \in [m]}$ are the Hessian matrix and the gradient of $\mathcal{L}_{\lambda}(\bm\theta)$, respectively. The equality in (\ref{Main}) implies that $\sum_{j=1}^m 
    \left(
\nabla^2\mathcal{L}_{\lambda}(\bm\theta_0)
    \right)_{(i,j)}(\widehat{\theta}_j-\theta_j^\star) = \left(
\nabla\mathcal{L}_{\lambda}(\bm\theta^\star)
    \right)_{i}$ for $i \in [m]$. Rearranging this equality yields that
\begin{align*}
    \left(
\nabla^2\mathcal{L}_{\lambda}(\bm\theta_0)
    \right)_{(i,i)}(\widehat{\theta}_i-\theta_i^\star) = \left(
\nabla\mathcal{L}_{\lambda}(\bm\theta^\star)
    \right)_{i} -  \sum_{j \in [m] \setminus \{i\}} 
    \left(
\nabla^2\mathcal{L}_{\lambda}(\bm\theta_0)
    \right)_{(i,j)}(\widehat{\theta}_j-\theta_j^\star).
\end{align*}
Take the absolute value of both sides, one has
\begin{align}
    \label{Second}
   \left| \left(
\nabla^2\mathcal{L}_{\lambda}(\bm\theta_0)
    \right)_{(i,i)} \right| \cdot \left| \widehat{\theta}_i-\theta_i^\star \right|  \leq \left|\left(
\nabla\mathcal{L}_{\lambda}(\bm\theta^\star)
    \right)_{i}\right|   + \sum_{j \in [m] \setminus \{i\}} 
    \left|  \left(
\nabla^2\mathcal{L}_{\lambda}(\bm\theta_0)
    \right)_{(i,j)} \right|
    \cdot \left| \widehat{\theta}_j-\theta_j^\star\right|.
\end{align}

Let $\nabla^2\mathcal{L}_{\lambda}(\bm\theta_0)=\bm{D}(\bm\theta_0)+\bm{A}(\bm\theta_0)$ with $\bm{D}(\bm\theta_0)$ and $\bm{A}(\bm\theta_0)$ being the diagonal and off-diagonal parts of $\nabla^2\mathcal{L}_{\lambda}(\bm\theta_0)$, respectively. For each $i\neq j$, $A_{ij}(\bm\theta_0)$ can be written as
\begin{align*}
 \left(
\nabla^2\mathcal{L}_{\lambda}(\bm\theta_0)
    \right)_{(i,j)}=   A_{ij}(\bm\theta_0) = \sum_{l=1}^L w_l
    a_{ij}^{(l)}\left(
\widetilde{z}_{ij}^{(l)}g'(\gamma_{ij}^0)+(1-\widetilde{z}_{ij}^{(l)})
g'(-\gamma_{ij}^0)
\right),
\end{align*}
where $\gamma_{ij}^0 = \theta_{0,i}-\theta_{0,j}$. Furthermore, we can write $D_{ii}(\bm\theta_0)$ as
\begin{align*}
    D_{ii}(\bm\theta_0) = &-
     \sum_{l=1}^L w_l\left\{
\sum_{j \in [m] \setminus \{i\}}a_{ij}^{(l)}\left(\widetilde{z}_{ij}^{(l)}g'(\gamma_{ij}^0)+(1-\widetilde{z}_{ij}^{(l)})
g'(-\gamma_{ij}^0)
\right)
\right\}+
    2\lambda 
    \\
    =&  -\sum_{j \in [m] \setminus \{i\}}A_{ij}(\bm\theta_0)+2\lambda,
\end{align*}
where the first equality holds by the facts that $\widetilde{z}_{ij}^{(l)}=1-\widetilde{z}_{ji}^{(l)}$ and $g'(\gamma_{ij})=g'(-\gamma_{ji})$ for any $i \neq j$. Given that $F(x)$ is any log-concave function. Using this property, we can show that $\mathcal{L}_{\lambda}(\bm\theta)$ is convex with high probability, or equivalently, $\nabla^2\mathcal{L}_{\lambda}(\bm\theta)$ is a positive definite matrix with high probability (Lemma \ref{Lemma:Hessian}).

In what follows, we turn to show that $D_{ii}(\bm\theta_0)$ is lower bounded by a positive constant relating to $m,L$, and $p$ with high probability. For any $v>0$,
\begin{align*}
    &\mathbb{P}
    \left(
D_{ii}(\bm\theta_0)>v
    \right) = 
        \mathbb{P}
    \left\{
D_{ii}(\bm\theta_0)-\mathbb{E}\left[D_{ii}(\bm\theta_0)\right]>v-\mathbb{E}\left[D_{ii}(\bm\theta_0)\right]
    \right\} \\
    =  &
    \mathbb{P}
    \left\{
D_{ii}(\bm\theta_0)-\mathbb{E}\left[D_{ii}(\bm\theta_0)\right]>v+\mathbb{E}\left[\sum_{j \neq i}A_{ij}(\bm\theta_0)\right]-2\lambda
    \right\} \\
    \geq &
        \mathbb{P}
    \left\{
\sum_{j \neq i}A_{ij}(\bm\theta_0)-\mathbb{E}\left[\sum_{j \neq i}A_{ij}(\bm\theta_0)\right]<-v+C_{g',L}(q) mp+2\lambda
    \right\}.
\end{align*}
To apply the Bernstein's inequality, we first show the following conditions.
\begin{align*}
& \left| w_l\left(
\widetilde{z}_{ij}^{(l)}g'(\gamma_{ij}^0)+(1-\widetilde{z}_{ij}^{(l)})
g'(-\gamma_{ij}^0)
\right)\right| \leq  \frac{e^{\epsilon_l}-1}{e^{\epsilon_l}+1}\frac{C_{g',U}(q)}{G(\bm{\epsilon})}
\leq \frac{C_{g',U}(q)}{G(\bm{\epsilon})}
, \\
&\text{Var}
\left(
A_{ij}(\bm\theta_0)
\right) \leq \frac{C_{g',U}^2(q) p}{4\sum_{l=1}^L
\left(
\frac{e^{\epsilon_l}-1}{e^{\epsilon_l}+1}
\right)^2}
 =
\frac{C_{g',U}^2(q)p }{4\sum_{l=1}^L
\left(
\frac{e^{\epsilon_l}-1}{e^{\epsilon_l}+1}
\right)^2} = 
\frac{C_{g',U}^2(q)p }{4G(\bm{\epsilon})}
\end{align*}
Let $v = C_{g',L}(q) mp/2+2\lambda$. It then follows that
\begin{align*}
     &   \mathbb{P}
    \left\{
\sum_{j \neq i}A_{ij}(\bm\theta_0)-\mathbb{E}\left[\sum_{j \neq i}A_{ij}(\bm\theta_0)\right]<-v+C_{g',L} mp+2\lambda
    \right\} \\
    = & 
    \mathbb{P}
    \left\{
\sum_{j \neq i}A_{ij}(\bm\theta_0)-\mathbb{E}\left[\sum_{j \neq i}A_{ij}(\bm\theta_0)\right]<C_{g',L}(q) mp/2
    \right\}  \\
    \geq & 1-
    \exp\left(-\frac{3C_{g',L}^2(q) m^2 p^2 G(\bm{\epsilon})}{6mp
 C_{g',U}^2(q)+8 mp C_{g',U}^2(q)}\right)
=  1- 
\exp\left(-\frac{3C_{g',L}^2(q) mLpB(\bm{\epsilon})}{14C_{g',U}^2(q)} \right).
\end{align*}
Therefore, we have $D_{ii}(\bm\theta_0) >C_{g',L}(q) mp/2$ with probability at least $1- \exp\left(-\frac{3C_{g',L}^2 (q)mLpB(\bm{\epsilon})}{14C_{g',U}^2(q)} \right)$.

Therefore, conditional on the event $D_{ii}(\bm\theta_0)>C_{g',L}(q) mp/2$, we have
\begin{align}
\label{Main2}
    C_{g',L}(q) mp|\widehat{\theta}_i - \theta_i^\star|/2 \leq &|D_{ii}(\bm\theta_0)|
    \cdot |\widehat{\theta}_i - \theta_i^\star| \leq   \sum_{j \in [m]\setminus \{i\}} |A_{ij}(\bm{\theta}_0)| \cdot|\widehat{\theta}_j - \theta_j^
    \star|+ |\xi_i| \notag\\
    \leq &
 C_{g',U}(q) p  \sqrt{m} \Vert \widehat{\bm{\theta}} - \bm{\theta}^\star\Vert_{2}+|\xi_i|,
\end{align}
where the last inequality holds with probability at least $1-\exp(-CmLpB(\bm{\epsilon}))$ and $\xi_i$ is the $i$-the element of $\nabla\mathcal{L}_{\lambda}(\bm\theta^\star)$ and given as
\begin{align*}
\xi_{i} \triangleq \frac{\partial \mathcal{L}_{\lambda}(\bm\theta)}{\partial \theta_i}\Big|_{\bm{\theta}=\bm{\theta}^\star}
=&
  \sum_{l=1}^L \left\{-
\sum_{j:i<j}w_l a_{ij}^{(l)}\left(\widetilde{z}_{ij}^{(l)}\frac{f(\gamma_{ij}^\star)}{F(\gamma_{ij}^\star)}-(1-\widetilde{z}_{ij}^{(l)})
\frac{f(\gamma_{ij}^\star)}{1-F(\gamma_{ij}^\star)}
\right)+ \right.\\
&\left.
\sum_{j:i>j}w_l a_{ji}^{(l)}\left(\widetilde{z}_{ji}^{(l)}\frac{f(\gamma_{ji}^\star)}{F(\gamma_{ji}^\star)}-(1-\widetilde{z}_{ji}^{(l)})
\frac{f(\gamma_{ji}^\star)}{1-F(\gamma_{ji}^\star)}
\right)
\right\}+2\lambda \theta_i^\star \\
=&
  \sum_{l=1}^L \left\{-\sum_{j:i<j}w_l a_{ij}^{(l)}\left(\widetilde{z}_{ij}^{(l)}\frac{f(\gamma_{ij}^\star)}{F(\gamma_{ij}^\star)}-(1-\widetilde{z}_{ij}^{(l)})
\frac{f(\gamma_{ij}^\star)}{1-F(\gamma_{ij}^\star)}
\right)+\right.\\
&\left.
\sum_{j:i>j}w_l a_{ji}^{(l)}\left(\widetilde{z}_{ji}^{(l)}\frac{f(\gamma_{ij}^\star)}{1-F(\gamma_{ij}^\star)}-(1-\widetilde{z}_{ji}^{(l)})
\frac{f(\gamma_{ij}^\star)}{F(\gamma_{ij}^\star)}
\right)
\right\}+2\lambda \theta_i^\star,
\end{align*}
where the last equality follows from the fact that $f(x)=f(-x)$ and $F(x)=1-F(-x)$ for any $x\in \mathbb{R}$. Furthermore, since $m-1 \geq m/2$, re-arranging the inequality in (\ref{Main2}) gives
\begin{align*}
|\widehat{\theta}_i-\theta_i^\star|
\leq &
    \frac{C_{g',U}(q)}{\sqrt{m}C_{g',L}(q)} \Vert \widehat{\bm{\theta}} - 
    \bm{\theta}^\star\Vert_2 + \left|
    \frac{2C_{g',U}(q)}{mpC_{g',L}(q)}
    \sum_{l =1}^L\sum_{j \in [m]\setminus\{i\} }w_l
a_{ij}^{(l)}\left(\widetilde{z}_{ij}^{(l)}\frac{f(\gamma_{ij}^\star)}{F(\gamma_{ij}^\star)}-(1-\widetilde{z}_{ij}^{(l)})
\frac{f(\gamma_{ij}^\star)}{1-F(\gamma_{ij}^\star)}
\right)\right| \\
+ &
    \frac{2\lambda \kappa}{mpC_{g',L}(q)}.
\end{align*}
Here we notice that
\begin{align*}
&    \mathbb{E}
    \left(
\widetilde{z}_{ij}^{(l)}\frac{f(\gamma_{ij}^\star)}{F(\gamma_{ij}^\star)}-(1-\widetilde{z}_{ij}^{(l)})
\frac{f(\gamma_{ij}^\star)}{1-F(\gamma_{ij}^\star)}
    \right)=
    F(\gamma_{ij}^\star)\frac{f(\gamma_{ij}^\star)}{F(\gamma_{ij}^\star)}-(1-F(\gamma_{ij}^\star))
\frac{f(\gamma_{ij}^\star)}{1-F(\gamma_{ij}^\star)}=0, \\
&\text{Var}
\left[
w_l
a_{ij}^{(l)}
\left(
\widetilde{z}_{ij}^{(l)}\frac{f(\gamma_{ij}^\star)}{F(\gamma_{ij}^\star)}-(1-\widetilde{z}_{ij}^{(l)})
\frac{f(\gamma_{ij}^\star)}{1-F(\gamma_{ij}^\star)}
\right)\right] \leq \frac{pC_{g,U}^2(q)\left(
\frac{e^{\epsilon_l}-1}{e^{\epsilon_l}+1}\right)^2}{4G^2(\bm{\epsilon})} , \\
& \max_{l \in [L]}w_l
\left|
    \widetilde{z}_{ij}^{(l)}\frac{f(\gamma_{ij}^\star)}{F(\gamma_{ij}^\star)}-(1-z_{ij}^{(l)})
\frac{f(\gamma_{ij}^\star)}{1-F(\gamma_{ij}^\star)}\right|
\leq  \frac{C_{g,U}(q)}{G(\bm{\epsilon})}.
\end{align*}
Then applying the Bernstein's inequality yields that
\begin{align}
\label{BernTD}
  &  \mathbb{P}
    \left(
\left|\frac{1}{mpC_{g',L}(q)}
    \sum_{l \in [L]} \sum_{j \in [m] \setminus \{i\}  }w_l
a_{ij}^{(l)}\left(\widetilde{z}_{ij}^{(l)}\frac{f(\gamma_{ij}^\star)}{F(\gamma_{ij}^\star)}-(1-\widetilde{z}_{ij}^{(l)})
\frac{f(\gamma_{ij}^\star)}{1-F(\gamma_{ij}^\star)}
\right)\right|>t
    \right)  \notag \\
    \leq &\exp\left(
-\frac{3t^2 mLpB(\bm{\epsilon})C_{g',L}^2(q)}{8C^2_{g,U}(q)}
    \right).
\end{align}
Choosing $t = 4\frac{C_{g,U}(q)}{C_{g',L}(q)}\sqrt{\frac{\log(m)}{mLpB(\bm{\epsilon})}}$ yields that
\begin{align*}
    |\widehat{\theta}_i-\theta_i^\star|
\lesssim &\frac{C_{g',U}(q)}{C_{g',L}(q)}
    m^{-\frac{1}{2}} \Vert \widehat{\bm{\theta}} - 
    \bm{\theta}^\star\Vert_2 + 16\frac{C_{g',U}(q)C_{g,U}(q)}{C_{g',L}(q)}\sqrt{\frac{\log(m)}{mLpB(\bm{\epsilon})}}+ \frac{2\lambda \kappa}{mpC_{g',L}(q)},
\end{align*}
with probability at least $
1-O(m^{-3})$. This completes the proof. \qed \\

\begin{lemma}
    \label{Lemma:Boundedness}
    Suppose $\frac{\log(m)}{mLpB(\bm{\epsilon})}=o(1)$. For a general $F(\cdot)$ satisfying Assumptions \ref{Ass1} and \ref{Ass2}, we define a gradient descent sequence for the optimization task in (\ref{Equ:OPT1}) as
    \begin{align*}
    \bm{\theta}^{(t+1)} = \bm{\theta}^{(t)}
    - \alpha 
    \left(
\nabla\mathcal{L}_{\lambda}(\bm{\theta}^{(t)})
    \right) = 
     \bm{\theta}^{(t)}- 
     \alpha \left(
\nabla\mathcal{L}_{0}(\bm{\theta}^{(t)})+2\lambda \bm{\theta}^{(t)}
     \right),
\end{align*}
with $\frac{1}{\sqrt{\log m}} \lesssim q_0$ and $\alpha \leq \frac{4}{C_{g',U}(3q_0) mp}$. Moreover, $\lambda$ satisfies the conditions in Theorem \ref{Thm:ConsisT1}. Then there exist an initializer $\bm{\theta}^{(0)}$ such that $\Vert \widehat{\bm{\theta}} - \bm{\theta}^\star \Vert_{\infty} \leq 3q_0$, with probability at least $1-O(m^{-8})$.
    
\end{lemma}

\noindent
\textbf{Proof of Lemma \ref{Lemma:Boundedness}.} First, we introduce a leave-one-out gradient descent sequence as in \citet{chen2019spectral} and \citet{AndersonZhang}. Specifically, we define
\begin{align*}
\mathcal{L}^{(k)}(\bm{\theta})=&- \frac{1}{2}
\sum_{i \neq j,i,j \in [m] \setminus \{k\}}\sum_{l=1}^L a_{ij}^{(l)}w_l
\left\{
\widetilde{z}_{ij}^{(l)} \log
F(\theta_i-\theta_j)
+
(1-\widetilde{z}_{ij}^{(l)})
\log
\left(
1-F(\theta_i-\theta_j)\right)\right\} \\
-&
L
\sum_{i \in [m]\setminus k}
p \left\{
F(\theta_i^\star-\theta_k^\star) \log
F(\theta_i-\theta_k)
+
(1-F(\theta_i^\star-\theta_k^\star))
\log
\left(
1-F(\theta_i-\theta_k)\right)\right\},
\end{align*}
where $k \in [m]$. For $\mathcal{L}^{(k)}(\bm{\theta})$, we define the associated gradient descent sequence as
\begin{align*}
    \bm{\theta}^{(t+1,k)}  = 
     \bm{\theta}^{(t,k)}- 
     \alpha \left(
\nabla\mathcal{L}^{(k)}(\bm{\theta}^{(t,k)})+2\lambda \bm{\theta}^{(t,k)}
     \right).
\end{align*}
We set $\bm{\theta}^{(0,k)}=\bm{\theta}^{(0)}$ for each $k \in [m]$. This implies $\Vert \bm{\theta}^{(0,k)} - \bm{\theta}^{(0)}\Vert_{\infty}=0$ for $k \in [m]$. Then we consider the following three conditions for $1 \leq t\leq t^\star$.
\begin{align}
&    \max_{k \in [m]}
    \Vert \bm{\theta}^{(t,k)} - \bm{\theta}^{(t)}\Vert_2 \leq q_0, \label{GDeq1}\\
 &    \Vert \bm{\theta}^{(t)} - \bm{\theta}^\star\Vert_2 \leq 
     4\sqrt{\frac{m}{\log m}} \kappa,\label{GDeq2} \\
  &   \max_{k\in [m] }\left|\theta_k^{(t,k)}-\theta_k^\star\right| \leq q_0.\label{GDeq3}
\end{align}
Clearly, (\ref{GDeq1}), (\ref{GDeq2}), and (\ref{GDeq3}) hold true for $t = 0$ if the starting point $\bm{\theta}^{(0)}$ is chosen such that $\Vert \bm{\theta}^{(0)} - \bm{\theta}^\star\Vert_{\infty} \leq q_0$ and $\Vert \bm{\theta}^{(0)} - \bm{\theta}^\star\Vert_{2} \leq 4 \sqrt{\frac{m}{\log m}}\kappa$. For example, $\bm{\theta}^{(0)}=\bm{\theta}^\star$ is a particular example as specified in \citet{chen2019spectral,AndersonZhang}. In what follows, we show that these three inequalities hold true for any $1\leq t\leq t^\star$ based on a mathematical induction argument.

\vspace{3mm}
\noindent
\textbf{(Step 1. Bounding $\max\limits_{k \in [m]}\Vert \bm{\theta}^{(t+1)} - \bm{\theta}^{(t+1,k)} \Vert_2$.)} First, we suppose (\ref{GDeq1})-(\ref{GDeq3}) are true at $t$. Then we consider the case of $t+1$ for (\ref{GDeq1}).
\begin{align*}
    &\bm{\theta}^{(t+1)} - \bm{\theta}^{(t+1,k)}\\
    =&
    (1-2\alpha \lambda)(\bm{\theta}^{(t)} - \bm{\theta}^{(t,k)})+
    \alpha
    \left(
    \nabla\mathcal{L}^{(k)}(\bm{\theta}^{(t,k)})-
\nabla\mathcal{L}_{0}(\bm{\theta}^{(t)})
    \right) \\
    =&
    (1-2\alpha \lambda)(\bm{\theta}^{(t)} - \bm{\theta}^{(t,k)})
    -\alpha\left(\nabla\mathcal{L}_{0}(\bm{\theta}^{(t)})-
    \nabla\mathcal{L}_{0}(\bm{\theta}^{(t,k)})\right)
    +
    \alpha
    \left(
    \nabla\mathcal{L}^{(k)}(\bm{\theta}^{(t,k)})-
\nabla\mathcal{L}_{0}(\bm{\theta}^{(t,k)})
    \right) \\
    =&\left[
    \left(1-2\alpha \lambda\right)\bm{I}_m-\alpha\nabla^2\mathcal{L}_{0}(\bm\theta_0^{(t,k)})\right](\bm{\theta}^{(t)} - \bm{\theta}^{(t,k)})+
    \alpha
    \left(
    \nabla\mathcal{L}^{(k)}(\bm{\theta}^{(t,k)})-
\nabla\mathcal{L}_{0}(\bm{\theta}^{(t,k)})
    \right),
\end{align*}
where $\bm\theta_0^{(t,k)}$ is a convex combination of $\bm{\theta}^{(t)}$ and $\bm{\theta}^{(t,k)}$ and $\bm{I}_m$ is a $m\times m$ identity matrix. For a sufficiently small $\alpha>0$, we have
\begin{align*}
   & \left[
    \left(1-2\alpha \lambda\right)\bm{I}_m-\alpha\nabla^2\mathcal{L}_{0}(\bm\theta_0^{(t,k)})\right](\bm{\theta}^{(t)} - \bm{\theta}^{(t,k)}) \\
    \leq &\Vert \left(1-2\alpha \lambda\right)\bm{I}_m-\alpha\nabla^2\mathcal{L}_{0}(\bm\theta_0^{(t,k)}) 
    \Vert_{op} \cdot \Vert \bm{\theta}^{(t)} - \bm{\theta}^{(t,k)} \Vert_2,
\end{align*}
where $\Vert \cdot \Vert_{op}$ denotes the operator norm. By (\ref{GDeq1}) and (\ref{GDeq3}), we have 
\begin{align*}
  \Vert \bm{\theta}^{(t,k)} - \bm{\theta}^\star\Vert_{\infty} \leq & \Vert \bm{\theta}^{(t)} - \bm{\theta}^{(t,k)} \Vert_{\infty}+\Vert \bm{\theta}^{(t)} - \bm{\theta}^\star\Vert_{\infty} \leq q_0+\Vert \bm{\theta}^{(t)} - \bm{\theta}^\star\Vert_{\infty} \\
\leq &  q_0+ \max_{i \in [m]} |\theta_i^{(t)}-\theta_i^{(t,i)}+\theta_i^{(t,i)}-\theta_i^\star| 
\leq 2q_0 + \max_{i \in [m]} |\theta_i^{(t)}-\theta_i^{(t,i)}| 
\leq 3q_0.
\end{align*}
Therefore, we have $\bm{\theta}^{(t,k)},\bm{\theta}^{(t)} \in \mathcal{C}(3q_0)$ with $\mathcal{C}(\cdot)$ being defined in Lemma \ref{Lemma:Hessian}. This then implies that $\bm\theta_0^{(t,k)} \in \mathcal{C}(3q_0)$. By Lemmas \ref{Lemma:Hessian} and \ref{Lemma:LargestEigen}, we obtain
\begin{align*}
E_1 = \left\{
C_{g',L}(3q_0)mp/4 \leq 
\Lambda_{min,\perp}
\left(
\nabla^2\mathcal{L}_{0}(\bm\theta_0^{(t,k)})
\right) \leq 
\Lambda_{max}
\left(
\nabla^2\mathcal{L}_{0}(\bm\theta_0^{(t,k)})
\right)  \leq 2C_{g',U} (3q_0)mp \right\},
\end{align*}
with probability at least $1- m \exp\left(-\frac{3mpLB(\bm{\epsilon})}{32}\right)-m \exp\left(-\frac{3C_{g',L}^2(3q_0)mLpB(\bm{\epsilon})}{128  C_{g',U}^2(3q_0) }\right)$. Here we simplify the probability as
\begin{align*}
    m \exp\left(-\frac{3mpLB(\bm{\epsilon})}{32}\right)+m \exp\left(-\frac{3C_{g',L}^2(3q_0)mLpB(\bm{\epsilon})}{128  C_{g',U}^2(3q_0) }\right) \lesssim 
    m \exp\left(-C_1 mLpB(\bm{\epsilon})\right),
\end{align*}
where $C_1$ denotes some universal constants.

Choosing $\alpha \leq \frac{1}{4 C_{g',U} (3q_0)mp}$. In what follows, we turn to prove $\max_{k \in [m]}
    \Vert \bm{\theta}^{(t+1,k)} - \bm{\theta}^{(t+1)}\Vert_2 \leq q_0$ conditional the event $E_1$. First, by the triangle inequality, we have
    \begin{align}
    \label{FirstBound}
        &\Vert \bm{\theta}^{(t+1)} - \bm{\theta}^{(t+1,k)} \Vert_2 \notag \\ \leq & 
        \left[1-2\alpha\lambda-\alpha\Lambda_{min,\perp}\left(\nabla^2\mathcal{L}_{0}(\bm\theta_0^{(t,k)})\right)\right] \cdot \Vert \bm{\theta}^{(t)} - \bm{\theta}^{(t,k)}  \Vert_2 
        + \alpha \Vert \nabla\mathcal{L}^{(k)}(\bm{\theta}^{(t,k)})-
\nabla\mathcal{L}_{0}(\bm{\theta}^{(t,k)})\Vert_2 \notag \\
\leq & \left[1-\alpha\Lambda_{min,\perp}\left(\nabla^2\mathcal{L}_{0}(\bm\theta_0^{(t,k)})\right)\right]q_0+\alpha \Vert \nabla\mathcal{L}^{(k)}(\bm{\theta}^{(t,k)})-
\nabla\mathcal{L}_{0}(\bm{\theta}^{(t,k)})\Vert_2,
    \end{align}
where the first inequality holds with $\alpha\Lambda_{min,\perp}\left(\nabla^2\mathcal{L}_{0}(\bm\theta_0^{(t,k)})\right)<1$ when $E_1$ is true. Therefore, we have
\begin{align*}
    \Vert \bm{\theta}^{(t+1)} - \bm{\theta}^{(t+1,k)} \Vert_2 \leq  \left[1-\alpha C_{g',L}(3q_0) mp/4 \right]+\alpha \Vert \nabla\mathcal{L}^{(k)}(\bm{\theta}^{(t,k)})-
\nabla\mathcal{L}_{0}(\bm{\theta}^{(t,k)})\Vert_2.
\end{align*}  
 Next, we turn to bound the second term. We first note that
\begin{align*}
    &\Vert \nabla\mathcal{L}^{(k)}(\bm{\theta}^{(t,k)})-
\nabla\mathcal{L}_{0}(\bm{\theta}^{(t,k)})\Vert_2^2 \\
= &
\sum_{i \in [m]\setminus \{k\}} \left(\sum_{l=1}^L w_l
\left(a_{ik}^{(l)}
\widetilde{z}_{ik}^{(l)} -p 
F(\theta_i^\star-\theta_k^\star)\right)
g(\gamma_{ij}^{(t,k)})
-w_l
\left(
a_{ik}^{(l)}(1-\widetilde{z}_{ik}^{(l)})-
p+pF(\theta_i^\star-\theta_k^\star)
\right)
g(-\gamma_{ij}^{(t,k)})
\right)^2
\\
+&
 \left(\sum_{i \in [m]\setminus \{k\}}\sum_{l=1}^L w_l
\left(a_{ik}^{(l)}
\widetilde{z}_{ik}^{(l)} -p 
F(\theta_i^\star-\theta_k^\star)\right)
g(\gamma_{ij}^{(t,k)})
-w_l
\left(
a_{ik}^{(l)}(1-\widetilde{z}_{ik}^{(l)})-
p+pF(\theta_i^\star-\theta_k^\star)
\right)
g(-\gamma_{ij}^{(t,k)})
\right)^2 \\
\triangleq & I_1 + I_2,
\end{align*}
where $\gamma_{ij}^{(t,k)}=\theta_i^{(t,k)}-\theta_j^{(t,k)}$. It remains to bound $I_1$ and $I_2$ with high probability based on the Bernstein's inequality. We first show the following facts:
\begin{align*}
&w_l\left| 
\widetilde{z}_{ik}^{(l)} 
g(\gamma_{ij}^{(t,k)})
-
(1-\widetilde{z}_{ik}^{(l)})
g(-\gamma_{ij}^{(t,k)})\right| \leq \frac{e^{\epsilon_1}-1}{e^{\epsilon_l}+1}
\frac{C_{g,U}(3q_0)}{G(\bm{\epsilon})}
\leq \frac{C_{g,U}(3q_0)}{G(\bm{\epsilon})}
, \\
&\text{Var}\left(w_l \left(
a_{ik}^{(l)}
\widetilde{z}_{ik}^{(l)} -p 
F(\theta_i^\star-\theta_k^\star)\right)
\right) \leq \frac{p}{4G^2(\bm{\epsilon})} \left(\frac{e^{\epsilon_l}-1}{e^{\epsilon_l}+1}\right)^2.
\end{align*}
Applying the Bernstein's inequality to $I_2$, we have
\begin{align}
\label{I_2_Bound}
    \mathbb{P}(I_2 \geq t^2)
    \leq & 8 
    \exp
    \left(
-\frac{3G(\bm{\epsilon})t^2}{6m pC_{g,U}^2(3q_0)+ 2tC_{g,U}(3q_0)}
    \right) \leq 8\exp
    \left(
-\frac{3G(\bm{\epsilon})t^2}{8m pC_{g,U}^2(3q_0)}
    \right),
\end{align}
where the last inequality follows from the condition that $t \leq mpC_{g,U}(3q_0)$.

Similarly, for $I_1$, one has
\begin{align}
\label{I_1_bound}
  & \mathbb{P}(I_1 \geq t^2) \leq  4
   \mathbb{P}\left(\sum_{i \in [m]\setminus \{k\}} \left(\sum_{l=1}^L w_l
\left(a_{ik}^{(l)}
\widetilde{z}_{ik}^{(l)} -p 
F(\theta_i^\star-\theta_k^\star)\right)
g(\gamma_{ij}^{(t,k)})
\right)^2
>t^2
\right) \notag \\
\leq &4
\sum_{i \in [m]\setminus \{k\}}
\mathbb{P}\left( \left|\sum_{l=1}^L  w_l
\left(a_{ik}^{(l)}
\widetilde{z}_{ik}^{(l)} -p 
F(\theta_i^\star-\theta_k^\star)\right)
g(\gamma_{ij}^{(t,k)})
\right|
>\frac{t}{\sqrt{m}}
\right) \notag \\
\leq & 8m
\exp\left(
-\frac{3t^2 G(\bm{\epsilon})}{6mpC_{g,U}^2(3q_0) + 2\sqrt{m}t C_{g,U}(3q_0)}
\right) 
\leq 8m
\exp\left(
-\frac{3t^2 G(\bm{\epsilon})}{8mpC_{g,U}^2(3q_0)}
\right) ,
\end{align}
for $0<t \leq \sqrt{m}p C_{g,U}(3q_0)$. Therefore, for (\ref{I_2_Bound}) and (\ref{I_1_bound}), we choose $t =4\sqrt{\frac{2C_{g,U}^2(3q_0)mp\log(m)}{G(\bm{\epsilon})} }$ and it follows that
\begin{align*}
    E_2= \left\{
    \Vert \nabla\mathcal{L}^{(k)}(\bm{\theta}^{(t,k)})-
\nabla\mathcal{L}_{0}(\bm{\theta}^{(t,k)})\Vert_2 \leq 
    \sqrt{I_1} + \sqrt{I_2} \leq 16\sqrt{\frac{C_{g,U}^2(3q_0)mp\log(m)}{G(\bm{\epsilon})}}\right\} ,
\end{align*}
with probability at least $1-16/m^{11}$. Therefore, conditional the event $E_2$, (\ref{FirstBound}) can be further bounded as
\begin{align}
\label{Step_1_K}
    \Vert \bm{\theta}^{(t+1)} - \bm{\theta}^{(t+1,k)} \Vert_2 \leq
    q_0-\alpha q_0 C_{g',L}(3q_0) mp/4 +
    16 \alpha C_{g,U}(3q_0) \sqrt{\frac{mp\log(m)}{G(\bm{\epsilon})} },
\end{align}
with probability at least $1- m \exp\left(-C_1 mLpB(\bm{\epsilon})\right)-\frac{16}{m^{11}}$. Suppose $\frac{32C_{g,U}(3q_0)}{C_{g',L}^2(3q_0)} \sqrt{\frac{ \log(m)}{mpLB(\bm{\epsilon})}}
   \leq q_0$. Applying a union bound for cases $k\in [m]$ yields that
\begin{align*}
\mathbb{P}\left\{
\max_{k \in [m]}
\Vert \bm{\theta}^{(t+1)} - \bm{\theta}^{(t+1,k)} \Vert_2 \leq q_0\right\} \geq 
1- m^2 \exp\left(-C_1 mLpB(\bm{\epsilon})\right)-\frac{16}{m^{10}}.
\end{align*}
This completes the proof of (\ref{GDeq1}) for the case of $t+1$.

\vspace{3mm}
\noindent
\textbf{(Step 2. Bounding $\Vert \bm{\theta}^{(t+1)}-\bm{\theta}^\star \Vert_2$.)} Applying a similar argument of gradient descent update as above, we have
\begin{align*}
    \bm{\theta}^{(t+1)}-\bm{\theta}^\star = &
    \bm{\theta}^{(t)}-\bm{\theta}^\star-
    \alpha
    \left(
\nabla\mathcal{L}_{0}(\bm{\theta}^{(t)})+2\lambda \bm{\theta}^{(t)}
    \right) \\
    =&
    (1-2\alpha \lambda)\left(\bm{\theta}^{(t)}-\bm{\theta}^\star\right)-
    \alpha\left(
\nabla\mathcal{L}_{0}(\bm{\theta}^{(t)})-
\nabla\mathcal{L}_{0}(\bm{\theta}^\star) 
    \right)-2\alpha \lambda \bm{\theta}^\star -
    \alpha \nabla\mathcal{L}_{0}(\bm{\theta}^\star) \\
    =& 
    \left(
    (1-2\alpha\lambda)\bm{I}_m -\alpha
    \nabla^2\mathcal{L}_{0}(\bm{\xi}^{(t)})\right)\left(\bm{\theta}^{(t)}-\bm{\theta}^\star\right)-2\alpha \lambda \bm{\theta}^\star -
    \alpha \nabla\mathcal{L}_{0}(\bm{\theta}^\star),
\end{align*}
where $\bm{\xi}^{(t)}$ is a convex combination of $\bm{\theta}^{(t)}$ and $\bm{\theta}^\star$. Here $\Vert \bm{\theta}^{(t)}-\bm{\theta}^\star\Vert_{\infty} \leq 3q_0$ using (\ref{GDeq1}) and (\ref{GDeq3}). Therefore, $\Vert \bm{\xi}^{(t)}-\bm{\theta}^\star\Vert_{\infty} \leq 3q_0$, i.e., $\bm{\xi}^{(t)} \in \mathcal{C}(3q_0)$. Next, we consider the events
\begin{align*}
&E_3 = \left\{
C_{g',L}(3q_0)mp/4 \leq 
\Lambda_{min,\perp}
\left(
\nabla^2\mathcal{L}_{0}(\bm{\xi}^{(t)})
\right) \leq 
\Lambda_{max}
\left(
\nabla^2\mathcal{L}_{0}(\bm{\xi}^{(t)})
\right)  \leq 2C_{g',U} (3q_0)mp \right\} \\
&E_4 = \left\{
    \Vert \nabla\mathcal{L}_{0}(\bm\theta^\star) \Vert_2
    \leq 
    \frac{m^{3/2}pC_{g,U}(3q_0)}{\log(m)}
\right\},
\end{align*}
By Lemmas \ref{Lemma:Hessian} and \ref{Lemma:Gradient}, $E_3$ and $E_4$ occur simultaneously with probability at least $1- m \exp\left(-C_2 mLpB(\bm{\epsilon})\right)$. Conditional the event $E_3$ and $E_4$, we have
\begin{align*}
    \Vert \bm{\theta}^{(t+1)}-\bm{\theta}^\star \Vert_2 
    \leq & 4\left[1-\alpha  C_{g',L}(3q_0) mp/4\right]
\sqrt{\frac{m}{\log m}}\kappa
+
    2\alpha \lambda \Vert \bm{\theta}^\star\Vert_2+
    \alpha  \Vert  \nabla\mathcal{L}_{0}(\bm{\theta}^\star) \Vert_2 \\
    \leq &4\sqrt{\frac{m}{\log m}}\kappa -
     \alpha C_{g',L}(3q_0) mp\sqrt{\frac{m}{\log m}}\kappa
    +\alpha  \Vert  \nabla\mathcal{L}_{0}(\bm{\theta}^\star) \Vert_2  +2 \alpha \lambda \sqrt{m} \kappa \\
    \leq &  4\sqrt{\frac{m}{\log m}}\kappa - 
    \alpha C_{g',L}(3q_0) pm^{3/2}\sqrt{\frac{1}{\log m}}\kappa
    +\alpha m^{3/2}p C_{g,U}(3q_0)/\log(m) \\
    \leq  & 4\sqrt{\frac{m}{\log m}}\kappa,
\end{align*}
with probability at least $1- m \exp\left(-C_2 mLpB(\bm{\epsilon})\right)$, where the last inequality follows by the condition that $2\lambda \lesssim C_0(3q_0) m p$. This completes the proof of second step. 

\vspace{3mm}
\noindent
\textbf{(Step 3. Bounding $\max_{k\in [m]}|\theta_k^{(t+1,k)}-\theta_k^\star|$.)} For each $k\in [m]$, we have
\begin{align*}
    &\theta_k^{(t+1,k)}-\theta_k^\star  \\
    = &
   \theta_k^{(t,k)}-\theta_k^\star+
  \alpha  p
\sum_{i \in [m]\setminus k}
 \left\{
F(\theta_i^\star-\theta_k^\star) g(\theta_i^{(t,k)}-\theta_k^{(t,k)})
-
(1-F(\theta_i^\star-\theta_k^\star))
g(\theta_k^{(t,k)}-\theta_i^{(t,k)})\right\}-2
\lambda \alpha \theta_k^{(t,k)}.
\end{align*}
Note that $F(\theta_i^\star-\theta_k^\star) g(\theta_i^\star-\theta_k^\star)-(1-F(\theta_i^\star-\theta_k^\star))g(\theta_k^\star-\theta_i^\star)=0$. By the mean value theorem, we have
\begin{align*}
  &  F(\theta_i^\star-\theta_k^\star) g(\theta_i^{(t,k)}-\theta_k^{(t,k)})
+
(1-F(\theta_i^\star-\theta_k^\star))
g(\theta_k^{(t,k)}-\theta_i^{(t,k)}) \\
=&\underbrace{
\left[
F(\theta_i^\star-\theta_k^\star) g'(\gamma_{ik}^{0})
+
(1-F(\theta_i^\star-\theta_k^\star))
g'(-\gamma_{ik}^{0})
\right]}_{\triangleq  J_{ik}} \cdot (\theta_i^{(t,k)}-\theta_k^{(t,k)} -\theta_i^\star+\theta_k^\star ),
\end{align*}
where $\gamma_{i,k}^{0}$ is a convex combination of $\theta_i^\star-\theta_k^\star$ and $\theta_i^{(t,k)}-\theta_k^{(t,k)}$. With this, we can rewrite $\theta_k^{(t+1,k)}-\theta_k^\star$ as 
\begin{align*}
    \theta_k^{(t+1,k)}-\theta_k^\star = 
    \left(1-2\lambda \alpha-
    \alpha p
    \sum_{i \in [m]\setminus \{k\}} J_{ik}
    \right)
    \left(
    \theta_k^{(t,k)}-\theta_k^\star\right)-2\lambda \alpha 
    \theta_k^\star+\alpha  p
\sum_{i \in [m]\setminus k} J_{ik}
    (\theta_i^{(t,k)}-\theta_i^\star).
\end{align*}
Next, we turn to bound all terms on the right-hand side. First, we note that $\Vert \bm{\theta}^{(t,k)} - \bm{\theta}^\star\Vert_{\infty} \leq 3q_0$ via (\ref{GDeq1}) and (\ref{GDeq3}). Therefore, we have
\begin{align*}
  & \left| \sum_{i \in [m]\setminus k} J_{ik}
    (\theta_i^{(t,k)}-\theta_i^\star)\right| \leq 
    C_{g',U}(3q_0) 
    \Vert 
    \bm{\theta}^{(t,k)} - \bm{\theta}^\star\Vert_1 \\
    \leq & 
    \sqrt{m}C_{g',U}(3q_0)\Vert 
    \bm{\theta}^{(t,k)} - \bm{\theta}^\star\Vert_2\leq 
    C_{g',U}(3q_0)\left(\sqrt{m}+\frac{4m\kappa}{\sqrt{\log m}}\right).
    \end{align*}
Similarly, we also have $\sum_{i \in [m]\setminus \{k\}} J_{ik} \geq m C_{g',L}(3q_0)/2$. To sum up, $|\theta_k^{(t+1,k)}-\theta_k^\star|$ can be bounded as
\begin{align*}
    \left|\theta_k^{(t+1,k)}-\theta_k^\star\right|
    \leq & \left[1-2\lambda \alpha - \alpha mp C_{g',L}(3q_0)/2\right]q_0  + \alpha  p C_{g',U}(3q_0)\left(\sqrt{m}+\frac{4m\kappa}{\sqrt{\log m}}\right) \\
    \leq &\left[1- \alpha mp C_{g',L}(3q_0)\right]q_0  + \frac{5\alpha m p C_{g',U}(3q_0)(1+\kappa)}{\sqrt{\log m}}
    \leq q_0,
\end{align*}
where the last inequality holds by choosing $5(1+\kappa)C_{g',U}/(C_{g',L}(3q_0)\sqrt{\log(m)}) \leq q_0 $. This completes the proof of the third step. Therefore, it follows that (\ref{GDeq1})-(\ref{GDeq3}) hold true for $0 \leq t \leq t^\star$ with probability at least $1-C t^\star/m^{10}$ for some universal constant $C$.

Next, we turn to analyze the convergence behavior of $\bm{\theta}^{(t^\star)}$ towards $\widehat{\bm{\theta}}$. Note that $\nabla\mathcal{L}_0(\widehat{\bm{\theta}})+2\lambda \widehat{\bm{\theta}}=0$ due to the optimality of $\widehat{\bm{\theta}}$.
\begin{align*}
    \bm{\theta}^{(t+1)}-\widehat{\bm{\theta}}
    = \bm{\theta}^{(t)}
    -\widehat{\bm{\theta}}-
         \alpha \left(
\nabla\mathcal{L}_0(\bm{\theta}^{(t)})+2\lambda \bm{\theta}^{(t)}\right) 
= \left[(1-2\alpha \lambda )\bm{I}_m-
\alpha \nabla^2\mathcal{L}_0(\bm{\xi}^{(t)})
\right]\left(\bm{\theta}^{(t)}
    -\widehat{\bm{\theta}}\right),
\end{align*}
where $\bm{\xi}^{(t)}$ is a convex combination of $\bm{\theta}^{(t)}$ and $\bm{\theta}^\star$. By Lemmas \ref{Lemma:Hessian} and \ref{Lemma:LargestEigen}, we have
\begin{align*}
    \Lambda_{max}(\nabla^2\mathcal{L}_0(\bm{\xi}^{(t)})) \leq 
    2C_{g',U} (q)mp
    \mbox{ and }
    \Lambda_{min}(\nabla^2\mathcal{L}_0(\bm{\xi}^{(t)})) \geq 0,
\end{align*}
with probability at least $1-m \exp\left(-\frac{3mpLB(\bm{\epsilon})}{32}\right)-
    m \exp\left(-\frac{3C_{g',L}^2(3q_0) mLpB(\bm{\epsilon})}{128  C_{g',U}^2(3q_0)}\right)$.

Therefore, we have
\begin{align*}
 \Vert \bm{\theta}^{(t+1)}-\widehat{\bm{\theta}}\Vert_2
 \leq &
 \left(
 1-\frac{\lambda}{\lambda+C_{g',U}(3q_0)mp}
 \right)
 \Vert \bm{\theta}^{(t)}-\widehat{\bm{\theta}}\Vert_2
\\
 \leq &
 \left(
 1-\frac{1}{1+CmpLB(\bm{\epsilon})}
 \right)
 \Vert \bm{\theta}^{(t)}-\widehat{\bm{\theta}}\Vert_2 \triangleq \rho \Vert \bm{\theta}^{(t)}-\widehat{\bm{\theta}}\Vert_2 ,
\end{align*}
where the second inequality follows by setting $\lambda \asymp \frac{1}{LB(\bm{\epsilon})}$. Clearly, if $t^\star \asymp m^2$, we have $\left(
 1-\frac{1}{1+CmpLB(\bm{\epsilon})}
 \right)^{t^\star} \sqrt{m} =o(1)$. With this, we have
\begin{align*}
\Vert \widehat{\bm{\theta}} - \bm{\theta}^\star\Vert_{\infty} 
    \leq &
    \Vert \bm{\theta}^{(t^\star)} - \widehat{\bm{\theta}} \Vert_{\infty}+
  \Vert \bm{\theta}^{(t^\star)} - 
    \bm{\theta}^\star\Vert_{\infty} \leq 
\Vert \bm{\theta}^{(t^\star)} - \widehat{\bm{\theta}} \Vert_{2}+
  \Vert \bm{\theta}^{(t^\star)} - 
    \bm{\theta}^\star\Vert_{\infty} \\
    \leq &
    \rho^{t^\star}
    \Vert \bm{\theta}^{(0)} - \widehat{\bm{\theta}} \Vert_{2}+
  \Vert \bm{\theta}^{(t^\star)} - 
    \bm{\theta}^\star\Vert_{\infty} 
    \leq \rho^{t^\star}
    \Vert  \bm{\theta}^\star - \widehat{\bm{\theta}} \Vert_{2}+
     \rho^{t^\star}
    \Vert  \bm{\theta}^\star - \bm{\theta}^{(0)}\Vert_{2}+q_0 \\
    \leq &
    \rho^{t^\star} \sqrt{m}
    \Vert  \bm{\theta}^\star - \widehat{\bm{\theta}} \Vert_{\infty}+
     \rho^{t^\star}
   \sqrt{m}q_0+q_0 
   \leq \frac{1}{3}\Vert \widehat{\bm{\theta}} - \bm{\theta}^\star\Vert_{\infty}+2q_0.
\end{align*}
Asymptotically, if we set $t^\star \asymp m^2$, we have $\left(1-\frac{1}{1+CmLpB(\bm{\epsilon})}\right)^{t^\star} \sqrt{m}$ converging to zero. Then, a union bound for ensuring the events (\ref{GDeq1})-(\ref{GDeq3}) for $1< t \leq t^\star$ show that it holds with probability at least $1-O(m^{-8})$ that $ \Vert \widehat{\bm{\theta}} - \bm{\theta}^\star\Vert_{\infty} \leq 3q_0$. The desired result immediately follows by setting $q = 3q_0$. 
This completes the proof. \qed  \\

\begin{lemma}
    \label{Lemma:Gradient}
    Under Assumptions \ref{Ass1} and \ref{Ass2}, for $q>0$, we define $\mathcal{C}(q)=\{\bm{\theta}: \Vert \bm{\theta} - \bm{\theta}^\star\Vert_{\infty} \leq q \}$. For any $0<v \leq  pm^{3/2}C_{g,U}(q) $, it holds true that
    \begin{align*}
      &  \mathbb{P}\left(
    \Vert \nabla\mathcal{L}_{0}(\bm\theta) \Vert_2>v
    \right) 
    \leq  m
    \exp\left(
-\frac{3v^2 G(\bm{\epsilon})}{8pm^2  C_{g,U}^2(q)}
    \right),
    \end{align*}
    where $C_{g,U}(q)=\max_{x \in [-\kappa-2q,\kappa+2q]}g(x)$ denotes the maximum of $g(x)$ on $[-\kappa-2q,\kappa+2q]$.
\end{lemma}

\noindent
\textbf{Proof of Lemma \ref{Lemma:Gradient}.} First, the $i$-th element of $\nabla\mathcal{L}_{0}(\bm\theta)$ is given as
\begin{align*}
\left[\nabla\mathcal{L}_{0}(\bm\theta)\right]_i=
    -
    \sum_{l=1}^L \left\{
\sum_{j \in [m]\setminus \{i\}}a_{ij}^{(l)}w_l\left(\widetilde{z}_{ij}^{(l)}g(\gamma_{ij})-(1-\widetilde{z}_{ij}^{(l)})
g(-\gamma_{ij})
\right)
\right\},
\end{align*}
where $\gamma_{ij}=\theta_i - \theta_j$. Therefore, $\Vert \nabla\mathcal{L}_{0}(\bm\theta) \Vert_2^2$ is given as
\begin{align*}
    \Vert \nabla\mathcal{L}_{0}(\bm\theta) \Vert_2^2 = 
    \sum_{i=1}^m 
    \left(
\sum_{l=1}^L \left\{
\sum_{j \in [m]\setminus \{i\}}a_{ij}^{(l)}w_l\left(z_{ij}^{(l)}g(\gamma_{ij})-(1-z_{ij}^{(l)})
g(-\gamma_{ij})
\right)
\right\}
    \right)^2.
\end{align*}
For each $i \neq j$ and $\bm{\theta} \in \mathcal{C}(q)$, we have
\begin{align*}
 &  w_l| \widetilde{z}_{ij}^{(l)}g(\gamma_{ij})-(1-\widetilde{z}_{ij}^{(l)})
g(-\gamma_{ij})| \leq  \frac{C_{g,U}(q)}{G(\bm{\epsilon})}  ,  \\
&\text{Var}\left[
a_{ij}^{(l)}w_l\left(
\widetilde{z}_{ij}^{(l)}g(\gamma_{ij})-(1-\widetilde{z}_{ij}^{(l)})
g(-\gamma_{ij}))\right)
\right] \leq \frac{p 
\left(\frac{e^{\epsilon_l}-1}{e^{\epsilon_l}+1}\right)^2
C_{g,U}^2(q)}{G^2(\bm{\epsilon})}.
\end{align*}
Applying the Berinstein's inequality yields that
\begin{align*}
    \mathbb{P}\left(
    \Vert \nabla\mathcal{L}_{0}(\bm\theta) \Vert_2^2>v^2
    \right) 
    \leq &
    m 
    \mathbb{P}\left\{
    \left[
\sum_{l=1}^L \left(
\sum_{j \in [m]\setminus \{i\}}w_l a_{ij}^{(l)}\left(\widetilde{z}_{ij}^{(l)}g(\gamma_{ij})-(1-\widetilde{z}_{ij}^{(l)})
g(-\gamma_{ij})
\right)
\right)
    \right]^2>v^2/m
    \right\}  \\
    \leq  & m
    \exp\left(
-\frac{3v^2 G(\bm{\epsilon})}{6pm^2  C_{g,U}^2(q)+2v \sqrt{m}C_{g,U}(q)}
    \right) \leq m
    \exp\left(
-\frac{3v^2 G(\bm{\epsilon})}{8pm^2  C_{g,U}^2(q)}
    \right),
\end{align*}
where the last inequality follows from the condition that $v \leq  pm^{3/2}C_{g,U}(q)$. This completes the proof. \qed \\

\begin{lemma}
    \label{Lemma:LargestEigen}
For any $\bm{\theta} \in \mathcal{C}(q)=\{\bm{\theta}: \Vert \bm{\theta} - \bm{\theta}^\star\Vert_{\infty} \leq q \}$, it holds true that 
\begin{align*}
\mathbb{P}
    \left(
\Lambda_{max}
\left(
\nabla^2\mathcal{L}_{0}(\bm\theta)
\right)  \leq 2C_{g',U} (q)mp
    \right)  
    \geq  1- m \exp\left(-\frac{3mpLB(\bm{\epsilon})}{32   }\right),
\end{align*}
where $\Lambda_{max}(\cdot)$ denotes the largest non-zero eigenvalue.
\end{lemma}

\noindent
\textbf{Proof of Lemma \ref{Lemma:LargestEigen}.} The proof is based on the matrix Bernstein inequality, as demonstrated in the proof of Lemma \ref{Lemma:Hessian}. The conditions for using the matrix Bernstein inequality are already verified in Lemma \ref{Lemma:Hessian}.
    \begin{align*}
    &\mathbb{P}
    \left(
\Lambda_{max}
\left(
\nabla^2\mathcal{L}_{0}(\bm\theta)
\right) >v
    \right)
    =
        \mathbb{P}
    \left(
\Lambda_{max}
\left(
\nabla^2\mathcal{L}_{0}(\bm\theta)-\mathbb{E}\left(\nabla^2\mathcal{L}_{0}(\bm\theta)\right)+\mathbb{E}\left(\nabla^2\mathcal{L}_{0}(\bm\theta)\right)
\right) >v
    \right) \\
    \leq &
            \mathbb{P}
    \left(
\Lambda_{max}
\left(
\nabla^2\mathcal{L}_{0}(\bm\theta)-\mathbb{E}\left(\nabla^2\mathcal{L}_{0}(\bm\theta)\right) \right)+
\Lambda_{max} \left(
\mathbb{E}\left(\nabla^2\mathcal{L}_{0}(\bm\theta)\right)
\right) >v
    \right) \\
    =& \mathbb{P}
    \left(
\Lambda_{max}
\left(
\nabla^2\mathcal{L}_{0}(\bm\theta)-\mathbb{E}\left(\nabla^2\mathcal{L}_{0}(\bm\theta)\right) \right) >v-
\Lambda_{max} \left(
\mathbb{E}\left(\nabla^2\mathcal{L}_{0}(\bm\theta)\right)
\right)
    \right).
\end{align*}
For any $\bm{u}\in \mathbb{R}^m$ such that $\bm{1}_m^T \bm{u}=0$ and $\Vert \bm{u}\Vert_2=1$, we have
\begin{align*}
  & \max_{\bm{\theta} \in \mathcal{C}(q) }\bm{u}^T   \mathbb{E}\left(\nabla^2\mathcal{L}_{0}(\bm\theta)\right)\bm{u} 
  = \max_{\bm{\theta} \in \mathcal{C}(q) } -p\sum_{i<j}
  \left[F(\theta_i^\star-\theta_j^\star)g'(\gamma_{ij})+(1-F(\theta_i^\star-\theta_j^\star))
g'(-\gamma_{ij}) 
\right](u_i-u_j)^2 \\
=&\max_{\bm{\theta} \in \mathcal{C}(q) }
-\frac{p}{2}\sum_{i \neq j}
  \left[F(\theta_i^\star-\theta_j^\star)g'(\gamma_{ij})+(1-F(\theta_i^\star-\theta_j^\star))
g'(-\gamma_{ij}) 
\right](u_i^2-2u_iu_j+u_j^2) \\
\leq &
C_{g',U}(q) \frac{p}{2}\sum_{i \neq j}(u_i^2-2u_iu_j+u_j^2)  \leq C_{g',U} (q)mp.
\end{align*}Therefore, we have
\begin{align*}
    \mathbb{P}
    \left(
\Lambda_{max}
\left(
\nabla^2\mathcal{L}_{0}(\bm\theta)
\right) >v
    \right) \leq 
    \mathbb{P}
    \left(
\Lambda_{max}
\left(
\nabla^2\mathcal{L}_{0}(\bm\theta)-\mathbb{E}\left(\nabla^2\mathcal{L}_{0}(\bm\theta)\right) \right) >v-
C_{g',U} (q)mp
    \right).
\end{align*}
Recall that $\nabla^2\mathcal{L}_{0}(\bm\theta)$ can be represented as $\nabla^2\mathcal{L}_{0}(\bm\theta)= \sum_{i < j}\sum_{l=1}^L Q_{ijl}(\bm{\theta})$, where $Q_{ijl}(\bm{\theta})$ is defined as
\begin{align*}
    \big(Q_{ijl}(\bm{\theta})\big)_{(k,m)} =  
    \begin{cases}
-a_{ij}^{(l)}w_l\left(\widetilde{z}_{ij}^{(l)}g'(\gamma_{ij})+(1-\widetilde{z}_{ij}^{(l)})
g'(-\gamma_{ij})
\right), k=m=i \mbox{ or } k=m=j,\\
a_{ij}^{(l)}w_l\left(\widetilde{z}_{ij}^{(l)}g'(\gamma_{ij})+(1-\widetilde{z}_{ij}^{(l)})
g'(-\gamma_{ij})
\right), (k,m)=(i,j) \mbox{ or } (k,m)=(j,i),\\
0, \mbox{otherwise}.\\
    \end{cases}
\end{align*}
Note that
\begin{align*}
  &  \Lambda_{max}\left(
Q_{ijl}(\bm{\theta})-\mathbb{E}
    \left( Q_{ijl}(\bm{\theta}) \right)
    \right) \leq 
    \frac{2C_{g',U}(q) }{G(\bm{\epsilon})}, \\
  &  \Lambda_{max}\left(
\mathbb{E}
\left(
    \sum_{i < j}\sum_{l=1}^L \left[Q_{ijl}(\bm{\theta})-\mathbb{E}
    \left( Q_{ijl}(\bm{\theta}) \right)\right]^T
    \left[Q_{ijl}(\bm{\theta})-\mathbb{E}
    \left( Q_{ijl}(\bm{\theta}) \right)\right]
    \right)
    \right)\leq \frac{4 p m C^2_{g',U}(q)  }{G(\bm{\epsilon})}.
\end{align*}
for each $i,j \in [m]$ and $l \in [L]$. 
Suppose that $0<v \leq  2 p m C_{g',U}(q) $, we have
\begin{align*}
    \mathbb{P}
    \left\{
\Lambda_{max}
\left(
\nabla^2\mathcal{L}_{0}(\bm\theta)-\mathbb{E}\left(\nabla^2\mathcal{L}_{0}(\bm\theta)\right)
\right) >v-
C_{g',U} (q)mp
    \right\} \leq m \exp\left(-\frac{3(v-
C_{g',U} (q)mp)^2G(\bm{\epsilon})}{32 p m C_{g',U}^2(q)  }\right).
\end{align*}
Choosing $v = 2C_{g',U} (q)mp$, we have
\begin{align*}
\mathbb{P}
    \left(
\Lambda_{max}
\left(
\nabla^2\mathcal{L}_{0}(\bm\theta)
\right) >2C_{g',U} (q)mp
    \right) 
    \leq   m \exp\left(-\frac{3mpLB(\bm{\epsilon})}{32   }\right).
\end{align*}
This completes the proof. \qed \\

\end{document}